Zagazig University
Faculty of Engineering
Department of Construction and Utilities

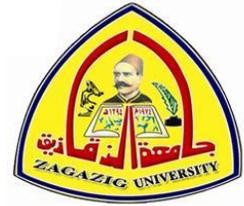

# PREDICTION OF CONSTRUCTION COST FOR FIELD CANALS IMPROVEMENT PROJECTS IN EGYPT

*Submitted by*

## Haytham H. Elmousalami

Bachelor of Civil Engineering
Zagazig University, 2013

A thesis submitted in partial fulfillment of the requirement for the degree of
**Master of Science in Construction Engineering and Utilities**

*Supervised by*

| **Ahmed H. Ibrahim** | **Ahmed H. Elyamany** |
|:---:|:---:|
| Associate professor | Assistant professor |
| Department of Construction and Utilities, | Department of Construction and Utilities, |
| Faculty of Engineering, | Faculty of Engineering |
| Zagazig University | Zagazig University |

**(January 2018)**

# Acknowledgement

First of all, all thanks and appreciations to Allah for his unlimited blessings. I wish to express my profound gratitude to:

| Name | Position | Role |
|---|---|---|
| Mohamed Mahdy Marzouk | Professor of Construction Engineering and Management, and Director of construction technology lab Structural Engineering Department, Faculty of Engineering, Cairo University | Examining committee |
| Osama Khairy Saleh | Professor of Water Engineering Department Water Engineering and Water Facilities Department Faculty of Engineering, Zagazig University | Examining committee |
| Ahmed Hussein Ibrahim | Associate Professor of Construction Engineering and Management Construction and Utilities Department, Faculty Of Engineering, Zagazig University | Supervision and Examining committee |
| Ahmed Hussein Elyamany | Associate Professor of Construction Engineering and Management Construction and Utilities Department, Faculty Of Engineering, Zagazig University | Supervision |

For their continued guidance, supervision, and comments throughout the course of this research. They have been ever-present force in helping me to mature as a student and as a researcher. Their dedication to helping me succeed is deeply appreciated.

My grateful thanks to all colleagues, contractors, consultants, and engineers who participated in filling questionnaires and provided important information for this study, and a special thank for my family members and the Ministry of Water resources and irrigation (MWRI) and Irrigation Improvement sector (IIS).

Zagazig University
Faculty of Engineering
Department of Construction and Utilities

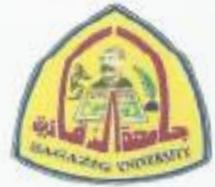

<u>**Approval Sheet**</u>

# PREDICTION OF CONSTRUCTION COST FOR FIELD CANALS IMPROVEMENT PROJECTS IN EGYPT

A dissertation by

### Haytham Hesham Elsayed Hashem ElMousalami

Submitted to the construction engineering and utility department, faculty of engineering, Zagazig University
In partial fulfillment of the requirement for the degree of Master of Science in Construction Engineering and Utilities

Approved as to style and content by the examining committee

| | | |
|---|---|---|
| Prof. Mohamed Mahdy Marzouk | Professor of Construction Engineering and Management Structural Engineering Department, Faculty of Engineering, Cairo University | |
| Prof. Osama Khairy Saleh | Professor of Water Engineering Department Water Engineering and Water Facilities Department Faculty of Engineering, Zagazig University | |
| Asso.Prof. Ahmed Hussein .Ibrahim | Associate Professor of Construction Engineering and Management Construction and Utilities Department, Faculty Of Engineering, Zagazig University | |

Approved Date: Thursday, 18/ January / 2018

# Abstract

Field canals improvement projects (FCIPs) are one of the ambitious projects constructed to save fresh water. To finance this project, Conceptual cost models are important to accurately predict preliminary costs at early stages of the project. The first step is to develop a conceptual cost model to identify key cost drivers affecting the project. Therefore, input variables selection remains an important part of model development, as the poor variables selection can decrease model precision. The study discovered the most important drivers of FCIPs based on a qualitative approach and a quantitative approach. Subsequently, the study has developed a parametric cost model based on machine learning methods such as regression methods, artificial neural networks, fuzzy model and case based reasoning.

There are several methods to achieve prediction for project preliminary cost. However, cost model inputs identification remains a challenging part during model development. Therefore, this study has conducted two procedures consisted of traditional Delphi method, Fuzzy Delphi Method (FDM) and the Fuzzy Analytic Hierarchy Process (FAHP) to determine these drivers. A Delphi rounds and Likert scale were used to determine the most important factors from viewpoints of consultant engineers and involved contractors. The study concluded that proposed approaches provided satisfying and consistent results. Finally, cost drivers of FCIPs were identified and can be used to develop a reliable conceptual cost model. On the other hand, the study has determined the key cost drivers for $FCIP_S$ based on quantitative data and statistical techniques. Factor analysis, regression methods and Correlation methods are utilized to identify cost drivers. In addition, this study has developed two hybrid models based on correlation matrix and stepwise regression which have identified the cost drivers more effectively that the other techniques. The key cost drivers are command area, PVC Length, construction year and a number of irrigation valves where the number of irrigation valves can be calculated as a function of the PVC length.

Once the key cost drivers of a project are identified, the parametric (algorithmic) cost model for the FCIPs is be developed. To develop the parametric cost model, two models are developed one by multiple linear regression and the other by artificial neural networks (ANNs). The results reveal the ability of both linear regression and ANNs model to predict cost estimate with an acceptable degree of accuracy. Sensitivity analysis is conducted to determine the contribution of selected key parameters. Finally, a simple friendly project data-input screen is created to facilitate usage and manipulation of the developed model. The research contribution has developed a reliable parametric model for predicting the conceptual cost of FCIPs with acceptable accuracy (9.12% and 7.82% for training and validation respectively).

Fuzzy systems have the ability to model numerous applications and to solve many kinds of problems with uncertainty nature such as cost prediction modeling. However, traditional fuzzy modeling cannot capture any kind of learning or adoption which formulates a problem in fuzzy rules generation. Therefore, hybrid fuzzy models can be conducted to automatically generate fuzzy rules and optimally adjust membership functions (MFs). This study has reviewed two types of hybrid fuzzy models: neuro-fuzzy and evolutionally fuzzy modeling. Moreover, a case study is applied to compare the accuracy and performance of traditional fuzzy model and hybrid fuzzy model for cost prediction where the results show a superior performance of hybrid fuzzy model than traditional fuzzy model.

# Contents

# List of Tables

# List of Figures

# List of Abbreviations

| | |
|---|---|
| **AI** | Artificial Intelligence |
| **AHP** | Analytic Hierarchy Process |
| **ANNS** | Artificial Neural Networks |
| **CBR** | Case-Based Reasoning |
| **CBR** | Case Based Reasoning |
| **CDPCE** | Cost Drivers Of Preliminary Cost Estimate |
| **CER** | Cost Estimation Relationships |
| **CFA** | Confirmatory Factor Analysis |
| **CI** | Computational Intelligence |
| **DSR** | Descriptive Statistics Ranking |
| **EC** | Evolutionary Computing |
| **EFA** | Exploratory Factor Analysis |
| **FAHP** | Fuzzy Analytic Hierarchy Process |
| **FCIPS** | Field Canal Improvement Projects |
| **FDM** | Fuzzy Delphi Method |
| **GA** | Genetic Algorithm |
| **GIS** | Geographical Information system |
| **KMO** | Kaiser-Meyer-Olken |
| **MAPE** | Mean Absolute Percentage Error |
| **MRA** | Multiple Regression Analysis |
| **MS** | Mean Score |
| **MF** | Membership Function |
| **ML** | Machine Learning |
| **MLP** | Multilayer Perceptron Network |
| **NA** | Note Available |
| **PCA** | Principal Component Analysis |
| **$R^2$** | The Coefficient Of Determination |
| **RII** | Relative Importance Index |
| **SCEA** | The Society Of Cost Estimating And Analysis |
| **SE** | The Standard Error |
| **SQRT** | Square Root |
| **TDM** | Traditional Delphi Method |

# CHAPTER 1

# INTRODUCTION

## 1.1 Background to Field Canal Improvement Projects (FCIPs)

Fresh water is naturally a limited resource on the earth plant. Many countries face a real challenge of future development due to water availability. In the twentieth century, population growth increased three-fold from 1.8 billion to 6 billion people. This reflects unsustainability where water usage during this period increased six-fold. 1.2 billion fellow humans have been expected to have no access to safe drinking water (Loucks et al, 2005). Therefore, a global trend is directed to save water and maintain its sustainability through several policies and projects. Field Canals Improvement Projects (FCIPs) are one of these projects where field canal's conveyance efficiency increased, on average, by 25% after improving the field canals during farm irrigation operations (Ministry of Public Works and Water Resources, 1998).

Field Canals Improvement Project (FCIP) is to construct a burden PVC pipeline instead of earthen field canal to decrease water losses due to water seepage losses and evaporations losses. FCIP consists of many simple structures and components. These components are plain concrete intake to take water from the water source (branch canal) where water is passing through suction pipes to a plain concrete sump, the objective of the sump is to accumulate water to be pumped by pumping sets located in a pump house. Water is pumped through PVC pipelines to be used by irrigation valves (Alfa - Alfa type). Based on the previous process, it can be concluded that the main components of FCIPs can be divided into three categories: civil works components, mechanical components, and electrical components. Civil works components are a pipeline, a pump house, a sump structure, suction pipes, and an intake. Mechanical components are pump sets, irrigation valves, and mechanical connections. Electrical components are electrical boards and electrical connections (Radwan, 2013). Fig.1.1 is a geographic information system (GIS) picture for FCIP planning where command area (area served) is 20.58 hectares. This figure illustrates PVC pipeline length, irrigation valves and location of the FCIP's station.

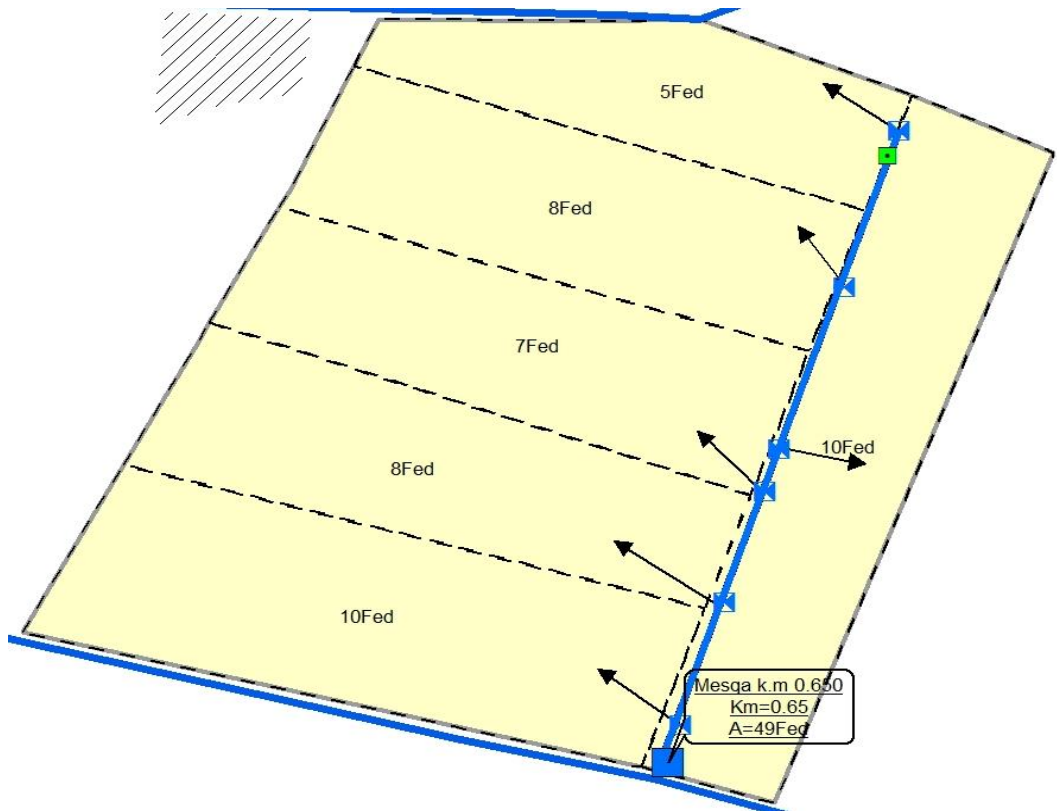

**Fig.1.1 GIS picture for FCIP planning at 0.65 km on Soltani Canal.**

## 1.2 Conceptual cost estimating

Conceptual cost estimating is one of the most important activities during project planning and feasibility study. The planning decisions of FCIPs in early stages are vital, as it can have the biggest influence on the total construction cost of the project. Conceptual cost estimating is the determination of the project's total costs based only on general early concepts of the project (Kan, 2002). Conceptual cost estimating is a challenging task that occurs at the early stages of a project where limited information is available and many factors affecting the project costs are unknown.

## 1.3 Research problem

Elfaki et al. (2014) indicated many research gaps where there is a crucial necessity need for a cost estimation method that covers all estimation factors. The study suggested a direction to avoid this gap by computerizing human knowledge. The challenge is identifying key cost drivers that have the highest influential impact on the final construction cost of FCIPs. Such parameters must be measurable for each new FCIP to be used in the conceptual cost estimation model. Conceptual cost estimating is the determination of the project's total costs based only on general early concepts of the project (Kan, 2002). Conceptual cost estimating is a challenging task that occurs at the early stages of a project where limited information is available and many factors affecting the project costs are unknown.

Inputs identification is one of the most important steps in developing a conceptual cost model. However, poor inputs selection can have the negative impact on the performance of the proposed model. Therefore, using experts' opinions helps the decision makers to evaluate the initial cost of FCIPs. Researchers usually depend on literature to know key cost drivers of a particular project. However, there is no sufficient literature about FCIPs cost drivers. Alternatively, researchers usually conduct interviews and Delphi rounds to discover these cost drivers. However, these methods cannot provide uncertainty that exists in the real words data.

Cost estimation traditionally starts with quantification that is a time intensive process. Currently, quantification is time-consuming which requires 50% to 80% of a cost estimator's time on a project (Sabol, 2008). Currently, both project owner and involving contractors uses traditional methods such as taking experts 'opinions to predict preliminary costs of FCIPs. A cost estimation tool is required to help the decision makers to take decisions regarding financing the construction of FCIPs.

During the initiation phase of FCIPs, A preliminary cost estimate is required to secure sufficient fund for such projects. Subsequently, the importance of using a precise cost model to predict the preliminary cost estimate exists. To develop a precise cost model, historical data of FCIPs have been collected to evaluate and select cost drivers of FCIPs. The purpose of variables selection is to improve the prediction accuracy and provide a better understanding of collected data (Guyon and Elisseeff, 2003).

## 1.4 Research Objectives

The research objective is developing a reliable parametric cost estimation model at the conceptual phase for Field Canals Improvement Projects (FCIPs). The objectives of this study are:

1. Identify the key conceptual cost drivers affecting the accuracy of cost estimation of FCIPs based on qualitative methods such as Delphi method that depends only on experts' judgments. The objective of this study is to identify FCIPs' cost drivers by qualitative methods such as Delphi method that depends only on experts' judgments for a process evaluation. To consider uncertainty, this study has applied fuzzy Delphi method and fuzzy analytical hierarchy process.

2. Identify the key conceptual cost drivers affecting the accuracy of cost estimation of FCIPs based on using historical quantitative data. The study objective aims identifying FCIPs' cost drivers of preliminary cost estimate (CDPCE) by using historical quantitative data. Experts' opinions are not utilized here to avoid biased selection when using human judgment. The purpose of the study is to discover and apply data-driven methods to select the key cost drivers based only on quantitative collected past data. The importance of cost drivers is to help decision makers to predict the preliminary cost of FCIPs and study the financial feasibility of these projects.

3. Develop a comprehensive tool for parametric cost estimation using multiple regression analysis and the optimum Neural Network model. The research objective is developing a reliable parametric cost estimation model before the construction of Field Canals Improvement Projects (FCIPs) by using Multiple Regression Analysis (MRA) and Artificial Neural Networks (ANNs). Therefore, a total of 144 FCIPs of constructed projects are collected to build up the proposed model.

## 1.5 Research Importance

The contributions of this thesis are expected to be relevant to both researchers and practitioners:

First, to researchers, the findings should help to investigate the accuracy of applying qualitative methods such as Delphi rounds and quantitative methods such as factor analysis to identify key cost drivers of a certain case study. In addition, this study will maintain the ability of regression analysis and Artificial Neural Network model to develop a reliable parametric cost estimation model.

Second, as for practitioners, the findings should help to easily estimate the cost of FCIPs after programing the developed model into a marketing program.

## 1.6 Research Scope and Limitation

This research focuses on Irrigation Improvement Sector of in Egypt; including the main projects of this sector that were implemented between 2010 and 2015 were collected.

## 1.7 Methodology Outline

The objectives of this study are achieved through performing the following steps:

1.1 Conduct a literature review of previous studies that are related to construction cost estimate and paying special attention of using Delphi rounds, Factor analysis, Regression analysis, and ANN.

1.2 Conduct quantitative and qualitative surveying techniques to identify the key factors on cost of FCIPs.

1.3 Conduct Delphi rounds and exploratory interviews with all experts to obtain the relevant data of FCIPs.

1.4 Conduct factor analysis and quantitative methods on historical data to identify key cost drivers.

1.5 Select the final key cost drivers based on both qualitative and quantitate methods.

1.6 Select the application SPSS software to be used in modeling regression analysis and the neural network.

1.7 Examine the validity of the adopted model by using statistical performance measurements and applying sensitivity analysis.

## 1.8 Research Layout

The current study was included eight chapters explained as follow:

Chapter (1) Introduction

An introductory chapter defines the problem statement, the objectives of this study, the methodology and an overview of this study.

Chapter (2) literature Review

Presents a literature review of traditional and present efforts that are related to the parametric cost estimating, and application of Delphi rounds, Factor Analysis, Regression Analysis and Artificial Neural Network (ANN) model in related field with its characteristics and structures.

Chapter (3) Research Methodology

The adopted methodology in this research was presented in this chapter including the data-acquisition process of cost drivers that relate to cost estimating of FCIPs

Chapter (4) Qualitative methods

Presents statistical analysis for questionnaire surveying, Delphi technique and hierarchy process. It also presents the cost drivers in this study

Chapter (5) Quantitative methods

Presents statistical analysis for collected historical data and Factor Analysis. It also presents the adopted cost drivers in this study and the encoded data for model implementation.

Chapter (6) Model Development

Presents the selected application software and type of model chosen and displays the model implementation, training and validation. As well, the results of the best model with a view of influence evaluation of the trained Regression mode and ANN model are showed.

Chapter (7) Automated fuzzy rules generation model

Presents two types of hybrid fuzzy models: neuro-fuzzy and evolutionally fuzzy modeling. Moreover, a case study is applied to compare the accuracy and performance of traditional fuzzy model and hybrid fuzzy model for cost prediction

Chapter (8) Conclusions and Recommendations

Presents conclusions and recommendations outlines for future work.

# CHAPTER 2

# LITERATURE REVIEW

## 2.1 Introduction

Cost estimating is a primary part of construction projects, where cost is considered as one of the major criteria in decision making at the early stages of the project. The accuracy of estimation is a critical factor in the success of any construction project, where cost overruns are a major problem, especially with current emphasis on tight budgets. Indeed, cost overruns can lead to cancellation of a project (Feng, et al., 2010; AACE, 2004).

Subsequently, the cost of construction project needs to be estimated within a specific accuracy range, but the largest obstacles standing in front of a cost estimate, particularly in early stage, are lack of preliminary information and larger uncertainties as a result of engineering solutions. As such, to overcome this lack of detailed information, cost estimation techniques are used to approximate the cost within an acceptable accuracy range (AACE, 2004).

Cost models provide an effective alternative for conceptual estimation of construction costs. However, development of cost models can be challenging as there are several factors affecting project costs. There are usually various and noisy data available for modeling. Parametric model mainly depends on parameters to simulate and describe the case studied (AACE, 2004; Elfaki, 2014). Poor identification of parameters means poor performance and accuracy of the parametric model. On the other hand, the optimal set of parameters produces the optimal performance of the developed model with less computation effort and less parameter needed to run the model (Kan, 2002).

## 2.2 Definitions

### 2.2.1 Cost Estimate

Dysert in (2006) defined a cost estimate as, "the predictive process used to quantify cost, and price the resources required by the scope of an investment option, activity, or project". Moreover, Akintoye & Fitzgerald (1999) defined cost estimate as, "is crucial to construction contact tendering, providing a basis for establishing the likely cost of resources elements of the tender price for construction work". Another definition was given by Smith & Mason (1997) which is "Cost estimation is the evaluation of many factors the most prominent of which are labor, and material".

The Society of Cost Estimating and Analysis (SCEA) defined the cost estimation as "the art of approximating the probable worth or cost of an activity based on information available at the time" (Stewart, 1991).

According to estimating methods, top-down and bottom-up approaches are the main two approaches of the cost estimate. On the one hand, top-down approach occurs at the conceptual phase and depends on the historical cost data where a similar project of the collect data is retrieved to estimate the current project. On the other hand, the bottom-up approach requires detailed information of the studied project. Firstly, all project is divided into items to create a cost breakdown structure (CBS). The main items of CBS are dependent on the amount of resources (labor, equipment, materials, and sub-contractors). The next step is to calculate the cost of each broken item and sum up for the total construction cost (AACE, 2004).

The estimating process consists of five main elements (Phaobunjong, 2002): project information, historical data, current data, estimating methodology, cost estimator, and estimates. Project information is the project characteristics that can be used as inputs to the cost model. Historical data are the collected data of the previous projects to statistically develop the cost model. Current data are the data extracted from the project information such as unit cost rates of material, labor, and equipment. Estimating methodology is the method used for cost estimate such as parametric cost model. The cost estimator is the user who uses the cost model and enter the input parameters or data to obtain the cost estimate. The estimates are the outputs of the cost model.

## 2.2.2 Construction Cost

The sum of all costs, direct and indirect, inherent in converting a design plan for material and equipment into a project ready for start-up, but not necessarily in production operation; the sum of field labor, supervision, administration, tools, field office expense, materials, equipment, taxes, and subcontracts (Humphreys, 2004; AACE, 2004).

## 2.2.3 Types of construction cost estimates

The type of estimate is a classification that is used to describe one of several estimate functions. However, there are different types of estimates which vary according to several factors including the purpose of estimates, available quantity and quality of information, range of accuracy desired in the estimate, calculation techniques used to prepare the estimate, time allotted to produce the estimate, phase of project, and perspective of estimate preparer (Humphreys, 2004). Generally, the main common types of cost estimates as outlined are:

(1) Conceptual estimate: a rough approximation of cost within a reasonable range of values, prepared for information purposes only, and it precedes design drawings. The accuracy range of this stage is -50% to +100%. Conceptual cost estimating is one of the most important activities during project planning and feasibility study. The planning decisions of FCIPs in early stages are vital, as it can have the biggest influence on the total construction cost of the project. Conceptual cost estimating is the determination of the project's total costs based only on general early concepts of the project (Kan, 2002). Conceptual cost estimating is a challenging task that occurs at the early stages of a project where limited information is available and many factors affecting the project costs are unknown (Choon & Ali, 2008; Abdal-Hadi, 2010).

(2) Preliminary estimate: an approximation based on well-defined cost data and established ground rules, prepared for allowing the owner a pause to review design before details. The accuracy range in this stage is -30% to +50%.
(3) Engineers estimate: Based on detailed design where all drawings are ready, prepared to ensure design is within financial resources and it assists in bids evaluating. The accuracy in this stage is -15% to +30%.
(4) Bid estimate: which done by contractor during tendering phase to price the contract. The accuracy in this stage is -5% to +15%.

For both preliminary and detailed technique its own methods, especially since preliminary methods are less numeric than detailed methods. However, most of researchers seek for perfect preliminary method that gives good results. Ostwald (2001) outlined commonly methods that are divided into two sets qualitative preliminary methods as opinion, conference, and comparison similarity or analogy and quantitative preliminary methods as unit method, unit quantity, linear regression...etc.

The following Table 2.1 summarizes the views of researchers about conceptual and detailed estimate (Al-Thunaian, 1996; Shehatto and EL-Sawalhi 2013").

**Table 2.1 Conceptual and Detailed Cost Estimates**

|  | Conceptual estimate | Detailed estimate |
|---|---|---|
| **When** | At the beginning of the project in feasibility stage and no drawing and details are available. | The scope of work is clearly defined and the detailed design is identified and a takeoff of their quantities is possible. |
| **Available of information** | No details of design and limited information on project scope are available. | Detailed specifications, drawings, subcontractors are available. |
| **Accuracy range** | -30% to +50% | -5% to +15% |
| **Purpose** | Determine the approximate cost of a project before making a final decision to construct it. | Determine the reliable cost of a project and make a contract. |
| **Requirements** | Clear understanding of what an owner wants and a good "feel" for the probable costs. | Analysis of the method of construction to be used, quantities of work, production rate and factors that affect each sub-item. |

### 2.2.4 Methods of cost estimation

Cost estimation methods can be categorized into several techniques as qualitative approaches and quantitative approach. Qualitative approaches rely on expert judgment or heuristic rules, and quantitative approaches classified into statistical models, analogous models and generative-analytical models (Duran, et al., 2009; Caputo & Pelagagge, 2008). Quantitative approach has been divided into three main techniques according to (Cavalieri, et al., 2004; Hinze, 1999).

(A) Analogy-based techniques

This kind of techniques allows obtaining a rough but reliable estimate of the future costs. It based on the definition and analysis of the degree of similarity between the new project and another one. The underlying concept is to derive the estimation from actual information. However, many problems exist in the application of this approach, such as:

1. The difficulties in the measure of the concept of degree of similarity.
2. The difficulty of incorporating in this parameter the effect of technological progress and of context factors.

(B) Parametric models

According to these techniques, the cost is expressed as an analytical function of a set of variables. These usually consist of some project features (performances, type of materials used), which are supposed to influence mainly the final cost of the project (cost drivers). Commonly, these analytical functions are named (Cost Estimation Relationships (CER)), and are built through the application of statistical methodologies. Parametric cost estimation is a method to develop cost estimating relationships with independent variables affecting the cost as a dependent variable. In addition, associated mathematical algorithms can be used to establish cost estimates (Hegazy and Ayed 1998).

(C) Engineering approaches

In this case, the estimation is based on the detailed analysis and features of the project. The estimated cost of the project is calculated in a very analytical way, as the sum of its elementary components, comprised by the value of the resources used in each step of the project process (raw materials, labor, equipment, etc.).
Due to this more details, the engineering approach can be used only when all the characteristics of the project process are well defined.

On the other hand, Cost estimating methods have been classified into four types (Dell'Isola, 2002): single-unit rate methods, parametric cost modeling, and elemental cost analysis and quantity survey. Single-unit rate methods are calculating the total cost of the project based on a unit such as the area of the building or an accommodation method such as cost per bed for hotels or hospitals. Parametric cost modeling is to develop a model based on parameters extracted from the collected data by conducting statistical analyses such as regression models, ANNs, and FL model. Elemental cost analysis is dividing the project into main elements and estimate the cost for each elements based on historical data. A quantity survey is a detailed cost estimate based on quantities surveyed and contract unit costs rates where such quantities include the resources used such as materials, labor, and equipment for each activity. Therefore, estimators usually apply single-unit rate method or parametric cost estimate in the conceptual stage.

## 2.3 Cost drivers identifications

Conceptual cost estimation mainly depends on the conceptual parameters of the project. Therefore, defining such parameters is the first and a critical step in the cost model development. This study has been conducted to review the common practices and procedures conducted to identify the cost drivers where the past literature have been classified into two main categories: qualitative and quantitative procedures. The objective is to review such procedures to get optimal cost model and to highlight of the future trends of cost estimation studies.

As illustrated in Fig.2.1, Such procedures can be categorized into two main procedures: qualitative procedure and quantitative procedure. The qualitative procedure includes all practices depends on the experts' questionnaire and gathering opinions. On the other hand, the quantitative procedure depends on the collected data where statistical techniques are required to discover and learn the patterns of data to extract the knowledge based on the collected data.

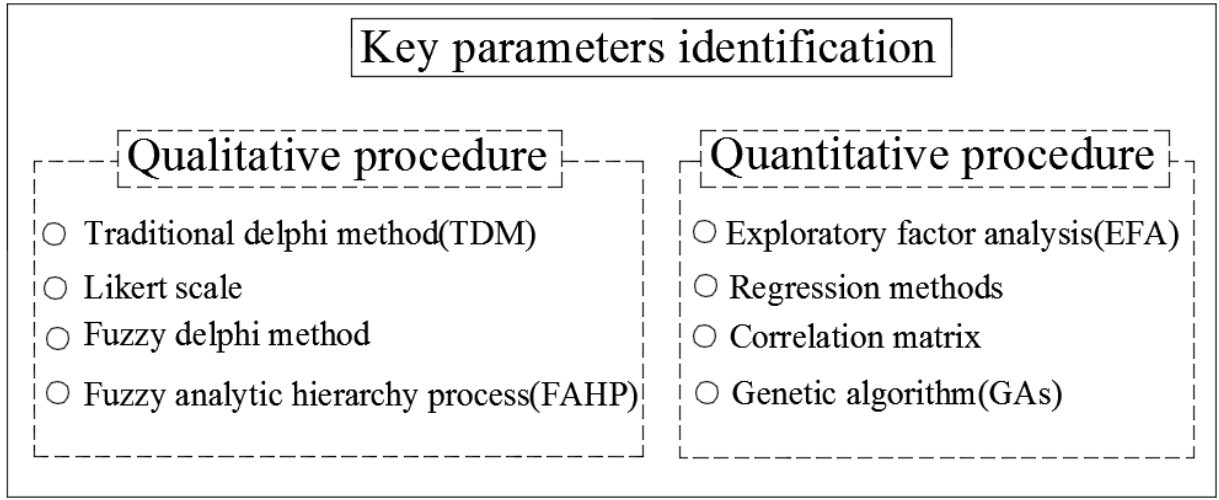

**Fig. 2.1. Qualitative and quantitative procedure.**

### 2.3.1 Qualitative procedure for key parameters identification

The qualitative procedures for key parameters identification are dependent on experts' interviews and field surveys. Many approaches such as traditional Delphi method, Likert scale, fuzzy Delphi method (FDM) and the fuzzy analytic hierarchy process (FAHP) have been conducted to select and evaluate the key cost drivers based on the viewpoints of experts.

#### 2.3.1.1 Traditional Delphi technique (TDM) and Likert scale

Traditional Delphi technique (TDM) is conducted to collect experts' opinions about a certain case. Based on experts' opinions, all parameters affecting a system can be identified. Delphi technique consists of several rounds for collecting, ranking and revising the collected parameters. Therefore, experts are also asked to give their feedback and revise their opinions to enhance the quality of the survey. Delphi rounds continue until no other opinions determined (Sandford and Hsu, 2007). Therefore, the first step is to select the experts to be asked based on their experience. The second step is to prepare a list of questions to discover the knowledge and parameters of the proposed case study. The third step is to apply Delphi rounds, where all experts should be asked through interviews or their answers can be collected via e-mails. The fourth step is to collect all experts' answers and make a list of all collected parameters. The fifth step is to ask experts again to assess and evaluate parameters. Finally, experts can revise their parameters and state the reasons for their rating (Sandford and Hsu, 2007).

Likert scale is a rating scale to represent the opinions of experts where Likert scale can be consisted of three points, five points or seven points. For example, a five-point Likert scale may be "Extremely Important", "Important", "Moderately Important", "Unimportant", and "Extremely Unimportant" where the experts will select these points to answer received questions (Bertram, 2017).

Based on the completed survey forms, statistical indices can be calculated to gather the final rank for each question or criteria based on experts ratings and to calculate sample adequacy of experts. Mean score (MS) used to gather the final rate for each criterion of the survey as Equation (2.1), whereas the standard error (SE) is calculated to check the sample size of experts as Equation (2.2).

$$MS = \frac{\sum(f*s)}{n} \qquad (1.1)$$

Where: (MS) is the man score to represent the impact of each parameter based on the respondent's answers, (S) is a score set to each parameter by the respondents, (f) is the frequency of responses to each rating for each impact of parameter, and (n) is total number of participants.

$$SE = \frac{\sigma}{\sqrt{n}} \qquad (1.2)$$

Where: (SE) is Standard Error, ($\sigma$) is the standard deviation among participants' opinions for each cost parameters, and (n) is a total number of participants. Thus, all parameters will be collected and ranked based on the experts' opinions.

### 2.3.1.2 Fuzzy Delphi Method (FDM)

FDM consists of the traditional Delphi method and fuzzy theory (Ishikawa et al., 1993). Maintaining the fuzziness and uncertainty in participants' opinions is the advantage of this method over traditional Delphi method. Instead of applying the experts' opinions as deterministic values, this method uses membership functions such as triangular, trapezoidal or Gaussian functions to map the deterministic numbers to fuzzy numbers. Accordingly, the reliability and quality of the Delphi method will be improved (Liu, 2013). The objective of the FDM is to avoid misunderstanding of the experts' opinions and to make a good generalization to the experts' opinions.

The first step of FDM is collecting initial parameters affecting on a proposed system like the first round of TDM. The second step is to assess each parameter by fuzzy terms where each Linguistic term consists of three fuzzy values (L, M, U) as shown in Fig.2.2 where μ(x) is a membership function. For example, unimportant term will be (0.00, 0.25, 0.50) and important term will be (0.50, 0.75, 1.00). The third step applies triangular fuzzy numbers to handle fuzziness of the experts' opinions where the minimum of the experts' common consensuses as point lij, and the maximum as point uij. This is illustrated in Equations 2.3, 2.4, 2.5, and 2.6 (Klir and Yuan, 1995):

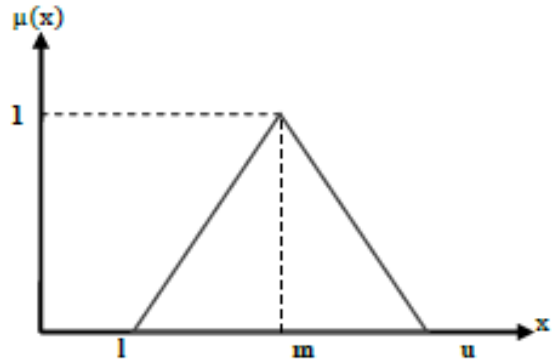

**Fig. 2.2. Triangular fuzzy number (Klir and Yuan, 1995).**

$$L_j = Min(L_{ij}), i = 1,2, \ldots \ldots n ; j = 1,2, \ldots m \tag{2.3}$$

$$M_j = (\prod_{i=1,j=1}^{n,m} m_{ij})^{1/n}, i = 1,2, \ldots \ldots n ; j = 1,2, \ldots m \tag{2.4}$$

$$U_j = Max(U_{ij}), i = 1,2, \ldots \ldots n ; j = 1,2, \ldots m \tag{2.5}$$

$$(W_{ij}) = (L_j, M_l, U_j) \tag{2.6}$$

Where:

i: an individual expert.

j: the cost parameter.

$l_{ij}$: the minimum of the experts' common consensuses.

$m_{ij}$: the average of the experts' common consensuses.

$u_{ij}$: the maximum of the experts' common consensuses.

$L_j$: opinions mean of the minimum of the experts' common consensuses ($l_{ij}$).

$M_j$: opinions mean of the average of the experts' common consensuses ($M_{ij}$).

$U_j$: opinions mean of the maximum of the experts' common consensuses ($U_{ij}$).

$W_{ij}$: The fuzzy number of all experts' opinions.

n: the number of experts.

m: the number of cost parameters.

The fourth Step is using a simple center of gravity method to defuzzify the fuzzy weight $w_j$ of each parameter to develop value $S_j$ by Equation (2.7).

$$Sj = \frac{Lj + Mj + Uj}{3} \qquad\qquad (2.7)$$

Where: $S_j$ is the crisp number after de-fuzzification process. Finally, the fifth step is that the experts provided a threshold to select or delete the collected parameters as following:

If $S_j \geq \alpha$, then the parameter should be selected.

If $Sj < \alpha$, then the parameter should be deleted.

Where $\alpha$ is the defined threshold. The FDM can be summarized as the following steps:

Step 1: Identify all possible variables affecting on a proposed system.
Step 2: Assess evaluation score for each parameter by fuzzy terms.
Step 3: Aggregate fuzzy numbers (Wij).
Step 4: Apply De-fuzzification (S).
Step 5: Defining a threshold (α).

### 2.3.1.3 Fuzzy Analytic Hierarchy Process (FAHP)

The Analytic Hierarchy Process (AHP) is a decision-making approach to evaluate and rank the priorities among different alternatives and criteria [(Saaty, 1980), and (Vaidya and Kumar, 2006)]. The conventional AHP cannot deal with the vague or imprecise nature of linguistic terms. Accordingly, Laarhoven and Pedrycz (1983) combined Fuzzy theory and AHP to develop FAHP. In traditional FAHP method, the deterministic values of AHP could be expressed by fuzzy values to apply uncertainty during making decisions. The aim is to assess the most critical cost parameters determined by FDM.

In FAHP, linguistic terms have been applied in pair-wise comparison which could be expresses by triangular fuzzy numbers (Srichetta and Thurachon, 2012) and (Erensal et al., 2006). (l, m, u) are triple triangular fuzzy set numbers that are used as a fuzzy values where $l \leq m \leq u$. Ma et al. (2010) applied the following steps:

Step 1: Identifying criteria and constructing the hierarchical structure.
Step 2: Setting up pairwise comparative matrices and transfer linguistic terms of positive triangular fuzzy numbers by linguistic scale of importance
Step 3: Generating group integration by Equation (2.8).

Step 4: Estimating the fuzzy weight.

Step 5: Defuzzify triangular fuzzy number into a crisp number.

Step 6: Ranking defuzzified numbers.

Experts' opinions are used to construct the fuzzy pair-wise comparison matrix to construct a fuzzy judgment matrix. After collecting the fuzzy judgment matrices from all experts using Equation (8), these matrices can be aggregated by using the fuzzy geometric mean (Buckley, 1985). The aggregated triangular fuzzy numbers of (n) decision makers' judgment in a certain case Wij = (Lij, Mij, Uij) where, for example, C, M, and, E refer to three different criteria respectively.

$$Wijn = \left( \ (\prod_{n=1}^{n} l\ ijn)^{\frac{1}{n}} \ , (\prod_{n=1}^{n} m\ ijn)^{\frac{1}{n}} \ , (\prod_{n=1}^{n} u\ ijn)^{\frac{1}{n}} \ \right) \qquad (2.8)$$

Where:

i: a criterion such as C, M or E.

j: the screened cost parameter for a defined case study.

n: the number of experts.

$l_{ij}$: the minimum of the experts' common consensuses.

$m_{ij}$: the average of the experts' common consensuses.

$u_{ij}$: the maximum of the experts' common consensuses.

$L_j$: opinions mean of the minimum of the experts' common consensuses ($l_{ij}$).

$M_j$: opinions mean of the average of the experts' common consensuses ($M_{ij}$).

$U_j$: opinions mean of the maximum of the experts' common consensuses ($U_{ij}$).

$W_{ij}$: the aggregated triangular fuzzy numbers of the $n^{th}$ expert's view.

Based on the aggregated pair-wise comparison matrix, the value of fuzzy synthetic extent Si with respect to the $i^{th}$ criterion can be computed by Equation (2.9) by algebraic operations on triangular fuzzy numbers [Saaty (1994) and Srichetta and Thurachon (2012)].

$$Si = \sum_{j=1}^{m} Wij * \left[ \sum_{i=1}^{n} \sum_{j=1}^{m} Wij \right]^{-1} \qquad (2.9)$$

Where i: a criterion, j: screened parameter, Wij: aggregated triangular fuzzy numbers of the $n^{th}$ expert's view, and Si: value of fuzzy synthetic extent. Based on the fuzzy synthetic extent values, this study used Chang's method (Saaty, 1980) to determine the degree of possibility by Equation (2.10). Accordingly, the degree of possibility can assess and evaluate the system alternatives.

$$V(S_m \geq S_c) = \left\{ \begin{array}{c} \mathbf{1}, \ if \ m_m \geq m_c \\ \mathbf{0}, \ if \ l_m \geq u_c \\ \dfrac{l_c - u_m}{(l_m - u_c) - (l_m - u_c)}, Otherwise \end{array} \right\} \quad (2.10)$$

Where:

V($S_m \geq S_c$): the degree of possibility between (C) criterion and (M) criterion.

($l_c$, $m_c$, $u_c$): the fuzzy synthetic extent of C criterion .

($l_m$, $m_m$, $u_m$): the fuzzy synthetic extent of M criterion.

## 2.3.2 Quantitative procedure for key parameters identification

The objective of variables identification is to increase the model prediction accuracy and provide a better understanding of collected data (Guyon and Elisseeff, 2003). Accurate cost drivers' identification leads to the optimal performance of the developed cost model. Quantitative methods depend on the collected data such as factor analysis, regression methods, and correlation methods.

### 2.3.2.1 Factor analysis

Factor analysis (FA) is a statistical method to cluster correlated variables to a lower number of factors where this method is used to filter data and determine key parameters. Many types of factoring exist such as principal component analysis (PCA), canonical factor analysis, and image factoring (Polit DF Beck CT, 2012). The advantage of exploratory factor analysis (EFA) is to combine two or more variables into a single factor that reduces the number of variables. However, factor analysis cannot provide results causality to interpret the data factored.

EFA is conducted by PCA to reduce the number of variables as well as to understand the structure of a set of variables (Field, 2009). The following questions should be answered before conducting EFA:

1) How large the sample needs to be?
2) Is there multicollinearity or singularity?
3) What is the method of data extraction?
4) What is the number of factors to retain?
5) What is the method of factor rotation?
6) Choosing between factor analysis and principal components analysis?

- **Sample size**

Factors obtained from small data sets cannot generalize as well as those derived from larger samples. Researchers have suggested using the ratio of sample size to the number of variables. Table 2.2 reviews such studies.

**Table 2.2. Survey of sample size for Factor Analysis.**

| | Reference | Summary of Findings |
|---|---|---|
| 1 | Nunnally (1978) | Sample size is 10 times of variables. |
| 2 | Kass and Tinsley (1979) | Sample size between 5 and 10 cases per variable. |
| 3 | Tabachnick and Fidell (2007) | Sample size is at least 300 cases. 50 observations are very poor, 100 are poor, 200 are fair, 300 are good, 500 are very good and 1000 or more is excellent. |
| 4 | Comrey and Lee (1992) | Sample size can be classified to 300 as a good sample size, 100 as poor and 1000 as excellent. |
| 5 | Guadagnoli and Velicer (1988) | A minimum sample size of 100 - 200 observations. |
| 6 | Allyn, Zhang, and Hong (1999) | The minimum sample size depends on the design of the study where a sample of 300 or more will probably provide a stable factor solution. |
| 7 | (Kaiser, 1970), Kaiser (1974), (Hutcheson & Sofroniou, 1999) | Based on the Kaiser–Meyer–Olkin measure of sampling adequacy (KMO) values (Kaiser, 1970), the values greater than 0.5 are barely acceptable (values below this should lead you to either collect more data or rethink which variables to include). Moreover, values between 0.5 and 0.7 are mediocre, values between 0.7 and 0.8 are good, values between 0.8 and 0.9 are great and values above 0.9 are superb. |
| 8 | (Kline, 1999) | The absolute minimum sample size required is 100 cases. |

- **Multicollinearity and singularity**

The first step is to check correlation among variables and avoid multicollinearity and singularity (Tabachnick and Fidell, 2007; Hays 1983). Multicollinearity means variables are correlated too highly, whereas singularity means variables are perfectly correlated. It is used to describe variables that are perfectly correlated (it means the correlation coefficient is 1 or -1). There are two methods for assessing multicollinearity or singularity:

1) The first method is conducted by scanning the correlation matrix among all independent variables to eliminate variables with correlation coefficients greater than 0.90 (Field, 2009; Hays 1983) or correlation coefficients greater than 0.80 (Rockwell, 1975).

2) The second method is to scan the determinant of the correlation matrix. Multicollinearity or singularity may be in existence if the determinant of the correlation matrix is less than 0.00001. One simple heuristic is that the determinant of the correlation matrix should be greater than 0.00001 (Field, 2009; Hays 1983). If the visual inspection reveals no substantial number of correlations greater than 0.3, PCA probably is not appropriate. Also, any variables that correlate with no others (r = 0) should be eliminated (Field, 2009; Hays 1983).

Bartlett's test can be used to test the adequacy of the correlation matrix. It tests the null hypothesis that the correlation matrix is an identity matrix where all the diagonal values are equal to 1 and all off-diagonal values are equal to 0. A significant test indicates that the correlation matrix is not an identity matrix where a significance value less than 0.05 and null hypothesis can be rejected (Dziuban et al, 1974).

According to factor extraction, Factor (component) extraction is conducting EFA to determine the smallest number of components that can be used to represent interrelations among a set of variables (Tabachnick and Fidell, 2007). Factors can be retained based on eigenvalues where a graph is known as a scree plot can be developed to retain factors (Cattell, 1966). All factors that have eigenvalues more than 1 can be retained (Kaiser, 1960). On the other hand, Jolliffe (1986) recommended to retaining factors that have eigenvalues more than (0.7).

According to Method of factor rotation, two type of rotation exists: orthogonal rotation and oblique rotation (Field, 2009). Orthogonal rotation can be varimax, quartimax and equamax. Whereas, oblique rotation can be direct oblimin and promax. Accordingly, the resulting outputs depend on the selected rotation method. For a first analysis, the varimax rotation should be selected to easily interpret the factors and this method can generally be conducted. The objective of the Varimax is to maximize the loadings dispersion within factors and to load a smaller number of clusters (Field, 2009).

Stevens (2002) concludes that no difference between factor analysis and component analysis exists if 30 or more variables and communalities greater than 0.7 for all variables. On the other hand, there is a difference between factor analysis and component analysis exists if the variables are fewer than 20 variables and low communalities (< 0.4).

### 2.3.2.2 Regression methods

Regression analysis can be used for both cost drivers selection and cost prediction modeling (Ratner, 2010). The current study focused on cost drivers' selection. Therefore, the forward, backward, stepwise methods are reviewed as follows.

Forward selection initiates with no variables in the model, where each added variable is tested by a comparison criterion to improve the model performance. If the independent variable significantly improves the ability of the model to predict the dependent variable, then this predictor is retained in the model and the method searches for a second independent variable (Field, 2009; Draper and Smith, 1998).

The backward method is the opposite of the forward method. In this method, all input independent variables are initially selected, and then the most unimportant independent variable are eliminated one-by-one based on the significance value of the t-test for each variable. The contribution of the remaining variable is then reassessed (Field, 2009; Draper and Smith, 1998).

Stepwise selection is an extension of the forward selection approach, where input variables may be removed at any subsequent iteration (Field, 2009; Draper and Smith, 1998). Despite forward selection, stepwise selection tests at each step for variables to be included or excluded where stepwise is a combination of backward and forward methods (Flom and Cassell, 2007).

### 2.3.2.3 Correlation method

The relation among all variables are shown in the correlation matrix, the aim is to screen variable based only on the correlation matrix. Therefore, all independent variables that are highly correlated with each other will be eliminated (R>= 0.8) and all dependent variables that are low correlated with the dependent variable (R <=0.3) will be eliminated. Such approach is dependent on the hypothesis that the relevant input independent variable is highly correlated with the output dependent variables and less correlated with the other input independent variables in the input subset (Ozdemir et al, 2001).

Pearson correlation is a measure of the linear correlation between two variables, giving a value between +1 and −1, where 1, 0, and -1 means positive correlation, no correlation, and negative correlation respectively. It is developed by Karl Pearson as a measure of the degree of linear dependence between two variables (Field, 2009).

Spearman correlation is a nonparametric measure of statistical dependence between two variables using a monotonic function. A perfect Spearman correlation of +1 or −1 occurs when each of the variables is a perfect monotone function of the other (Field, 2009).

### 2.3.2.4 Feature selection by GA

Genetic algorithm (GA) is an evolutionary algorithm (EA) used for search and optimization based on a fitness function (Siddique and Adeli, 2013). GA can be applied to select the input parameters of a prediction model such as artificial neural networks (ANNs). All irrelevant, redundant and useless parameters can be removed to reduce the size of the ANNs. The chromosome can be represented in a binary-coded where the bits number in the chromosome string equals the input variables number. This approach proposed by (Kohavi and John, 1997; Siedlecki and Sklansky (1988, 1989)) and called the wrapper approach to screen input variables (features).

As shown in Fig (2.3), data can be screened to key cost drivers where each chromosome represents a possible solution for input parameters. The chromosome consists of a binary gene where one represents the existence of a parameter, and zero represents the absent of the parameter. Each gene in the chromosome is associated with an input feature where the value of 1 represents the input feature existence and the value of 0 represents the elimination of this variable. Thus, the number of 1's in a chromosome is the number of the screened variables by GA. Chromosomes build a population of a set of possible solutions (Si). The objective of EA is to select the best subset of parameters (P) based on fitness function (F) that inherently minimizes the total system error. The fitness function is minimizing ANNs' prediction error. The main disadvantage of this methodology is high computation effort (Siddique and Adeli, 2013).

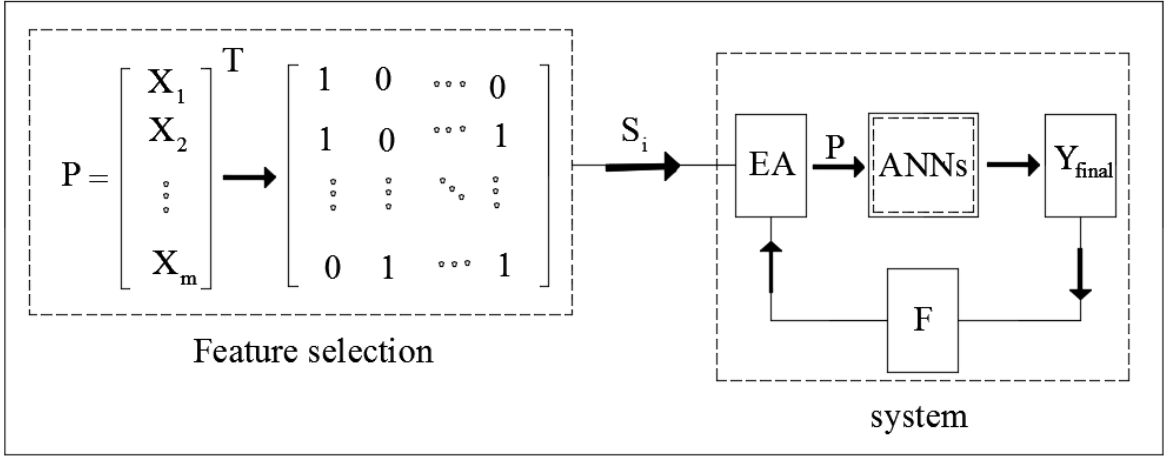

**Fig. 2.3. GA for cost driver identification (Siddique and Adeli, 2013).**

Fitness function is the guide to EA to convers, wrong fitness function formulation means false search and inaccurate optimization. Fitness function can be formulated in terms of minimum number of selected ANNs features, maximum accuracy and minimize computational cost (Yang and Honavar, 1998). Evaluation function can be formulated based on (Ozdemir et al., 2001).

### 2.3.2.5 Review of cost drivers identifications and discussion

Many past literature have been surveyed to identify the practices for cost drivers' identification in the construction industry. Many journals have been revised such as journal of construction engineering and management, the journal of construction engineering and management, the journal of civil engineering and management and construction management and economics.

Based on Table.2.3, the questionnaire survey approach is the most common approach conducted to identify and assess the cost drivers of a certain case study. Therefore, the qualitative approach is the most common than quantitative method. Such claim can be a result of no availability of data for the studied cases. Moreover, asking experts is a simple approach and needs no deep statistical knowledge, whereas the data-driven procedure requires statistical methods.

On the one hand, the most common methods in qualitative methods are questionnaire survey and AHP. On the other hand, the most common methods in quantitative methods are factor analysis and regression methods.

**Table. 2.3. Review of cost drivers' identification.**

| Reference | Method | Key findings |
|---|---|---|
| **Moselhi and hegazy, 1993** | questionnaire survey | A questionnaire survey has been conducted to discover the input variables for ANNs for markup estimation. |
| **Attalla and hegazy, 2003** | questionnaire survey | A questionnaire survey has been operated to identify the input variables for ANNs for cost deviation where 36 factors have been identified. |
| **Stoy et al., 2008 and Stoy et al. , 2007** | questionnaire survey + regression method | Based on 70 residential properties in German, Stoy et al have used regression method to select cost drivers. |
| **ElSawy, et al, 2011** | questionnaire survey | Based on Fifty-two s of building in Egypt, ten cost drivers have been selected by questionnaire survey of experts for ANNs cost model. |
| **Park and Kwon, 2011** | questionnaire survey + factor analysis | A questionnaire survey is applied to gather experts' opinions, whereas factor analysis is conducted to group the collected parameters into six groups. |
| **Marzouk and Ahmed, 2011** | questionnaire survey | A questionnaire survey has conducted to identify and evaluate fourteen parameters affecting on the costs of pump station projects. |
| **Petroutsatou et al, 2012** | questionnaire survey | A questionnaire survey has been conducted to determine significant parameters for ANNs cost prediction model for tunnel construction in Greece. |
| **El Sawalhi, 2012** | questionnaire survey | Both a questionnaire survey and relative index ranking technique have been conducted to investigate and rank the factors affecting the cost of building construction for fuzzy logic model. |
| **Petroutsatou et al, 2012** | questionnaire survey | A structured questionnaires have been conducted to collect data and all corresponding parameters for ANNs model development from different tunnel construction sites. |
| **Alroomi et al. ,2012** | questionnaire survey + factor analysis | Based on 228 completed questionnaires, all relevant cost data of competencies have been collected by experts, whereas the factor analysis has been conducted to investigate the correlation effects of the estimating competencies. |

| Reference | Method | Key findings |
|---|---|---|
| **Continue:** | **Table. 2.3. Review of cost drivers' identification.** | |
| **El-Sawalhi and Shehatto, 2014** | questionnaire survey | Eighty questionnaires have been conducted to determine significant variables for cost prediction for building project. |
| **El-Sawah and Moselhi, 2014** | trial and error approach | Based on trial and error approach and combination of input variables, ANNs models have been built for cost prediction of steel building and timber bridge. |
| **Choi et al, 2014** | questionnaire survey | Based on questionnaire survey, attributes of road construction project have been identified. |
| **Marzouk and Elkadi, 2016** | questionnaire survey + factor analysis | EFA is conducted to select cost drivers of water treatment plants where a total of 33 variables have been reduced to four components. Such components are used as inputs to ANNs model. |
| **Emsley et al, 2002** | literature survey + questionnaire survey + factor analysis | Based on 300 building projects, FL is investigated to select key cost drivers to be used by ANNs and regression models. |
| **Knight and Fayek, 2002** | literature survey + questionnaire survey | Based on past related literature and interview surveys, all parameters affecting on cost overruns for building projects have been identified and ranked for fuzzy logic model. |
| **Kim, 2013** | literature survey+ questionnaire survey | Based on past literature and interview surveys with experts, all parameters affecting on cost of highways project are identified for hybrid prediction model. |
| **Williams, 2002** | regression methods | Based on biding data, the stepwise regression method has been utilized to check the significance of each parameter and select the key cost drivers for regression model. |
| **Lowe and Emsley , 2006** | regression methods | Based on 286 sets of data collected in the United Kingdom, Both forward and backward stepwise regression have been used to develop six parametric cost models. |
| **Stoy, 2012** | regression methods | Backward regression method has been computed to determine key cost drives based on a total of 75 residential projects. |
| **Ranasinghe, 2000** | correlation method | This study presents induced correlation concept to analysis input cost variables for residential building projects in German. |
| **Yang , 2005** | correlation method | Correlation matrix should be scanned to reduce variable and to detect redundant variables. |

| Continue: | Table. 2.3. Review of cost drivers' identification. | |
|---|---|---|
| **Reference** | **Method** | **Key findings** |
| **Kim et al, 2005** | GA for parameter selection | This study has built three cost NN model by back propagation (PB) algorithm, GA for optimizing NN weights and GA for parameters optimization of BP algorithm. Optimizing parameters of BP algorithm produces the better results. |
| **Xu et al, 2015** | GA + Correlation method | Correlation method is used to rank model features, where GA is used for selecting the optimal subset of features for the model. |
| **Saaty, 2008** | AHP | For several application, the Analytic Hierarchy Process (AHP) has been conducted as a powerful decision-making procedure among different criteria and alternatives. |
| **Laarhoven and Pedrycz, 1983** | FAHP | AHP and Fuzzy Theory are combined to produce FAHP where the objective is to evaluate the most important cost parameters. |
| **Erensal et al., 2006** | FAHP | FAHP is conducted for evaluating key parameters in technology management. |
| **Pan, 2008** | FAHP | Fuzzy AHP is conducted to provide the vagueness and uncertainty for selecting a bridge construction method. According, FAHP obtains more reliable results than the conventional AHP. |
| **Manoliadis et al. , 2009** | FAHP | Based on qualifications survey, FDM is conducted to assess bidders' suitability for improving bidder selection. |
| **Ma et al., 2010** | FAHP | FAHP is conducted for Pile-type selection based on the collected field factors where fuzzy AHP approach produces an efficient performance for pile-type selection. |
| **Srichetta and Thurachon, 2012** | FAHP | FAHP is conducted for evaluating the notebook computers products. |
| **Hsu et al, 2010** | FDM + FAHP | This study utilizes two process of selection and decision making. FDM is the first process to identify the most important factors, whereas the second process is FAHP to identify the importance of each factor. |
| **Liu, 2013** | FDM + FAHP | Both FDM and FAHP are conducted to evaluate and filter all factors affecting on indicators of managerial competence. |

| Continue: | Table. 2.3. Review of cost drivers' identification. | |
|---|---|---|
| **Reference** | **Method** | **Key findings** |
| **Elmousalami et al, 2017** | FDM + FAHP + traditional Delphi method | This study has compared traditional Delphi method, FDM and FAHP to evaluate and select the key cost drivers of field canals improvement projects in Egypt. |

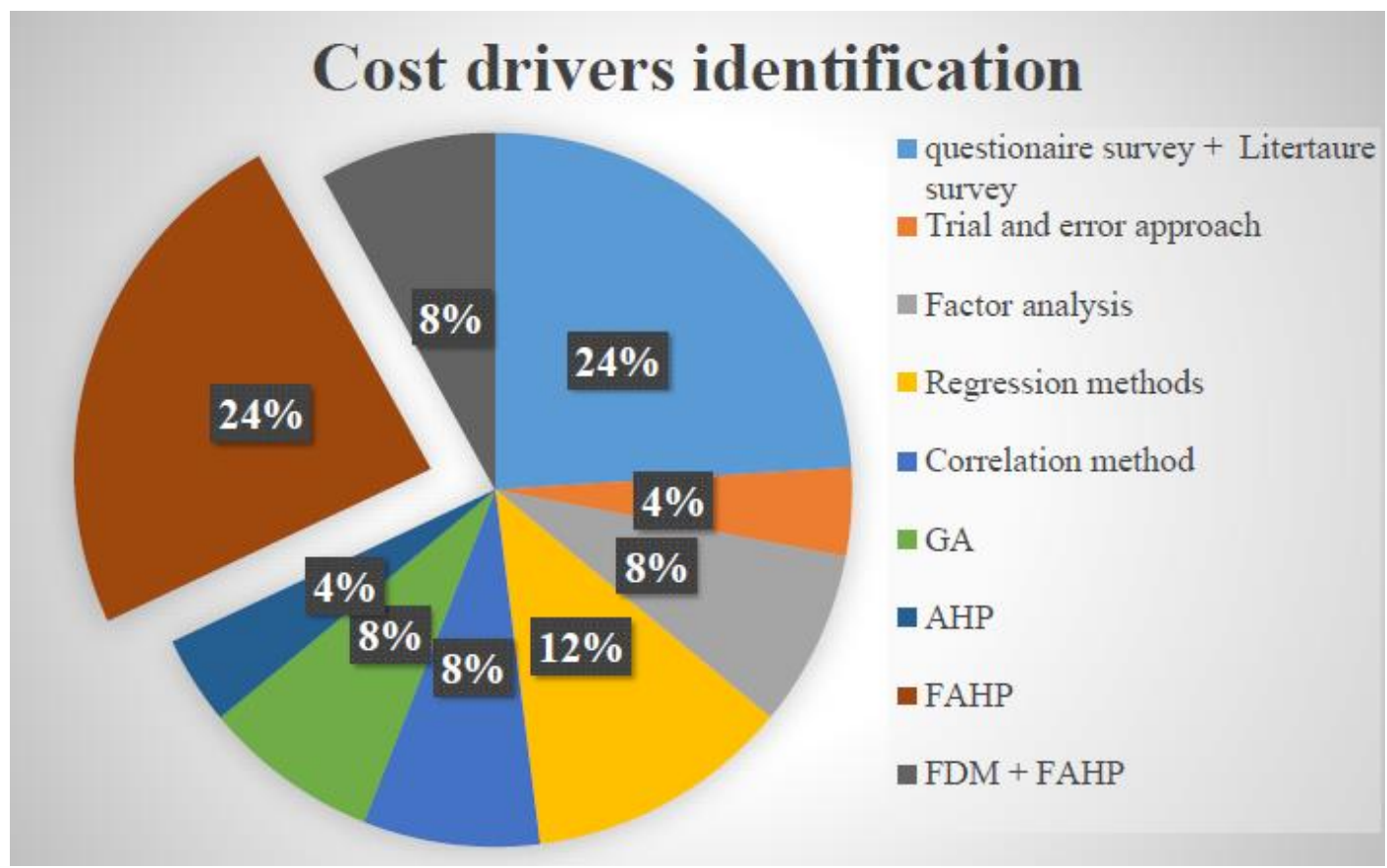

**Fig. 2.4 Cost drivers identification.**

As shown in Fig. 2.4, based on the survey literature, there are a variety of the used techniques where the FAHP is the most commonly used techniques. However, traditional techniques have many advantages such as identifying divergence of opinions among participants and share of knowledge and reasoning among participants. In addition, rounds enable participants to review, re-evaluate and revise all their previous opinions. Moreover, these methods need simple calculations and statistical equations. However, the major disadvantage of the traditional techniques is their inability to maintain uncertainty among different participants 'opinions. Accordingly, Information extracted from a selected group of experts may not be representative. Alternatively, the advanced techniques have applied fuzzy methods to maintain uncertainty among participants' opinions where this feature is the major advantage of the advanced technique. However, the advanced techniques require more calculations and statistical forms to be conducted. Generally, both methods are time-consuming to collect participant's opinions. As a result, it can be concluded that the advanced methods may more practical than traditional method in cost drivers' identification.

## 2.4 Parametric (algorithmic) construction cost estimate modeling

The conceptual cost estimate is one of the most critical processes during the project management. Parametric cost estimate modeling is one of the approaches used in the conceptual stage of the project. This study has discussed different computational intelligence techniques conducted to develop practical cost prediction models. Moreover, this study has discussed the hybridization of this model and the future trends for cost model development, limitations, and recommendations. The study focuses on reviewing the most common techniques which are conducted for cost modeling such as fuzzy logic (FL) model, artificial neural networks (ANNs), regression model, cased based reasoning (CBR), supportive vector machines (SVM), hybrid models, and evolutionary computing (EC) such as genetic algorithm (GA).

Different models can be conducted to predict the conceptual cost estimate for a project based on the key parameters of the project. This research aims to review the common computational intelligence (CI) techniques used for parametric cost models and to highlight the future trends. The accurate cost estimate is a critical aspect of the project's success (AACE, 2004; Hegazy, 2002). At conceptual stage, cost prediction models can be based on numerous techniques including statistical techniques such as regression analysis, probabilistic techniques such as Monte Carlo simulation, and artificial intelligence-based techniques such as SVM, ANNs and GA (Elfaki et al, 2014). Selecting the optimal technique for cost modeling aims to provide accurate results, minimizes the cost prediction error, and provides more practical model.

The study scope is conceptual cost estimate modeling. Conceptual estimating works as the main stage of project planning where limited project information is available and high level of uncertainty exists. Moreover, the estimating should be completed during a limited time period. Therefore, the accurate conceptual cost estimate is a challengeable task and a crucial process for project managers (Jrade, 2000).

### 2.4.1 Computational intelligence (CI)

Computational intelligence (CI) techniques are aspects of human knowledge and computational adaptively to become more vital in system modeling than classical mathematical modeling (Bezdek, 1994). Based on CI, an intelligent system can be developed to produce consequent outputs and actions depend on the observed input and output behavior of the system (Siddique and Adeli, 2013).

The objective is to solve complex real-world problems based on data analytics (such as classification, regression, prediction) and optimization in an uncertain environment. The core advantage of the intelligent systems are their human-like capability to make decisions depended on information with uncertainty and imprecision. The basic approaches to computational intelligence are fuzzy logic (FL), artificial neural networks (ANNs), evolutionary computing (EC) (Engelbrecht, 2002). Accordingly, CI is a combination of FL, neuro-computing and EC. The scope of this study will focus on the three methodologies of computational intelligence: FL, ANNs, EC and, its fusion.

### 2.4.2 Multiple regression analysis (MRA)

Multiple regression analysis (MRA) is a statistical analysis that uses given data for prediction applications. Based on historical cases, regression analysis develops a mathematical form to fit the given data (Field, 2009). This mathematical form can be formulated as Equation (2.11).

$$Y = B_0 + B_1X_1 + B_2X_2 + \ldots\ldots B_nX_n \qquad (2.11)$$

Where Y is dependent variable, $B_0$ is constant, $B_i$ is variable coefficients, and $X_i$ is independent variables. The change by 1 unit of the independent variable $X_1$ causes a change by $B_1$ in the dependent variable Y. Similarly, the change by 1 unit of the independent variable $X_2$ causes a change by $B_2$ in the dependent variable Y. In addition, the sign of $B_1$ and $B_2$ determines the decrease or increase in the dependent variable Y . The objective of the regression model is to mathematically represent data with the minimal prediction error .Therefore, regression analysis is applied in cost estimate modeling to represent the cost-estimate relationships where the cost prediction is represented as the dependent variable and the cost drivers are represented as the independent variables.

According to sample size, $(50 + 8k)$ may be the minimum sample size, where k is the number of predictors Green (1991). According to deleting outliers, Cook's distance detects the impact of a certain case on the regression model (Cook and Weisberg, 1982). If the Cook's values are < 1, there is no need to delete that case (Stevens, 2002). Otherwise, if the Cook's values are >1, there is a need to delete that case. According to multicollinearity and singularity, multicollinearity is the case that the variables are highly correlated whereas, the singularity is the case that the variables are perfectly correlated (Field, 2009). Variables are high correlated where the coefficient of determination is higher than 0.8 (r > 0.8) (Rockwell, 1975). The variance inflation factor (VIF) examines the linear relationship with the other variable (Field, 2009) where if average VIF is greater than 1, then multicollinearity occurs and can be detected (Bowerman and O'Connell, 1990; Myers, 1990). Homoscedasticity occurs when the residual terms vary constantly where the residual variance should be constant to avoid biased regression model (Field, 2009). The Durbin–Watson test is conducted to check the correlations among errors where the test values range between 0 and 4. The value of two refers that residuals are uncorrelated (Durbin and Watson's, 1951). Accordingly, regression models can be summarized in the following steps:

I.    Collect and prepare the historical cases.
II.   Divide the collected cases into a training set and a validation set.
III.  Check sample size of the collected training data (Green, 1991; Stevens, 2002).
IV.   Define key independent parameters (cost drivers) and dependent parameter (cost variable).
V.    Develop a regression model and check the significance (P-value) of each coefficient (Field, 2009).
VI.   Check outliers (Cook and Weisberg, 1982).
VII.  Check the variance inflation factor (VIF) (Bowerman and O'Connell, 1990; Myers, 1990).
VIII. Check homoscedasticity (Durbin and Watson's, 1951).
IX.   Calculate the resulting error such as mean absolute percentage error (MAPE).

### 2.4.3 Fuzzy logic (FL)

Fuzzy logic (FL) is modeling the human decision making by representing uncertainty, incompleteness, and randomness of the real world system (Zadeh, 1965, 1973). In addition, FL represents the experts' experience and knowledge by developing fuzzy rules. Such knowledge represents in fuzzy systems by membership functions (MFs) where MFs ranges from zero to one. MFs can be triangular, trapezoidal, Gaussian and bell-shaped functions where the selection of the MF function is problem-dependent. Fig.2.5. illustrates a trapezoidal MF consists of core set {$a_2$, $a_3$} and support set {$a_1$, $a_2$, $a_3$, $a_4$}. The shape of MF significantly influences the performance of a fuzzy model (Wang, 1997; Chi et al., 1996). Therefore, many methods are applied to develop MFs automatically such as clustering approach and to select the optimal shape of MFs.

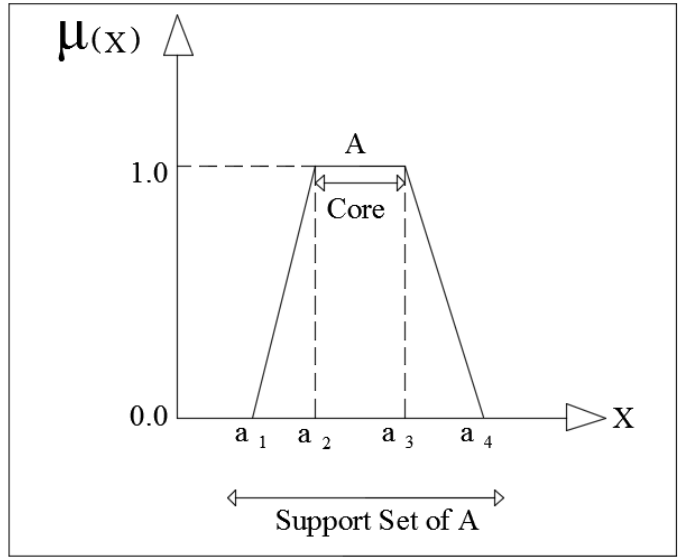

**Fig.2.5 Fuzzy trapezoidal membership function (Siddique and Adeli, 2013).**

Once MFs can be identified for each dependent and independent parameters, a set of operations on fuzzy sets can be conducted. Such operations are union of fuzzy sets, intersection of fuzzy sets, and complement of fuzzy set and α-cut of a fuzzy set. Linguistic terms are used to approximately represent the system features where such terms cannot be represented as quantitative terms (Zadeh, 1976). Once MFs and linguistic terms have been defined, IF-then rules can be developed to establish rule-based systems. Each rule presents rule to represent human logic and experience where all rules represent the brain of the fuzzy system.

Fuzzification is transforming crisp values into fuzzy inputs. Conversely, defuzzification is transforming of a fuzzy quantity into a crisp output. Many different methods of defuzzification exist such as max-membership, centre of gravity, weighted average, mean-max and centre of sums Runker (1997). Inference mechanism is the process of converting input space to output space such as Mamdani fuzzy inference, Sugeno fuzzy inference, and Tsukamoto fuzzy inference (Mamdani and Assilian, 1974; Takagi and Sugeno, 1985; Sugeno and Kang, 1988; Tsukamoto, 1979).

Fuzzy modeling identification includes two phases: structure identification and parameter identification (Emami et al., 1998). Structure identification is to define input and output variables and to develop input and output relations through if–then rules. The following points summarize the structure identification of fuzzy system

I.    Determine of relevant inputs and outputs.

II.    Selection of fuzzy inference system, e.g. Mamdani, Sugeno or Tsukamoto.

III.    Defining the linguistic terms associated with each input and output variable.

IV.    Developing a set of fuzzy if – then rules.

Whereas, the parameters identification is an optimization problem where the objective is to maximize the performance of the developed system. Defining MFs such as shape of MF (triangular, trapezoidal, Gaussian and bell-shaped functions) and its corresponding values can significantly optimize the system performance.

### 2.4.4 Artificial neural networks (ANNs)

ANNs are biologically inspired model to mimic human neural system for information-processing and computation purposes. ANNs is a machine learning (ML) technique where can learn from past data. Learning forms can be supervised, unsupervised and reinforcement learning. Contrary to traditional modeling technique such as linear regression analysis, ANNs have ability to approximate any nonlinear function to a specified accuracy. The first model of artificial neural networks came in 1943 when Warren McCulloch, a neurophysiologist and Walter Pitts, a young mathematician outlined the first formal model of an elementary computing neuron (McCulloch and Pitts, 1943). The first model of ANNs proposed by Warren McCulloch to mimic human neural system, the model is based on electrical circuits' concept where the output is zero or one. This model is called as perceptron or neuron where such neuron is the unit of ANNs (McCulloch and Pitts, 1943). Hopfield connected theses neurons and develop a network to create ANNs (Hopfield, 1982). Generally, ANNs can be categorized to two main categories: feedforward network and recurrent network.

In a feedforward network, all neurons are connected together with connections. The feedforward network consists of input vector (x), a weight matrix (W), a bias vector (b) and an output vector (Y) where it can be formulated as Equation (2.12).

$$Y = f(W \cdot x + b) \qquad (2.12)$$

Where f (.) includes a nonlinear activation function. Different types of activation function exist such as linear function, step function, ramp function and Tan sigmoidal function. Selecting of ANNs parameters such as the number of neurons, connections transfer functions and hidden layers mainly depend on its application. Several types of feedforward neural network architectures exist such as multilayer perceptron networks (MLP), radial basis function networks, generalized regression neural networks, probabilistic neural networks, belief networks, Hamming networks and stochastic networks where each architecture is problem-dependent (Siddique and Adeli, 2013). In this study, multilayer perceptron networks (MLP) will explained in some detail.

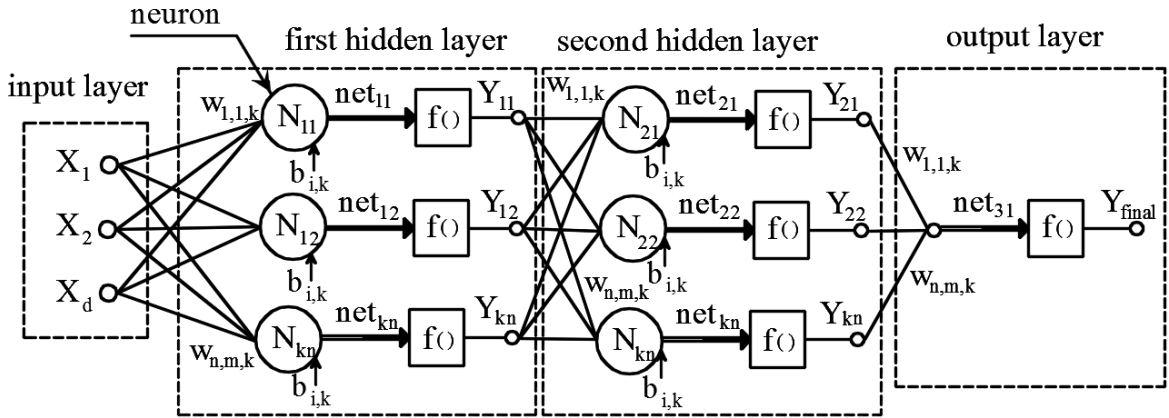

**Fig.2.6 Multilayer perceptron network (Siddique and Adeli, 2013).**

As shown in Fig.2.6, MLP network is a network with several layers of perceptions where each layer has a weight matrix (W), a bias vector (b) and output vector (Y). The input vector X = {$X_1$,$X_2$,$X_3$, ....} feed forward to (n) neurons in the hidden layer with a transfer function f(.) where weights w ={$w_1$,$w_2$,$w_3$,...} combined to produce the output. The outputs of each layer are computed as $Y_{kn}$ = f ($W_{n,m,k}$ · $X_m$ + b $_{i,k}$) where (k) is the number of layer, (d) is the number of inputs, (i) is the number of bias nodes, (n) is the number of neurons, (m) is the number of weight for each sending neuron and f (.) is the activation function (e.g., sigmoid and Tan sigmoid functions). There is no exact rule for determining the number of hidden layers and neurons in the hidden layer. No exact rule exists to determine the number of hidden layers and neurons in the hidden layer. (Huang and Huang, 1991; Choi et al., 2001) stated that a one hidden layer MLP needs at least (P − 1) hidden neurons to classify (P) patterns.

A three-layer network (input layer, hidden layer, and output layer) can solve a wide range of prediction, approximation and classification problems. Moreover, to avoid over-fitting problems and enhance the generalization capability, the number of training cases should be more than the size of the network (Rutkowski, 2005). Learning mechanism of ANNs is modifying the weights and biases of the network to minimize the in-sample error. The developing ANNs can be summarized in the following steps:

I. Collect and prepare the historical cases.
II. Divide the collected cases into a training set and a validation set.
III. Determine of relevant inputs and outputs.
IV. Select the number of hidden layers.
V. Select the number of neurons in each hidden layer.
VI. Selecting the transfer function.
VII. Set initial weights.
VIII. Select the learning algorithm to develop the ANNs' weights.
IX. Calculate the resulting error such as mean absolute percentage error (MAPE).

## 2.4.5 Evolutionary computing (EC)

Evolutionary computing (EC) is a natural selection inspired based on evolutionary theory (Darwin, 1859) such as Genetic algorithm (GA) (Holland, 1975). The genetic information can be represented as chromosomes where it is a powerful tool for optimization and search problems. Chromosome representation is the first design step in EC where chromosome is a possible candidate solution to search and an optimization problem. The gene is the functional unit of inheritance where any chromosome is expressed by a number of genes. A chromosome can be as a vector (C) consisting of (n) genes denoted by $(c_n)$ as follows: C = {$c_1$, $c_2$, $c_3$… $c_n$}. Each chromosome (C) represents a point in the n-dimensional search space.

The first task in chromosome construction is to encode the genetic information. Binary coding is the one of the most commonly used chromosome representation where $(b_i)$ is a binary values consists of zero or one as follows:

$$X = \{(b_1, b2\ldots b_l ), (b_1, b_2, \ldots , b_l ), \ldots\}, \quad bi \in \{0, 1\}$$

Fitness function $(F_{Ci})$ is a problem-dependent which guides the search model to converge and get the optimal solution. $F_{Ci}$ is applied to evaluate the fitness of each chromosome to select the best subset of chromosomes for crossover and mutation processes. As illustrated in example.1 where two chromosomes A and B are consisted of seven genes. Offspring 1 and Offspring 2 are produced by crossover of chromosome1 and chromosome 2 where the one point crossover is applied at the third gen of the chromosomes. The nest process is mutation where the sixth gene of the offspring 2 is mutated to 1 value.

Example 1:

Crossover process:

Chromosome A      **101**1001
Chromosome B      111**1111**

|            |          |
|------------|----------|
| Offspring 1 | 1011<u>111</u> |
| Offspring 2 | <u>111</u>1001 |

Mutation process:

|            |          |
|------------|----------|
| Offspring 1 | 1011111 |
| Offspring 2 | 11110**1**1 |

Relative fitness is the fitness function for each chromosome where relative fitness is criterion to select the next generation of chromosomes. Genetic operators are selection process of the fittest chromosomes and then conducting crossover and mutation processes subsequently. Many selection approaches exist such as random selection, proportional selection (roulette wheel selection), tournament selection and rank-based selection. Five main steps are required to develop an optimization problem by EC:

(i) Chromosome representation.

(ii) An initial population representation.

(iii) Definition of the fitness function as a chromosome selection criterion.

(v) EA parameter values determination such as probabilities of genetic operator's population size, and the maximum number of generations.

## 2.4.6 Case based reasoning (CBR)

Case-based reasoning (CBR) is a sustained learning and incremental approach that solves problems by searching the most similar past case and reusing it for the new problem situation (Aamodt and Plaza, 1994). Therefore, CBR mimics a human problem solving (Ross, 1989; Kolodner, 1992). As illustrated in Fig.2.7, CBR is a cyclic process learning from past cases to solve a new case. The main processes of CBR are retrieving, reusing, revising and retaining. Retrieving process is solving a new case by retrieving the past cases. The case can be defined by key attributes. Such attributes are used to retrieve the most similar case, whereas, reuse process is utilizing the new case information to solve the problem. Revise process is evaluating the suggested solution for the problem. Finally, retain process is to update the stored past cases with such new case by incorporating the new case to the existing case-base (Aamodt and Plaza, 1994).

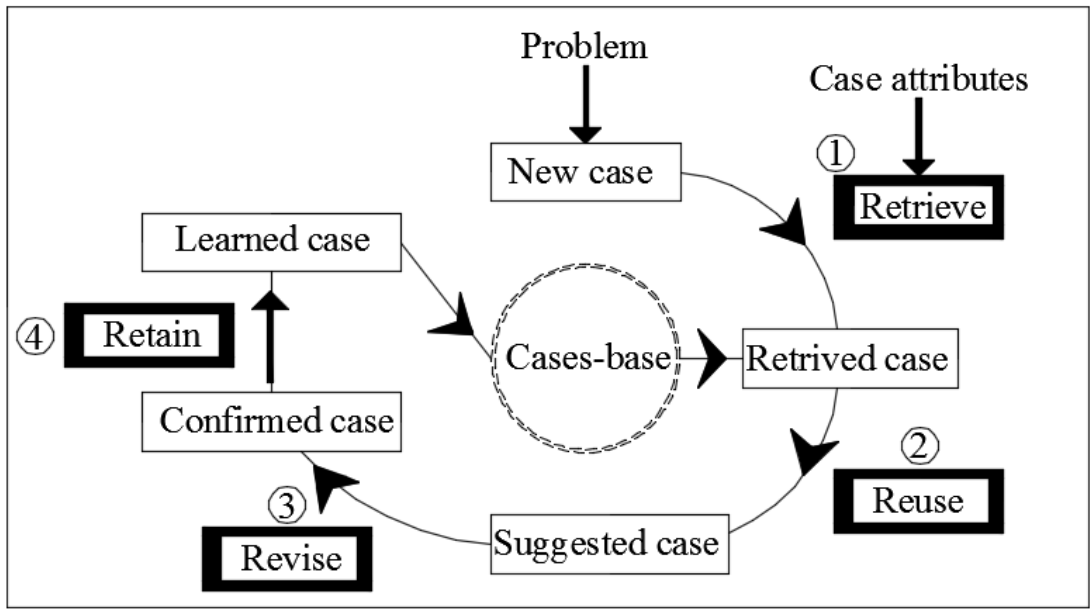

**Fig. 2.7 CBR processes (Aamodt and Plaza, 1994).**

The advantage of CBR is dealing with a vast amount of data where all past cases and new cases are stored in database techniques (Kim & Kang, 2004). The developing CBR can be summarized in the following steps:

  I.   Collect and prepare the historical cases.
  II.   Divide the collected cases into a training set and a validation set.
  III.  Determine of relevant inputs attributes and outputs.
  IV.  Identify the similarity function and conduct CBR processes.
  V.   Calculate the resulting error such as mean absolute percentage error (MAPE).

**2.4.7 Hybrid intelligent system**

The Fusion of this CI methodologies is called a hybrid intelligent system where Zadeh (1994) has predicted that the hybrid intelligent systems will be the way of the future. FL is an approximate reasoning technique. However, it does not have any adaptive capacity or learning ability. On the other hand, ANNs is an efficient mechanism in learning from given data and on uncertainty nature exist. EC enables optimization structure to the developed system. Combining these methodologies can enhance the computational model where the limitations of any single method can be compensate by other methods (Siddique and Adeli, 2013). Fig 2.8 illustrates a fusion of three basic model: FL, ANNs, and EC. Based on such models, many hybrid models can be evolved such as neuro-fuzzy model, evolutionary neural networks.

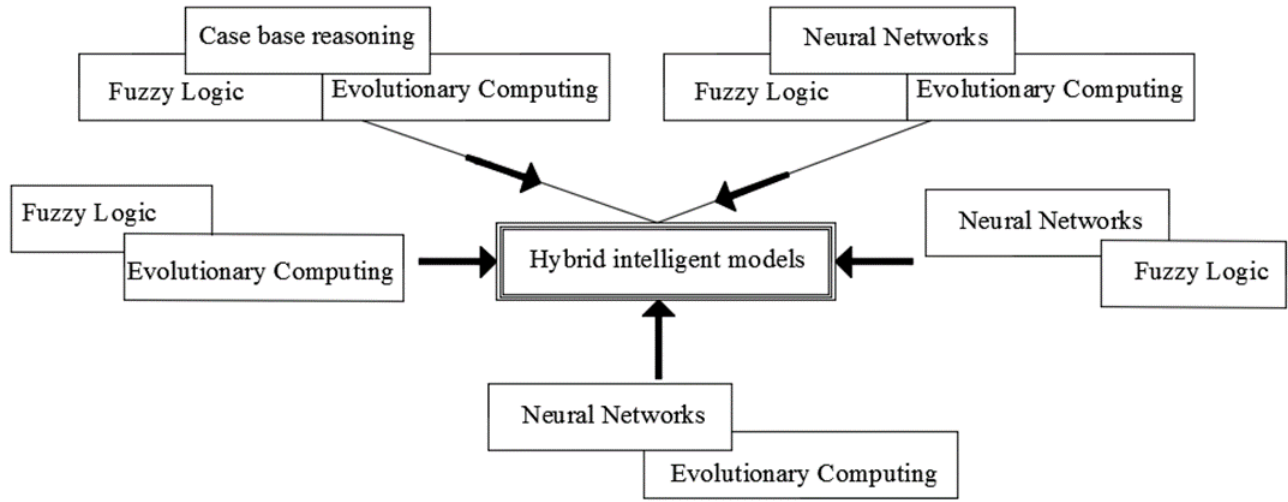

**Fig. 2.8 Hybrid intelligent systems (Siddique and Adeli, 2013).**

### 2.4.8 Data transformation

The objective of data transformation is to address the normality assumption of data distribution where the probability distribution shape has an important role in statistical modeling to convert error terms for linear models (Tabachnick and Fidell, 2007). Data transformation may produce more accurate results. (Stoy et al, 2008, 2012) have developed a semilog model to predict the cost of residential construction where the MAPE for the semilog model (9.6%) was better than linear regression model (9.7%). The previous result proved that semilog models may produce a more accurate model than a plain regression model. However, this is not a rule, in other words, plain regression models may produce more accurate and simple models than transformed models.

Lowe et al, (2006) have established a predictive model based on based on 286 historical cases where three alternatives: cost/m2, the log of cost variable, and the log of cost/m2 have been developed instead of raw cost data model where such data transformation approach has more accurate results than untransformed data model. Love et al, (2005) have represented the project time–cost relationship by a logarithmic regression model. (Wheaton and Simonton, 2007) have performed a semilog regression model to assess a building cost index. Thus, transformation of raw data can help to produce more reliable cost model.

### 2.4.9 Cost modeling review

The objective is to provide an overview of the recent and future trends in construction cost model development. The study has reviewed the past practices of parametric cost estimate at the conceptual stage for the construction project. Recently, many international journals have been reviewed such as the journal of construction engineering and management, journal of computing in civil engineering, and automation in construction, construction management, and economics. Such journals represent the most common and high-ranked journals for construction cost modeling.

The study focuses on the survey the most common techniques used to build a reliable parametric cost model based the collected data. Many machine learning (ML) and statistical techniques have been conducted such as regression model, ANNs, CBR and SVM. Moreover, hybrid models and fuzzy models have been reviewed to provide an overall perspective of the cost models developments as shown in Table.2.4.

**Table. 2.4. The review of the past practices of cost model development.**

| Method / technique | Project / Purpose | Findings and characteristics | Reference |
|---|---|---|---|
| ANNs | mark up estimation | In 1992, This study showed that ANNs produces better performance than hierarchical model for markup estimation. Moreover, GA is used for optimizing ANNs weights for markup estimation. Such model has displayed good generalization results. | Moselhi and Hegazy, 1993 |
| ANNs | highway construction | In 1998, This study conducted a regularization neural networks model that produced better predictable and reliable model for Highway construction projects. | Adeli and Wu, 1998 |
| ANNs | building projects | Based on 300 examples, a three cost models consisted of five, nine and 15 input parameters have been developed to predict the cost per m2 and the log of cost per $m^2$ in United Kingdom, in 2002.cost per $m^2$ model produces higher $R^2$. Whereas, the log model produces lower values of MAPE. For the selected model, $R^2$ value is 0.789 and a MAPE is 16.6% | Emsley et al, 2002 |
| ANNs | structural systems of residential buildings | Based on 30 examples, a cost model consisted of eight design parameters is developed to predict the cost per $m^2$ of reinforced concrete for 4–8 stories residential buildings in Turkey, in 2004. the cost estimation accuracy is 93% | Günaydın and Doğan, 2004 |
| ANNs | highway construction | In 2005, a NN model is built for Highway Construction Costs where the index of highway construction cost reflects the change in overall cost over time. | Wilmot and Mei, 2005 |

**Continue: Table. 2.4. The review of the past practices of cost model development.**

| Method / technique | Project / Purpose | Findings and characteristics | Reference |
|---|---|---|---|
| ANNs | building projects | Based on 286 past cases of data collected in the United Kingdom, a linear regression models and ANNs model have been established to assess the cost of buildings. Three alternatives: cost/m2, the log of cost variable, and the log of cost/m2 have been conducted instead of raw cost data where such data transformation approaches have better results than untransformed data model. A total of six models have been developed based on forward and backward stepwise regression analyses. The best regression model was the log of cost backward model. | Lowe et al, 2006 |
| ANNs | highway | Based on 67 examples, a reliable NN cost model has been developed for highway projects where MAPE is 1.55% | Salem et al, 2008 |
| ANNs | site overhead cost estimating | Based on 52 examples, a cost model consisted of ten input parameters is developed to predict the site overhead cost in Egypt in 2011. | ElSawy, et al, 2011 |
| ANNs | building projects | Based on 169 examples, a NN cost model has been developed for building construction projects with acceptable prediction error. | El-Sawalhi and Shehatto, 2014 |
| ANNs | highway construction | A prediction model has been developed with a MAPE of 1.4% for the unit cost of the highway project in Libya by changing ANNs structure, training functions and training algorithms until optimum model reached. | Elbeltagi et al, 2014 |

**Continue:** **Table. 2.4. The review of the past practices of cost model development.**

| Method / technique | Project / Purpose | Findings and characteristics | Reference |
|---|---|---|---|
| ANNs | public construction projects | Based on 232 public construction projects in Turkey, a multilayer perceptron (MLP) model and Radial basis function (RBF) model are developed to estimate for construction cost. RBF shows superior performance than MLP with approximately 0.7 %. | Bayram et al, 2015 |
| ANNs | building projects | Based on 657 building projects in Germany, a multistep ahead approach is conducted to increase the accuracy of model's prediction. | Dursun and Stoy, 2016 |
| ANNs | water treatment plants costs | First, cost drivers that influence construction costs of water treatment plants have been identified. Cost drivers have been determined through Descriptive Statistics Ranking (DSR) and Exploratory Factor Analysis. Principal component analysis (PCA) with VARIMAX rotation through five iterations have been used to minimize multicollinearity problem. Kaiser Criterion can be used so that a total of 33 variables were reduced to eight components while using Cattell's Scree test has reduced variables to four components. | Marzouk and Elkadi, 2016 |
| ANNs and regression | highway projects | Radial basis neural networks and regression model are developed for completed project cost estimation. The regression model produces better performance than ANNS model. Moreover, a hybrid model is developed and produces reliable results. A natural log transformation has helped to improve the linear relationship between variables. | Williams, 2002 |

| Method / technique | Project / Purpose | Findings and characteristics | Reference |
|---|---|---|---|
| ANNs and regression | cost deviation | Based on 41 examples, this study compared an ANNs model with regression model for cost deviation in reconstruction projects, in 2003. | Attalla and hegazy, 2003 |
| ANNs and regression | tunnel construction | Based on 33 constructed tunnels, both ANNs and regression models have been developed for Tunnel construction where the developed models are fitted for their purpose and reliable for cost prediction. | Petroutsat-ou et al, 2012 |
| ANNs and Regression | structural steel buildings | Based on 35 examples, a cost model consisted of three input parameters is developed to predict the preliminary cost of structural steel buildings in 2014. ANNs produces better performance than regression models where ANNs model has improved the MAPE by approximately 4 % than regression model. | El-Sawah and Moselhi, 2014 |
| CBR | building projects | This study has incorporated the decision tree into CBR to identify attribute weights of CBR. Such approach shows more reliable results for residential building projects cost assessment. | Doğan et al,2008 |
| CBR | pavement maintenance | Based on library of past cases, This study has developed a CBR model for pavement maintenance operations costs based on computing case similarity. | Chou ,2009 |
| CBR | pump stations | A parametric cost model is presented where a questionnaire survey has organized to analyze the most critical factors affecting the final cost of pump stations. Using Likert scale, these factors are screened to determine the key factors. A case-based reasoning has been built and tested to develop the proposed model. | Marzouk and Ahmed, 2011 |

**Table. 2.4. The review of the past practices of cost model development.**

| Method / technique | Project / Purpose | Findings and characteristics | Reference |
|---|---|---|---|
| CBR | military barrack projects | based on 129 military barrack projects, CBR model has been developed for cost estimation where the model produces reliable results | Ji et al, 2012 |
| CBR | building projects | This study has introduced an improved CBR model based on multiple regression analysis (MRA) technique where MRA has been applied in the revision stage of CBR model. Such model significantly improves the prediction accuracy where the performance of business facilities model has improved by 17.23%. | Jin et al, 2012 |
| CBR | storage structures | This study has built a CBR model to estimate resources and quantities in construction projects. The nearest neighbor technique has been conducted to measure the retrieval phase similarity of CBR model. The model has showed reliable MAPE ranging from 8.16% to 28.40%. | De soto and Adey, 2015 |
| CBR and GA | bridges | A cost estimation model has been developed based on CBR and GA for bridge projects which is used for optimizing parameters of CBR. Such methodology improves the accuracy than conventional cost model. | K. J. Kim and K. Kim, 2010 |
| CBR and AHP | highway | Analytic hierarchy process (AHP) has incorporated into CBR to build a reliable cost estimation model for highway projects in South Korea. | Kim, 2013 |
| Evolution--ary NN | highway | Based on 18 examples, a reliable NN cost model has been developed based on optimizing NN weights for highway projects. Simplex optimization of neural network weights is more accurate than trial and error and GA optimization where MAPE was 1%. | Hegazy and Ayed, 1998 |

**Continue:**  **Table. 2.4. The review of the past practices of cost model development.**

| Method / technique | Project / Purpose | Findings and characteristics | Reference |
|---|---|---|---|
| Evolutiona--ry NN | residential buildings | Based on 498 cases, a reliable NN cost model has been developed based on optimizing NN weights for residential buildings. GA optimization of NN parameters is more accurate than trial and error model where MAPE was 4.63%. | Kim et al, 2005 |
| Evolution ary Fuzzy Neural Inference Model (EFNIM), | building projects | This study has incorporate computation intelligence models such as ANNs, FL and EA to make a hybrid model which improves the prediction accuracy in complex project. As a result, an Evolutionary Fuzzy Neural model has been developed for conceptual cost estimation for building projects with reliable accuracy | Cheng et al, 2009 |
| Evolution ary fuzzy hybrid neural network | building projects | An Evolutionary Fuzzy Hybrid Neural Network model has been developed for conceptual cost estimation. FL is used for fuzzification and defuzzification for inputs and outputs respectively. GA is used for optimizing the parameters of such model such as NN layer connections and FL memberships. | Cheng et al, 2010 |
| Evolution ary fuzzy and SVM | building projects | Hybrid AI system based on SVM, FL and GA has been built for decision making for project construction management. The system has used FL to handle uncertainty to the system, SVM to map fuzzy inputs and outputs and GA to optimize the FL and SVM parameters. The objective of such system is to produce accurate results with less human interventions where MF shapes and distributions can be automatically mapped. | Cheng and Roy, 2010 |

**Table. 2.4. The review of the past practices of cost model development.**

| Method / technique | Project / Purpose | Findings and characteristics | Reference |
|---|---|---|---|
| GA for ANNs | residential buildings | This study has built three cost NN models by back propagation (BP) algorithm, GA for optimizing NN weights and GA for parameters optimization of BP algorithm. Optimizing parameters of BP algorithm produces the best results. | Kim et al, 2005 |
| GA for ANNs and CBR | bridge construction projects | GA is used as optimizing tool for ANNs and CBR cost models where such two model have been developed for bridge projects in Taiwan. Both models have displayed reliable results. | Chou et al, 2015 |
| Fuzzy linear regression | wastewater treatment plants | Based on 48 wastewater treatment plants, a fuzzy logic model is developed with acceptable error and uncertainty considerations. | Chen and Chang, 2002 |
| Fuzzy logic | design cost overruns on building projects | Based on the collected building projects in 2002, a fuzzy logic model is developed for estimating design cost overruns on building projects with acceptable error and uncertainty considerations. | Knight and Fayek, 2002 |
| Fuzzy sets | cost range estimating | This study proposed the use of fuzzy numbers for cost range estimating and claimed the fuzzy numbers for fuzzy scheduling range assessment. | Shaheen et al, 2007 |
| Fuzzy model | wastewater treatment projects | This study has compared Linear Regression model with Fuzzy Linear Regression model for wastewater treatment plants in Greece. The results of both models are similar and reliable. | Papadopoulos et al, 2007 |
| Fuzzy model | building projects | A fuzzy model is built based on four inputs and one output where a set of IF-then rules, center of gravity fuzzufucation, the product inference engine and singleton fuzzifier are applied. 3.2% is the maximal error. | Yang and Xu, 2010 |

 **Table. 2.4. The review of the past practices of cost model development.**

| Method / technique | Project / Purpose | Findings and characteristics | Reference |
|---|---|---|---|
| Fuzzy model | building projects | This study has applied index values for membership degree and exponential smoothing method to develop a construction cost model. | Shi et al, 2010 |
| Fuzzy neural network | cost estimation | Evolutionary fuzzy neural network model has developed for cost estimation based on 18 examples and 2 examples for training and testing respectively. GA is used to avoid sinking in local minimum results. | Zhu et al, 2010 |
| Fuzzy logic | building projects | Based on 106 building projects in Gaza trip in 2012, a fuzzy logic model is developed for building projects with acceptable error and good generalization. | El Sawalhi, 2012 |
| Fuzzy model | cost prediction | An improved fuzzy system is established based on fuzzy c-means (FCM) to solve the problem of fuzzy rules generation. Such model has produced better results for scientific cost prediction. | Zhai et al, 2012 |
| Fuzzy logic and neural networks | construction materials prices | This study has developed an ANNs model for predicting construction materials prices where as a fuzzy logic model is applied to determine the degree of importance of each material to use for ANNs model. Such modelling has acceptable accuracy in training and testing phases. | Marzouk and Amin, 2013 |

**Continue:** **Table. 2.4. The review of the past practices of cost model development.**

| Method / technique | Project / Purpose | Findings and characteristics | Reference |
|---|---|---|---|
| Fuzzy subtractive clustering | Telecomm uni-cation towers | Based on 568 cases, a four inputs fuzzy clustering model and sensitivity analysis are conducted for estimating telecommunication towers construction cost with acceptable MAPE. | Marzouk and Alaraby, 2014 |
| Fuzzy logic | satellite cost estimation | Based on two input parameters, a fuzzy logic model is developed for Satellite cost estimation. Such models works as Fuzzy Expert Tool for satellite cost prediction. | Karatas and Ince, 2016 |
| regression analysis, NN and CBR | Building projects | Based on 530 examples, three cost models consisted of nine parameters is developed to predict the cost buildings in Korea in 2004. NN model produces better results than CBR and regression model. However, CBR produces better results than NN as long term use due to updating cases to CBR system. | Kim et al, 2004 |
| Neuro-Genetic | residential buildings | Based on 530 cases of residential buildings, Neuro-Fuzzy cost estimation model has built where GA is applied for optimizing BP algorithm parameters. Such approach has more accurate results than trial and error BP algorithm. | Kim et al, 2005 |
| Neuro-fuzzy | residential constructi on projects | This study has developed adaptive neuro-fuzzy model for cost estimation for residential construction projects. Such model is integration of ratio estimation method with the adaptive neuro-fuzzy to obtain mining assessment knowledge that is not available in traditional approaches. | Yu and Skibniew ski, 2010 |

| Method / technique | Project / Purpose | Findings and characteristics | Reference |
|---|---|---|---|
| Neuro-Fuzzy and GA | semiconductor hookup construction | Based on 54 case studies of Semiconductor hookup construction, Neuro-Fuzzy cost estimation model has built and optimized by GA. Such model has accuracy better than the conventional cost method by approximately 20%. | Hsiao et al, 2012 |
| Neuro-fuzzy | water infrastructure | Based on 98 examples, a combination of neural networks and fuzzy set theory is incorporated to develop more accurate precise model for water infrastructure projects where MAPE is 0.8%. | Ahiaga et al, 2013 |
| Neuro-fuzzy | water infrastructure projects | Based on 1600 water infrastructure projects in UK, Neuro-Fuzzy hybrid cost model has been built where max-product composition produces better results than the max-min composition. | Tokede et al, 2014 |
| Regression | Building projects | A logarithmic regression model has been developed to examine the project time–cost relationship. Projects in various Australian states have performed a transformed regression model (semilog) to estimate a building cost index based on historical construction projects in several markets (Wheaton and Simonton, 2007). | Love et al, 2005 |
| Regression | Building projects | A semilog model has used to predict the cost of residential construction projects where the MAPE for the semilog model (9.6%) is more than linear regression model (9.7%). The previous result proved that semilog models may produce a more accurate model that plain regression model. However, this is not a rule, in other words, plain regression models may produce more accurate and simple models than transformed models. | Stoy et al, 2008 |

| | | **Continue:** Table. 2.4. The review of the past practices of cost model development. | |
|---|---|---|---|
| **Method / technique** | **Project / Purpose** | **Findings and characteristics** | **Reference** |
| Regression | Building projects | A semilog regression model has performed to develop cost models for residential building projects in German. The most significant variables have been identified by backward regression method. For the selected population, the proposed model has prediction accuracy of 7.55%. | Stoy et al, 2012 |
| SVM | building construction project | Based on 62 cases of building construction project in Korea, a SVM model has been developed to evaluate conceptual cost estimate. Such model can help clients to know the quality and accuracy of cost prediction. | An et al, 2007 |
| SVM | Building projects | This study has utilized the theory of the Rough Set (RS) with SVM to improve the prediction accuracy. RS is used for attributes reduction. | Wei, 2009 |
| SVM and ANNs | Building projects | Based on 92 building projects, ANNs and SVM have been used to predicted cost and schedule success at conceptual stage. Such model has prediction accuracy of 92% and 80% for cost success and schedule success, respectively. | Wang et al, 2012 |
| SVM | commercial building projects | Based on 84 cases of commercial building projects, a principal component analysis method has been developed into SVM to predict cost estimate based on project parameters. | Son et al, 2012 |

**Table. 2.4. The review of the past practices of cost model development.**

| Method / technique | Project / Purpose | Findings and characteristics | Reference |
|---|---|---|---|
| SVM | Building projects | This study has incorporated Least Squares Support Vector Machine (LS-SVM), Differential Evolution (DE) and machine learning based interval estimation (MLIE) for Interval estimation of construction cost. DE is used for optimizing cross-validation process to avoid over-fitting. | Cheng and Hoang, 2013 |
| SVM | Building projects | Based on 122 historical cases, hybrid intelligence model has been developed for construction cost index estimation with 1% MAPE. Such model consists of Least Squares Support Vector Machine (LS-SVM) and Differential Evolution (DE) that DE is applied to optimize LS-SVM tuning parameters. | Cheng et al, 2013 |
| SVM | Building projects | This study has developed a hybrid cost prediction model for building based on machine learning based interval, Least Squares Support Vector Machine (LS-SVM), and estimation (MLIE), and Differential Evolution (DE). | Cheng and Hoang, 2014 |
| SVM | predicting bidding price | Based on fifty four tenders, a SVM model has been developed with 2.5% MAPE for bidding price prediction. | Petruseva et al, 2016 |

## 2.4.10 Discussion of review

Based on the reviewed studies as shown Table 2.4, the survey has been classified into six main categories to represent models used for cost model development. These categories are ANNs model, FL model, regression model, SVM model, CBR model and hybrid models where hybrid models represents all combined methods such as fuzzy neural network and evolutionary fuzzy hybrid neural network .. etc.

Table 2.4 has reviewed a total of 52 studies about parametric cost modeling. As shown in Fig. 2.9 (A), the percentages of the categories, based on the reviewed 52 studies, are 27%, 25%, 14%, 13%, 11% and 10% for hybrid models, ANNs, fuzzy models, regression, SVM, and CBR, respectively. These percentages indicates that hybrid models are the current trend in parametric cost estimate modeling where the researches use such hybrid models to enhance the performance of the developed model and accuracy of the prediction results. In addition, hybrid models avoid limitations of a single method. For example, hybrid model of ANNs and FL produces a neuro-fuzzy model that provide uncertainty for ANNs. On the other hand ANNs provide learning ability to the fuzzy system.

The second percent of 25% represents the use of ANNs where it is a powerful ML technique to represent nonlinear data. The third percent is 14% that represent fuzzy models. Fuzzy model should be widely conducted since the fuzzy model provides vagueness and uncertainty to the results and produces more reliable prediction to the future real world cases. The fourth percentage is 13% that represents the regression model. Generally, regression model has been widely conducted due to its simplicity .However, ANNs can provide better results that regression model specifically with nonlinear data. SVM and CBR have similar small percent. However, CBR represents a promising technique where a CBR works as an incremental search engine for similar cases.

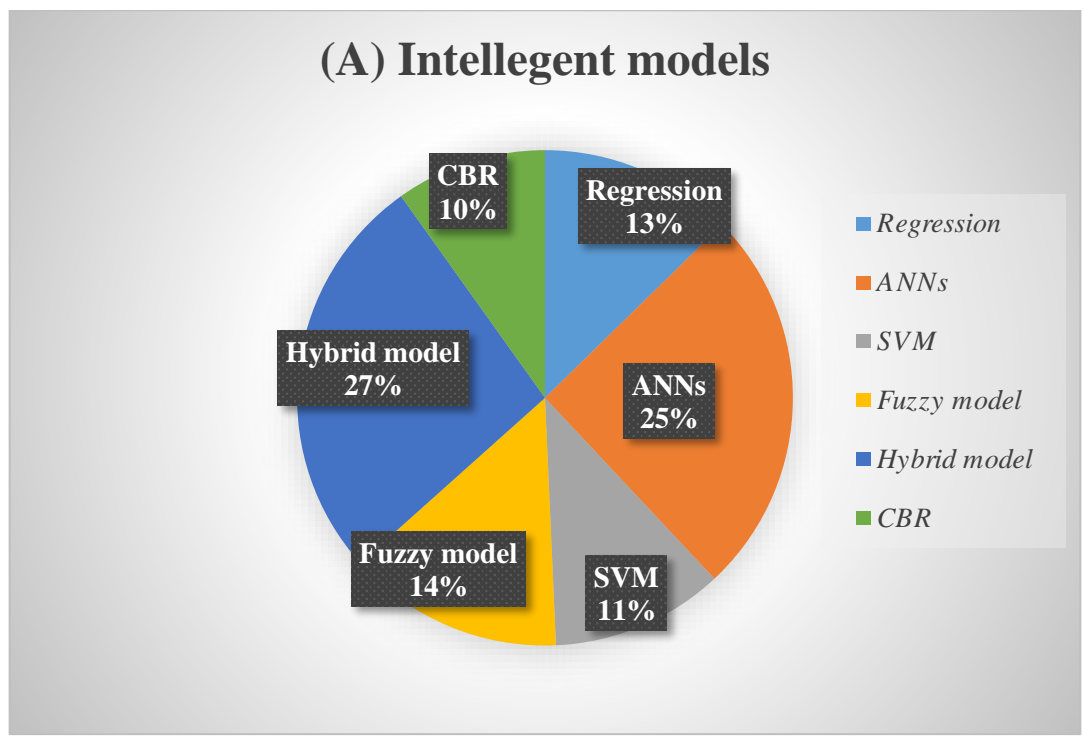

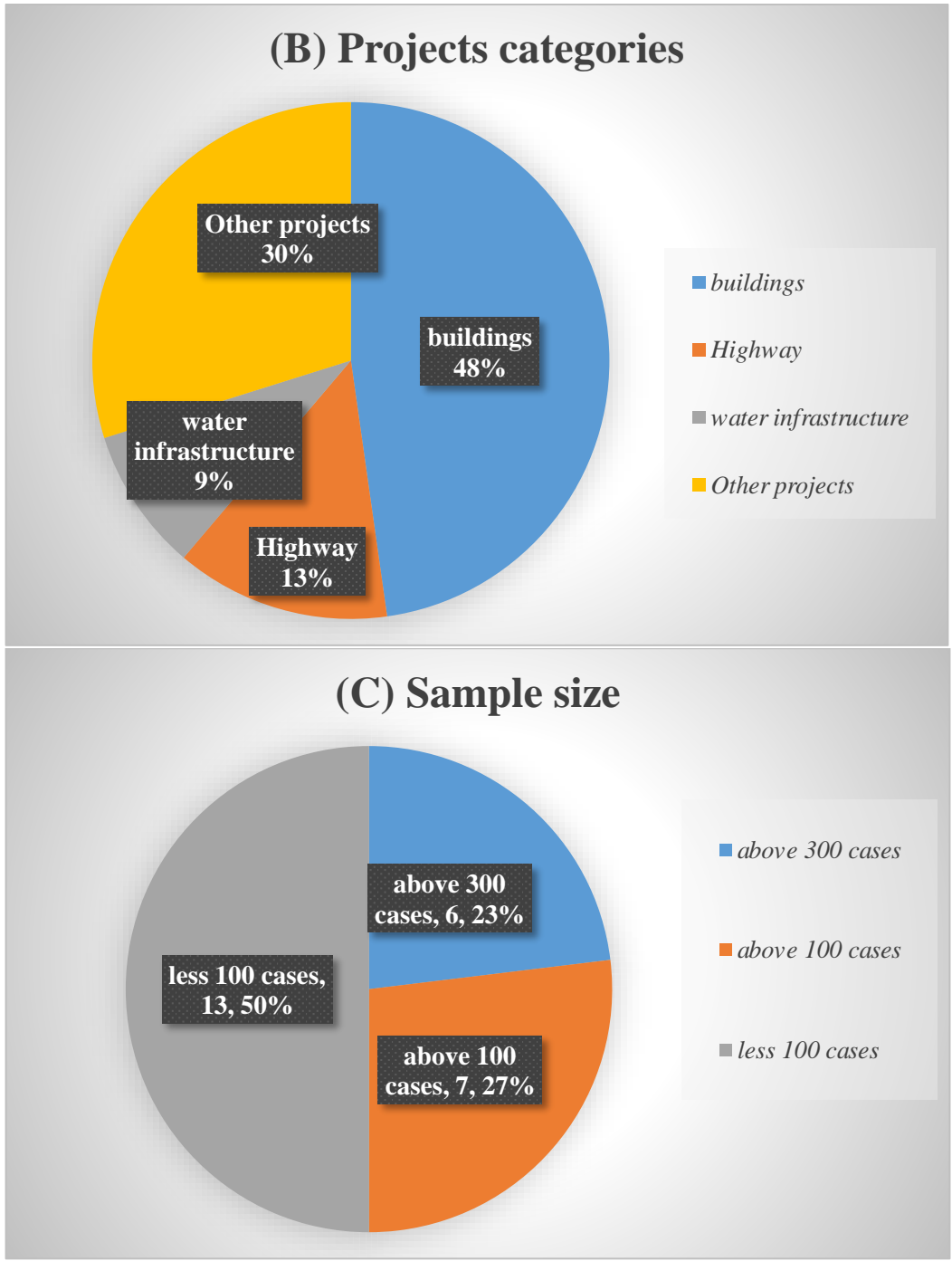

**Fig. 2.9 Classification of the previous study by (A) intelligent model, (B) project categories, and (C) sample size.**

Based on the reviewed studies shown in Table.2.4, the survey of the reviewed studies has been classified into four main categories to represent the projects used for the cost estimate. These categories are buildings, highway, water infrastructure and other projects. Building category includes residential building, industrial building, and commercial building projects. Whereas, highway category includes highway, road, pavement maintenance and bridges construction project. Water infrastructure includes waste water treatment and water infrastructure projects. Other projects include tunnel projects, steel projects, and telecommunication towers…etc.

As shown in Fig 2.9 (B), building category represents 48% of all collected projects whereas other projects category represents 30%, highway category represents 13% of all collected projects, whereas water infrastructure category represents 9%. Subsequently, building projects and highway projects have the greatest share in researchers' interest, whereas the other projects have fewer research efforts.

Based on the surveyed studies, the collected sample sizes range from 18 cases to 1600 cases. The sample size can be categorized into three categories: less than 100 cases, above 100 cases and above 300 cases. As shown in Fig.2.9 (C), about 50% of cases are less than 100 cases, whereas the cases above 100 and 300 have 27% and 23%, respectively. Most of the sample sizes are less than 100 cases and that may reflect a bias model and less model ability for generalization.

# CHAPTER 3

# RESEARCH METHODOLOGY

## 3.1 Introduction

This chapter discusses the methodology conducted in this study. The developed methodology to accomplish this study used both qualitative approaches and quantitative approaches to identify cost drivers affecting the cost of FCIPS. In addition, historical data analysis was used as the base of providing a relation between the main factors affecting the cost of the FCIPs to make estimates for new projects. This chapter provides the information about the steps followed to obtain the research objectives.

## 3.2 Research Design

The current research has two main objectives. The first objective is to identify key cost drivers affecting on cost of FCIPs where the second objective is to develop a precise parametric cost model to predict conceptual cost of FCIPs. As shown in Fig.3.1, the research methodology consists of two main processes. The first process is to identify key cost drivers based on quantitative and qualitative approaches. The second process is to develop a parametric cost model. These process will be discussed in details in the following sections.

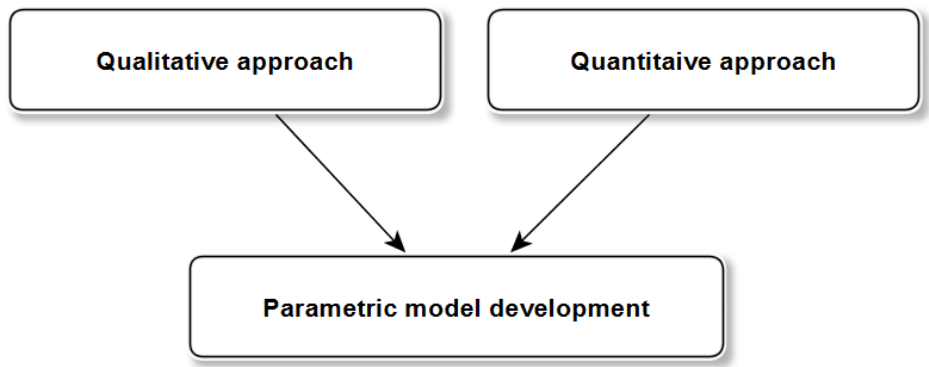

**Fig. 3.1. Research Methodology.**

## 3.3 Qualitative approach for cost drivers' selection

The objective of the current process is to determine the cost drivers of the FCIPs where these cost drivers can be used to develop a cost prediction model. This process has been designed to use two procedures to determine and evaluate the key cost drivers of FCIPs. The first procedure consists of TDM and Likert scale as traditional methods. The second procedure consists of FDM and FAHP as advanced methods. Accordingly, the cost drivers will be identified where that is the major objective of this study. Subsequently, results from both procedures can be compared to evaluate these two methods where Fig.3.2 illustrated the general idea of this methodology.

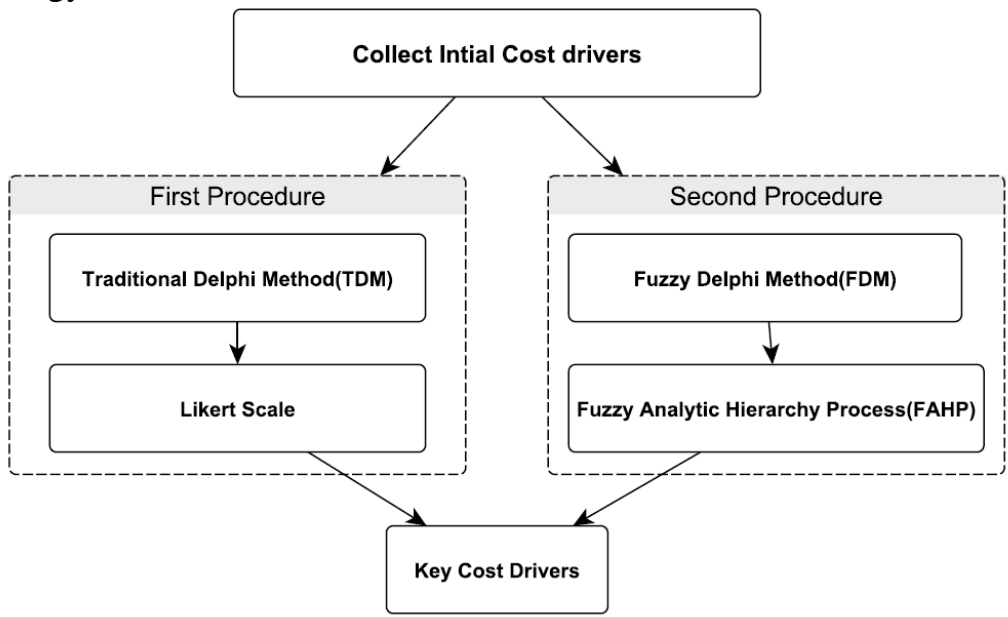

**Fig. 3.2. Qualitative approach methodology.**

## 3.4 Quantitative approach for cost drivers' selection.

The objective of this process is to identify the effective predictors among the complete set of predictors. This could be achieved by deleting both irrelevant predictors (i.e. variables not affecting the dependent variable) and redundant predictors (i.e. variables not adding value to the outcome). Therefore, key predictors selection methodology is based on a trinity of selection methodology:

1. Statistical tests (for example, F, chi-square and t-tests, and significance testing);
2. Statistical criteria (for example, R-squared, adjusted R-squared, Mallows' C p and MSE);
3. Statistical stopping rules (for example, P -values flags for variable entry / deletion / staying in a model) (Ratner, 2010).

As illustrated in Table 3.1, step 1 and 2 of the study methodology include literature survey to review previous practices for key parameters identification, collecting data of FCIPs, respectively. Step 3 is to conduct four statistical methods to analyze data and identify cost drivers. These methods are Exploratory Factor Analysis (EFA), Regression methods, correlation matrixes and hybrid methods. Finally, Step 4 compares results of statistical methods to select the best logic set of key cost drivers suitable for the conceptual stage of FCIPs. Fig. 3.3 showed the process of the cost driver's identification. Moreover, Fig.3.4 summarize all such methodology procedures. After collecting relevant data which represents all variables, statistical methods can be used to analysis data and screen such variables. All methods are developed using software SPSS 19 for Windows.

**Table. 3.1. Quantitative approach Methodology**

| Steps | Methods | Description |
|---|---|---|
| **Step 1** | | Literature survey |
| **Step 2** | | Data Collection |
| **Step 3** | | Apply statistical methods |
| | **Method 1** | **Exploratory Factor Analysis (EFA)** |
| | 1 | Review and Check sample size |
| | 2 | Scan correlation matrix, check multicollinearity and singularity |
| | 3 | KMO and  Bartlett's test |
| | 4 | Conduct PCA , select rotation method and criteria to retain variables |
| | **Method 2** | **Regression Methods** |
| | 1 | Forward Selection (FS) |
| | 2 | Backward Elimination (BE) |
| | 3 | Stepwise |
| | **Method 3** | **Correlation matrix scan** |
| | 1 | Pearson correlation |
| | 2 | Spearman correlation |
| | **Method 4** | **Hybrid models** |
| **Step 4** | | Compare results and suggest final key cost drivers |

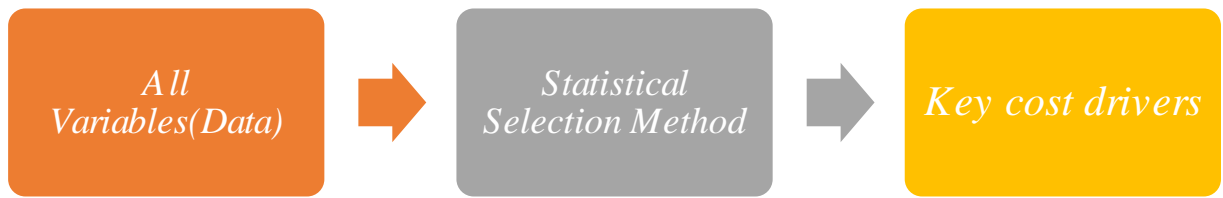

**Fig. 3.3. The process of the cost drivers' identification.**

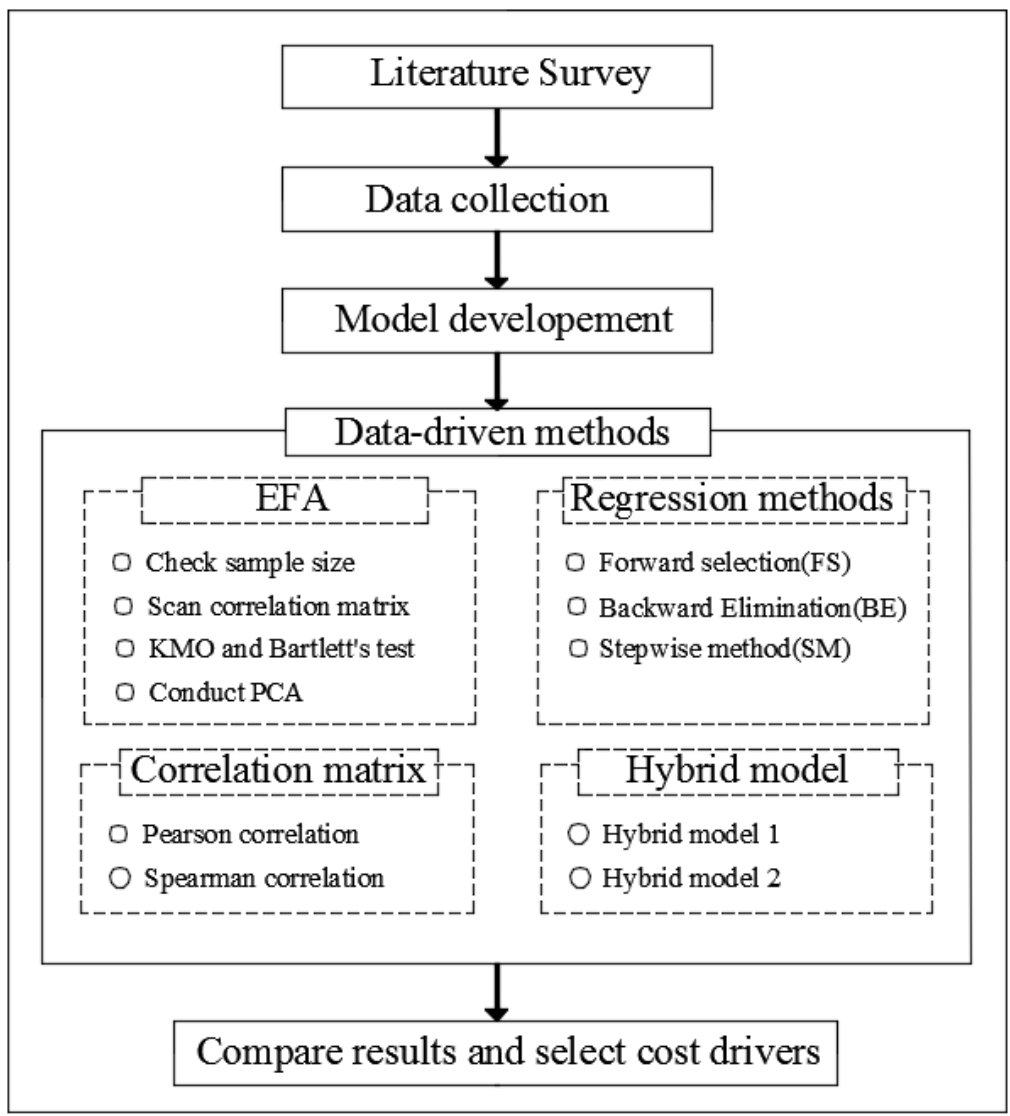

**Fig. 3.4. A research methodology for data-driven cost drivers' identification**

## 3.5 Model development

Once key cost drivers are identified, the parametric model can be developed. The purpose of this research is to develop a reliable cost estimating model to be used in early cost estimation. This research consists of six steps, the first step is to review the past literature. The second step includes data collection of real historical construction FCIPs whereas the third step includes a model development. The fourth step is to select the most accurate model based on the coefficient of determination ($R^2$) and Mean Absolute Percentage Error (MAPE). The fifth step focused on model validity by comparing the model results of 22 cases to compute MAPE for the selected models. The sixth step is to conduct sensitivity analysis to determine the contribution of each key parameter on the total cost of FCIPs. Fig.3.5 illustrates this process steps.

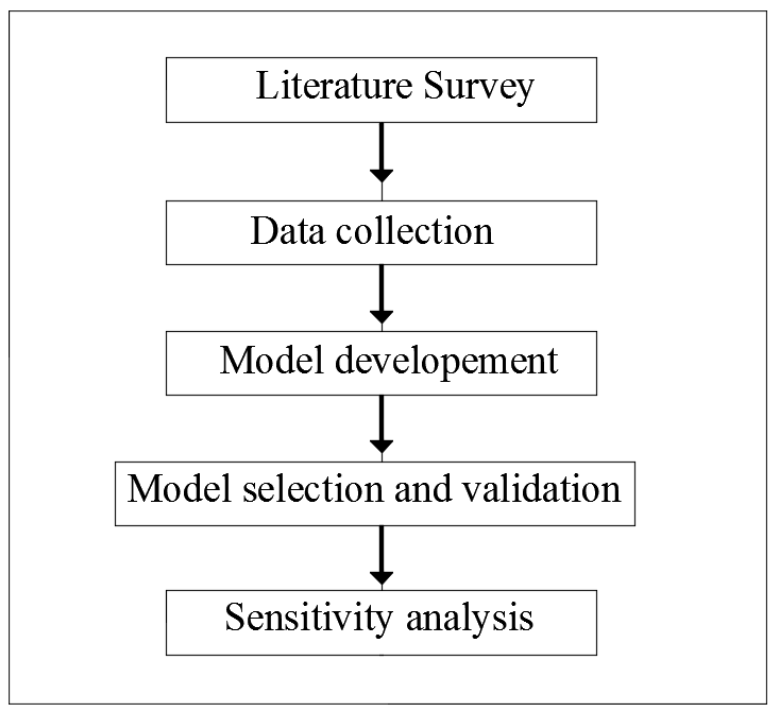

**Fig. 3.5. A research methodology for model development.**

# CHAPTER 4

# QUALITATIVE APPROACH

## 4.1 Introduction

The objective of the current study is to determine the cost drivers of the FCIPs where these cost drivers can be used to develop a cost prediction model. This study has been designed to use two procedures to determine and evaluate the key cost drivers of FCIPs. The first procedure consists of TDM and Likert scale as traditional methods. The second procedure consists of FDM and FAHP as advanced methods. Accordingly, the cost drivers will be identified where that is one of the major objectives of this study. Subsequently, results from both procedures can be compared to evaluate these two procedures and to select key cost drivers.

## 4.2 The first procedure: Traditional Delphi Method (TDM) and Likert scale

The first step, Appling Delphi rounds to detect all possible parameters and to rank the collected parameters. The second step, eliminate all parameters that were less than three on the 5-ponit Likert scale. Delphi technique was used to collect experts' opinions about parameters affecting the conceptual cost of the FCIPs. It provided feedback to experts in the form of distributions of their opinions and reasons. They were then asked to revise their opinions in light of the information contained in the feedback and to give reasons for their rating and selections. This sequence of questionnaire and revision was repeated until no other opinions detected (Hsu and Sandford, 2007).

The Delphi method has been achieved through the following three rounds. The first round included 15 exploratory interviews with experts. The participants were asked to present their opinions about parameters affecting the total construction cost of FCIPs. Specializations of the Interviewed personnel were illustrated in Fig .4.1. Interviewed personnel work in the ministry of water resources and irrigation, irrigation improvement sector and contractor companies to determine all expected parameters. As shown in Fig.4.1, a total of 23 % and 13 % of the respondents were consultant managers and contractor managers, respectively. The average years of experience of the respondents were about 10 years' in irrigation improvement projects and 15 years in civil engineering. This experience enhances the level of completeness, consistency and precision of the information provided.

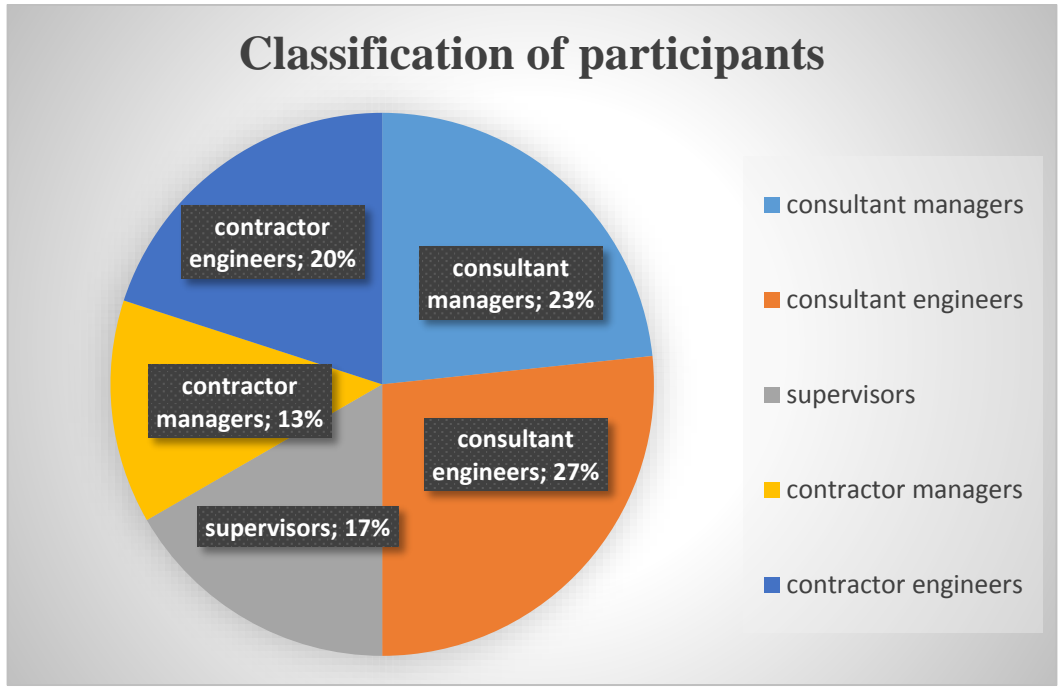

**Fig. 4.1. Classification of the participants.**

The second round, the factors that affect the cost of FCIPs which have been drawn from opinions of experts and previous studies, were presented to them to be ranked and evaluated. The third round, all collected factors have been revised and experts have been asked: "why have they been chosen these rates for each parameter?" .This round is important to emphases the factors rating and to ensure rating confidence. The participants were requested to indicate the degree of importance associated with each factor on the five point Likert scale of five categories ''Extremely Important'', ''Important'', "Moderately Important'', ''Unimportant'', and ''Extremely Unimportant''.

Based on the completed survey forms, a total number of 35 parameters affecting the conceptual cost of FCIPs have been collected. After completing the basic statistics that measure the frequency of responses (on the five- point Likert scale) for each of the 35 parameters, the values were used to develop common statistical indices such as the mean score (MS) using Equation (1.1) and the standard error (SE) Equation (1.2).

Where: Mean represents the impact of each parameter based on the respondents answers, (S) is a score set to each parameter by the respondents and it ranges from 1 to 5 where "1" is ''Extremely Unimportant'' parameter and "5" is "Extremely

Important" parameter, (f) is the frequency of responses to each rating for each impact of parameter, and (n) is total number of participants. The Field survey module is used in this survey showed in appendix (A: Field survey module) and the results shows in appendix (B: Delphi Rounds).

Table.4.1 listed these Parameters ($P_i$) along with their mean and SE. Based on collected data, SE is used for measuring the sufficiency of the sample size where the sample size is acceptable as long as SE does not exceed (0.2) (Marzouk and Ahmed, 2011). As a result of the previous Delphi rounds, all the 35 parameters were grouped into five categories. These categories were civil, mechanical, electrical, location and miscellaneous parameters. For parameters screening, the parameters whose mean value was calculated to be less than 3.0 were eliminated in order to keep the most important ones. Therefore, a total of 12 parameters were determined as the most important cost drivers of FCIPs. These parameters were illustrated in Table.4.1 and the twelve screened parameters are from P1 to P12.

### Table. 4.1 Parameters Affecting Construction Cost of FCIPs Projects.

| ID | categories | Parameters | Mean | SE |
|----|-----------|-----------|------|-----|
| P1 | Civil | Command Area (hectare) | 4.97 | 0.03 |
| P2 | Civil | PVC Length (m) | 4.43 | 0.09 |
| P3 | Civil | Construction year and inflation rate | 4.33 | 0.15 |
| P4 | Civil | Mesqa discharge ( capacity ) | 4.33 | 0.17 |
| P5 | Mechanical | Number of Irrigation Valves ( alfa-alfa valve ) | 4.20 | 0.15 |
| P6 | Civil | Consultant performance and errors in design | 4.13 | 0.18 |
| P7 | Electrical | Number of electrical pumps | 4.10 | 0.15 |
| P8 | Civil | PVC pipe diameter | 3.90 | 0.15 |
| P9 | Location | Orientation of mesqa ( intersecting with drains or roads or both) | 3.80 | 0.18 |
| P10 | Mechanical | Electrical and diesel pumps discharge | 3.57 | 0.13 |
| P11 | Civil | PC Intake , steel gate and  Pitching with cement mortar | 3.40 | 0.14 |
| P12 | Location | Type of mesqa (  Parallel to branch canal (Gannabya) , Perpendicular on branch canal) | 3.07 | 0.16 |
| P13 | miscellaneous | Farmers Objections | 2.93 | 0.20 |
| P14 | Electrical | Electrical consumption board type | 2.87 | 0.18 |
| P15 | Location | location of governorate (Al Sharqia  , Dakahlia , …) | 2.63 | 0.18 |
| P16 | Civil | Pump house size  3m*3m  or 3m*4m | 2.60 | 0.18 |
| P17 | miscellaneous | cement  price | 2.53 | 0.16 |
| P18 | Mechanical | Head of electrical and diesel pumps | 2.53 | 0.18 |
| P19 | miscellaneous | Farmers adjustments | 2.50 | 0.16 |
| P20 | Civil | Sand filling | 2.40 | 0.16 |
| P21 | Civil | Sump size | 2.37 | 0.17 |
| P22 | Civil | Contractor performance and bad construction works | 2.20 | 0.19 |
| P23 | miscellaneous | pump price | 2.13 | 0.20 |
| P24 | Civil | Crops on submerged soils ( Rice) and its season | 2.10 | 0.20 |
| P25 | miscellaneous | pipe price | 2.10 | 0.19 |
| P26 | Location | Topography and land levels of command area | 2.10 | 0.14 |
| P27 | Civil | Construction durations | 2.07 | 0.19 |
| P28 | Civil | Pumping and suction  pipes | 2.07 | 0.19 |
| P29 | Mechanical | Steel  mechanical connections | 2.00 | 0.17 |
| P30 | Civil | Difference between land and water levels | 1.90 | 0.14 |
| P31 | miscellaneous | steel price | 1.80 | 0.18 |
| P32 | Civil | Number of PVC branches | 1.80 | 0.17 |
| P33 | miscellaneous | Cash for damaged crops | 1.73 | 0.11 |
| P34 | Mechanical | Air / Pressure relief valve | 1.60 | 0.15 |
| P35 | miscellaneous | Crops on unsubmerged soils (wheat, corn, cotton, etc.) | 1.50 | 0.09 |

## 4.3 The second procedure

### 4.3.1 Fuzzy Delphi Technique (FDT)

Ishikawa et al. (1993) proposed Fuzzy Delphi Method where this method was extracted from the traditional Delphi method and fuzzy theory. The advantage of this method than traditional Delphi method is to solve the fuzziness of common understanding of experts' opinions. Moreover, the method takes into account the uncertainty among participants' opinions. Instead of gathering the experts' opinions as deterministic values, this method converts deterministic numbers to fuzzy numbers such as trapezoidal fuzzy number or Gaussian fuzzy number. Therefore, fuzzy theory can be applied to evaluate and rank these collected opinions. Accordingly, the efficiency and quality of questionnaires will be improved

(Liu, W.-K, 2013). The FDM steps were as following:

Step.1: Collect all possible parameters (similarly to TDM: Round 1).
Step.2: Collect evaluation score for each parameter.
Step.3: Set-up fuzzy number (Klir and Yuan, 1995).
Step.4: De-fuzzification (S).
Step.5: Setting a threshold (α).

The first step was collecting initial cost parameters similar to round one in TDM. The second step was evaluating each parameter by FDM conducted by five experts as illustrated in Table.4.2 and Fig.4.2.

**Table.4.2 the Fuzziness of linguistic terms for FDM for five-point Likert scale.**

| Linguistic terms | Fuzzy Delphi | | | TDM |
|---|---|---|---|---|
| | L | M | U | 0 |
| Extremely Unimportant | 0 | 0 | 0.25 | 1 |
| Unimportant | 0 | 0.25 | 0.5 | 2 |
| Moderately Important | 0.25 | 0.5 | 0.75 | 3 |
| Important | 0.5 | 0.75 | 1 | 4 |
| Extremely Important | 0.75 | 1 | 1 | 5 |

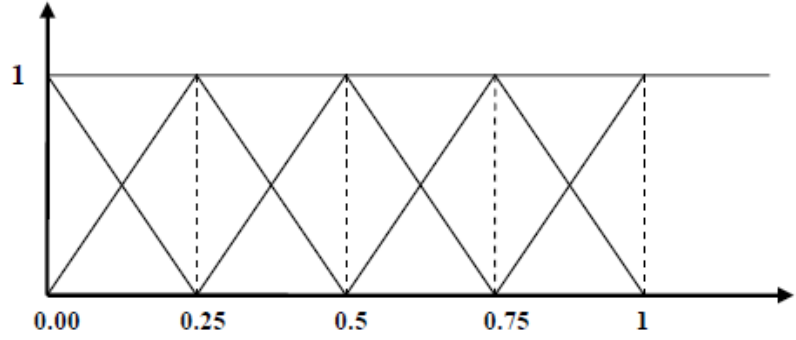

**Fig. 4.2. triangular fuzzy numbers for five-point Likert scale (Klir and Yuan, 1995).**

Third Step was that this study has used triangular fuzzy numbers to represent fuzziness of the experts' opinions where the minimum of the experts' common consensuses as point $l_{ij}$, and the maximum as point $u_{ij}$ as shown in Fig.4.3, and Equations (4.3,4.4,4.5,4.6) as follows (Klir and Yuan, 1995):

$$Lj = Min(Lij), i = 1,2, \dots \dots n \; ; j = 1,2, \dots m \qquad (4.3)$$

$$Mj = (\prod_{i=1,j=1}^{n,m} mij)^{1/n}, i = 1,2, \dots \dots n \; ; j = 1,2, \dots m \qquad (4.4)$$

$$Uj = Max(Uij), i = 1,2, \dots \dots n \; ; j = 1,2, \dots m \qquad (4.5)$$

$$(Wij) = (Lj, Ml, Uj) \qquad (4.6)$$

Where:

i: an individual expert.

j: the cost parameter for FCIPs.

$l_{ij}$: the minimum of the experts' common consensuses.

$m_{ij}$: the average of the experts' common consensuses.

$u_{ij}$: the maximum of the experts' common consensuses.

$L_j$: opinions mean of the minimum of the experts' common consensuses ($l_{ij}$).

$M_j$: opinions mean of the average of the experts' common consensuses ($M_{ij}$).

$U_j$: opinions mean of the maximum of the experts' common consensuses ($U_{ij}$).

$W_{ij}$: The fuzzy number of all experts' opinions.

n: the number of experts.

The fourth Step was using a simple center of gravity method to defuzzify the fuzzy weight $w_j$ of each parameter to develop value $S_j$ by Equation (4.7).

$$Sj = \frac{Lj + Mj + Uj}{3} \qquad (4.7)$$

Where: $S_j$ is the crisp number after de-fuzzification where is 0< S <1.

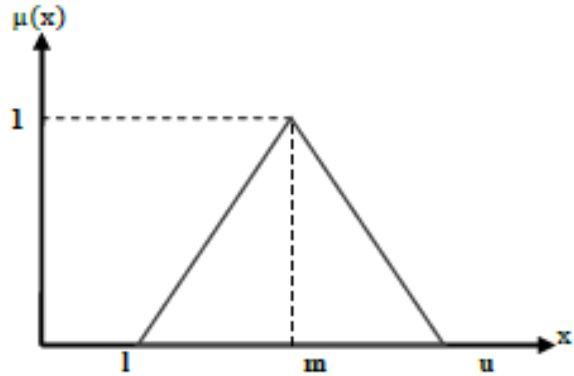

**Fig. 4.3. Triangular fuzzy number (Siddique and Adeli, 2013).**

Finally, the fifth step was that the experts provided a threshold to select or delete theses parameters. The threshold of FCIPs parameters was set as $\alpha = 0.6$.

If $S_j \geq \alpha$, then the parameter should be selected.

If $Sj < \alpha$, then the parameter should be deleted.

As shown in Table.4.3, each expert (j) presented his opinion as fuzzy numbers for each parameter $P_i$, and then all opinions of each parameter have been collected by the grasp equation to rank each parameter $P_i$. The initial 35 parameters as cost drivers of FCIPs were reduced to six parameters as illustrated in Table.4.4. However, $P_{22}$ has a crisp value of 0.62, this parameter has been deleted because it represents the construction cost from the contractor's point of view and this study aims to develop a prediction cost model from the MWRI's point of view. Moreover this parameters is not convenient and available at the conceptual stage of the FCIP.

**Table 4.3. The calculated results of the FDM.**

| Code | Opinions mean(L) | Opinions mean(M) | Opinions mean(U) | Crisp Value($S_j$) | Result |
|------|------------------|------------------|------------------|-------------------|--------|
| P1 | 0.75 | 1.00 | 1.00 | 0.92 | **Select** |
| P2 | 0.5 | 0.94 | 1.00 | 0.81 | **Select** |
| P3 | 0.5 | 0.84 | 1.00 | 0.78 | **Select** |
| P4 | 0.5 | 0.84 | 1.00 | 0.78 | **Select** |
| P5 | 0.25 | 0.59 | 1.00 | 0.61 | **Select** |
| P6 | 0 | 0.54 | 1.00 | 0.51 | Delete |
| P7 | 0.25 | 0.68 | 1.00 | 0.64 | **Select** |
| P8 | 0 | 0.33 | 0.75 | 0.36 | Delete |
| P9 | 0 | 0.00 | 0.75 | 0.25 | Delete |
| P10 | 0 | 0.41 | 1.00 | 0.47 | Delete |
| P11 | 0 | 0.00 | 0.50 | 0.17 | Delete |
| P12 | 0 | 0.00 | 0.75 | 0.25 | Delete |
| P13 | 0 | 0.00 | 1.00 | 0.33 | Delete |
| P14 | 0 | 0.00 | 0.50 | 0.17 | Delete |
| P15 | 0 | 0.33 | 0.75 | 0.36 | Delete |
| P16 | 0 | 0.00 | 0.50 | 0.17 | Delete |
| P17 | 0 | 0.00 | 0.50 | 0.17 | Delete |
| P18 | 0 | 0.00 | 0.50 | 0.17 | Delete |
| P19 | 0 | 0.00 | 0.50 | 0.17 | Delete |
| P20 | 0 | 0.00 | 0.75 | 0.25 | Delete |
| P21 | 0 | 0.00 | 0.25 | 0.08 | Delete |
| P22 | 0.25 | 0.62 | 1.00 | 0.62 | Delete |
| P23 | 0 | 0.00 | 0.75 | 0.25 | Delete |
| P24 | 0 | 0.00 | 0.50 | 0.17 | Delete |
| P25 | 0 | 0.00 | 0.50 | 0.17 | Delete |
| P26 | 0 | 0.00 | 0.50 | 0.17 | Delete |
| P27 | 0 | 0.00 | 0.50 | 0.17 | Delete |
| P28 | 0 | 0.00 | 0.50 | 0.17 | Delete |
| P29 | 0 | 0.00 | 0.50 | 0.17 | Delete |
| P30 | 0 | 0.00 | 0.50 | 0.17 | Delete |
| P31 | 0 | 0.00 | 0.50 | 0.17 | Delete |
| P32 | 0 | 0.00 | 0.50 | 0.17 | Delete |
| P33 | 0 | 0.00 | 0.50 | 0.17 | Delete |
| P34 | 0 | 0.00 | 0.50 | 0.17 | Delete |
| P35 | 0 | 0 | 0.5 | 0.17 | Delete |

**Table 4.4. The most important cost drivers based on only FDM.**

| Code | Parameters categories | Parameters | Initial Rank based on crisp value |
|------|----------------------|------------|-----------------------------------|
| **P1** | Civil | Command Area (hectare) | 1 |
| **P2** | Civil | PVC Length (m) | 2 |
| **P3** | Civil | Construction year and inflation rate | 3 |
| **P4** | Civil | Mesqa discharge ( capacity ) | 4 |
| **P5** | Mechanical | Number of Irrigation Valves ( alfa-alfa valve ) | 6 |
| **P7** | Electrical | Number of electrical pumps | 5 |

## 4.3.2 Fuzzy Analytic Hierarchy Process

The concept of Analytic Hierarchy Process (AHP) is a traditional powerful decision-making procedure to determine the priorities among different criteria and alternatives and to find an overall ranking of the alternatives Saaty (1980). The conventional AHP has no ability to deal with the imprecise or vague nature of linguistic assessment. Therefore, Laarhoven and Pedrycz (1983) combined Analytic Hierarchy Process (AHP) and Fuzzy Theory to produce FAHP. The deterministic numbers of traditional AHP method could be express by fuzzy numbers to obtain uncertainty when a decision maker is making a decision. The objective is to evaluate the most important cost parameters that were screened by Fuzzy Delphi method (6 parameters out of 35 parameters).

In FAHP, linguistic terms have been used in pair-wise comparison which can be represented by triangular fuzzy numbers (Erensal et al., 2006) and (Srichetta and Thurachon, 2012). Triple triangular fuzzy set numbers (l, m, u) were used as a fuzzy event where $l \leq m \leq u$ as shown in Table 4. Ma et al. (2010) have applied these steps:

Step.1: Defining criteria and build the hierarchical structure as shown in Fig.4.4.

Step.2: Set up pairwise comparative matrixes and transfer linguistic terms of positive triangular fuzzy numbers by Table 4.5.

Step.3: Generate group integration by Equation (4.8) as shown in Table 4.6.

Step.4: Calculate the fuzzy weight.

Step.5: De-fuzzify triangular fuzzy number into a crisp number.

Step.6: Rank de-fuzzified numbers.

**Table 4.5. The Fuzziness of linguistic terms for FAHP.**

| Linguistic Scale for Important | Triangular Fuzzy | | |
|---|---|---|---|
| | (L) | (M) | (U) |
| Just equal | 1.00 | 1.00 | 1.00 |
| Equally important | 1.00 | 1.00 | 3.00 |
| Weakly important | 1.00 | 3.00 | 5.00 |
| Essential or Strongly important | 3.00 | 5.00 | 7.00 |
| Very strongly important | 5.00 | 7.00 | 9.00 |
| Extremely Preferred | 7.00 | 9.00 | 9.00 |

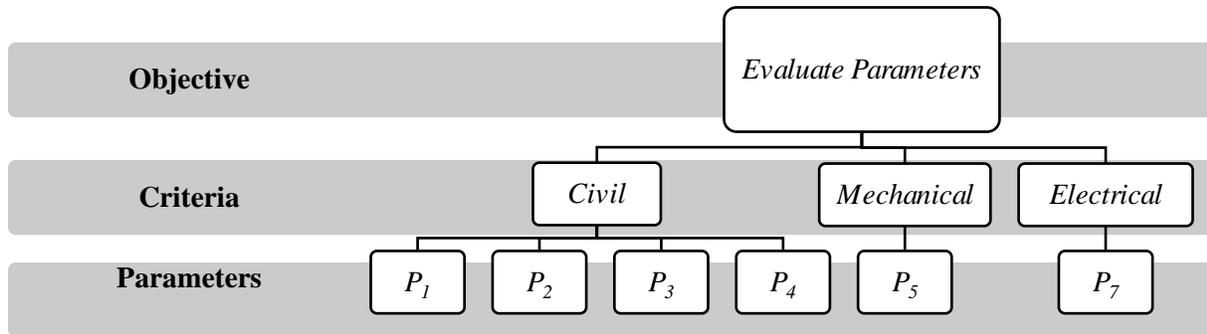

**Fig. 4.4. The hierarchical structure of selecting cost parameters for FCIPs.**

The current study questioned four experts who were FCIPs' managers and their experience more than 20 years. The experts' opinions were used to construct the fuzzy pair-wise comparison matrix to construct a fuzzy judgment matrix as illustrated in Table 4.6. After collecting the fuzzy judgment matrices from all experts by Equation (4.8), these matrices can be aggregated by using the fuzzy geometric mean (Buckley, 1985). The aggregated triangular fuzzy numbers of (n) decision makers' judgment in a certain case $W_{ij} = (L_{ij}, M_{ij}, U_{ij})$ as shown in Table 4.6 where C, M, and E were notations for civil, mechanical and electrical criteria respectively.

$$Wijn = \left( (\textstyle\prod_{n=1}^{n} l\ ijn)^{\frac{1}{n}}, (\textstyle\prod_{n=1}^{n} m\ ijn)^{\frac{1}{n}}, (\textstyle\prod_{n=1}^{n} u\ ijn)^{\frac{1}{n}} \right) \qquad (4.8)$$

Where:

i: a criterion such as civil, mechanical or electrical.

j: the screened cost parameter for FCIPs.

n: the number of experts.

$l_{ij}$: the minimum of the experts' common consensuses.

$m_{ij}$: the average of the experts' common consensuses.

$u_{ij}$: the maximum of the experts' common consensuses.

$L_j$: opinions mean of the minimum of the experts' common consensuses ($l_{ij}$).

$M_j$: opinions mean of the average of the experts' common consensuses ($M_{ij}$).

$U_j$: opinions mean of the maximum of the experts' common consensuses ($U_{ij}$).

$W_{ij}$: the aggregated Triangular Fuzzy Numbers of the $n^{th}$ expert's view.

**Table.4.6 The aggregate expert's fuzzy opinions about the main criteria.**

| Criteria | Aggregate experts' opinions | | | | | | | | |
|----------|------|------|------|------|------|------|------|------|------|
|          | C(l) | C (m) | C (u) | M(l) | M(m) | M(u) | E(l) | E(m) | E(u) |
| Civil    | 1.00 | 1.00 | 1.00 | 1.73 | 3.87 | 5.92 | 3.87 | 5.92 | 7.94 |
| Mechanical | 0.20 | 0.26 | 0.58 | 1.00 | 1.00 | 1.00 | 1.00 | 1.73 | 3.87 |
| Electrical | 0.13 | 0.17 | 0.26 | 0.26 | 0.58 | 1.00 | 1.00 | 1.00 | 1.00 |

Where:

C (l): the aggregation of the minimum of the experts' common consensuses for civil criterion(C).

C (m): the aggregation of the average of the experts' common consensuses for civil criterion(C).

C (u): the aggregation of the maximum of the experts' common consensuses for civil criterion(C).

M (l): the aggregation of the minimum of the experts' common consensuses for mechanical criterion (M).

M (M): the aggregation of the average of the experts' common consensuses for mechanical criterion (M).

M (u): the aggregation of the maximum of the experts' common consensuses for mechanical criterion (M).

E (l): the aggregation of the minimum of the experts' common consensuses for electrical criterion (E).

E (M): the aggregation of the average of the experts' common consensuses for electrical criterion (E).

E (u): the aggregation of the maximum of the experts' common consensuses for electrical criterion (E).

Based on the aggregated pair-wise comparison matrix , the value of fuzzy synthetic extent $S_i$ with respect to the $i^{th}$ criterion can be computed by Equation (4.9) by algebraic operations on triangular fuzzy numbers [(Saaty , 1994) and (Srichetta and Thurachon ,2012)]. Table 4.7 and Table 4.8 showed these operations.

$$Si = \sum_{j=1}^{m} Wij * \left[\sum_{i=1}^{n}\sum_{j=1}^{m} Wij\right]^{-1} \qquad (4.9)$$

Where:

i: a criterion such as civil, mechanical or electrical.

j: the screened cost parameter for FCIPs.

$W_{ij}$: the aggregated triangular fuzzy numbers of the $n^{th}$ expert's view.

$S_i$: the value of fuzzy synthetic extent.

**Table 4.7. The sum of horizontal and vertical directions.**

| | Row | | | Column | | | the fuzzy synthetic extent of each criterion | | |
|---|---|---|---|---|---|---|---|---|---|
| | C (l) | C (m) | C (u) | M(l) | M(m) | M(u) | (l) | (m) | (u) |
| **C** | 6.61 | 10.79 | 14.85 | 1.33 | 1.43 | 1.84 | 0.29 | 0.69 | 1.46 |
| **M** | 2.20 | 2.99 | 5.45 | 2.99 | 5.45 | 7.92 | 0.10 | 0.19 | 0.53 |
| **E** | 1.38 | 1.75 | 2.26 | 5.87 | 8.65 | 12.81 | 0.06 | 0.11 | 0.22 |
| | | | | 10.19 | 15.53 | 22.56 | | | |

Based on the fuzzy synthetic extent values, this study used Chang's method (Saaty, 1980) to determine the degree of possibility by Equation (4.10). The final result of priority rate was shown in Table 4.8.

$$V(S_m \geq S_c) = \left\{ \begin{array}{c} 1, \ \ if \ m_m \geq m_c \\ 0, \ \ \ \ if \ l_m \geq u_c \\ \dfrac{l_c - u_m}{(l_m - u_c) - (l_m - u_c)} , Otherwise \end{array} \right\} \qquad (4.10)$$

Where:

$V(S_m \geq S_c)$: the degree of possibility between a civil criterion (c) and a mechanical criterion (m).

$(l_c, m_c, u_c)$: the fuzzy synthetic extent of civil criterion from (Table 4.7).

$(l_m, m_m, u_m)$: the fuzzy synthetic extent of mechanical criterion from (Table 4.7).

**Table.4.8. the weights and normalized weights.**

|  |  | Weight = Minimum value | Normalized weight |
|---|---|---|---|
| **S C > S M** | 1.00 | 1.00 | 0.75 |
| **S C > S E** | 1.00 |  |  |
| **S M > S E** | 1.00 | 0.33 | 0.25 |
| **S M > S C** | 0.33 |  |  |
| **S E > S C** | 0.00 | 0.00 | 0.00 |
| **S E > S M** | 0.00 |  |  |
|  | Sum | 1.33 |  |

There are two priorities, one for criteria and the other for parameters. Therefore, the next step after criteria priority calculations is to calculate priority of the parameters. Similarity, the transformation procedures for comparison between criteria based on each parameter will be calculated. The relative weights of criteria based on each parameter are shown in Table 4.8. Subsequently, the final results of normalized weights from this table with respect to the overall criteria weights are computed and illustrated in Table 4.9.

**Table 4.9. Final priorities of parameters.**

| Criteria | priority of criteria | Parameters | priority of parameters | Final priority | CR of Parameters | CR of Criteria |
|---|---|---|---|---|---|---|
| Civil ( C ) | 0.75 | P1 | 0.45 | 0.34 | | |
| Civil ( C ) | 0.75 | P2 | 0.29 | 0.22 | | |
| Civil ( C ) | 0.75 | P3 | 0.26 | 0.19 | Acceptable | Acceptable |
| Civil ( C ) | 0.75 | P4 | 0.00 | 0.00 | | |
| Mechanical (M) | 0.25 | P5 | 1.00 | 0.25 | | |
| Electrical (E) | 0.00 | P6 | 0.00 | **0.00** | | |

Finally, Consistency Test of the Comparison Matrix has been conducted to measure the consistency of judgment of the decision maker (Saaty, 1980). The maximum value of Consistency Ratio (CR) was 0.028. This value is acceptable where its value should not exceed 0.1 for a matrix larger than 4x4 (Srichetta and Thurachon, 2012).

## 4.4 DISCUSSION

The purpose of the current study is to identify and evaluate the key cost drivers where two different procedures were used. A total of 35 parameters were collected and evaluated by traditional Delphi methods (TDM) and then screened by Likert scale. The results of the first procedure (TDM and Likert scale) were seven parameters out of 35 parameters. The screened parameters by the first procedure were command area (hectare), PVC length (m), construction year and inflation rate, a number of irrigation valves (alfalfa valve), mesqa discharge (capacity), consultant performance and errors in design, a number of electrical pumps.

In contrast, the second procedure has applied FDM and FAHM where four parameters out of 35 parameters have been identified. The screened parameters by the first procedure were command area (hectare), PVC length (m), and construction year and inflation rate, a number of irrigation valves. The number of key drivers by the first procedure (seven cost drivers) was more than the number of cost drivers by the second procedure (four cost drivers). In addition, by using only FDM (without FAHP), the cost drivers were command area (hectare), PVC length (m), construction year and inflation rate, mesqa discharge number of electrical pumps ( capacity ), a number of irrigation valves, a number of electrical pumps (six cost drivers).

Accordingly, as showed Table 4.10, the cost model developer has three options to develop a cost model where four, six, or seven parameters can be used to be the model's inputs. In addition, the key cost drivers by second procedure existed in the first procedure. Thus proved that two procedure were reliable to identify the cost drivers. Generally, model developers and model users prefer to use fewer input parameters as fewer inputs mean little data collection, little effort, a little time needed. Accordingly, the second procedure is more practical than the first procedure. Moreover, applying only FDM were better than applying the first procedure due to uncertainty considerations.

Moreover, the second procedure takes into account uncertainty by applying fuzzy theory. Finally, this study recommended conducting the second procedure than the first one to perform high efficiency for cost driver's identification and to develop a more reliable cost model.

**Table.*4*.10 Comparison among 1st approach and (2nd approach)**

| Comparison criteria | TDM (1st approach) | FDM (without FAHP) | FDM + FAHP (2nd approach) |
|---|---|---|---|
| Number of key cost drivers | 7 | 6 | 4 |
| Calculations | simple | more complicated than TDM and less than (FDM + FAHP) | complicated |
| Uncertainty | no uncertainty | exist | exist |

## 4.5 CONCLUSION

This chapter discussed the qualitative methods such as Delphi techniques and Fuzzy Analytical hierarchy Process (FAHP) to collect, rank and evaluate the cost drivers of the FCIPs. The current study has used two procedures where both Traditional Delphi Method (TDM) and Fuzzy Delphi Method (FDM) were used to collect and initially rank the cost drivers. Based on the second approach, The FAHP was used to finally rank the screened parameters by FDM. Out of 35 cost drivers, only four parameters were selected as final parameters. The contribution of this study was to find out and evaluate these parameters and to maintain the ability of FDT and FAHP to collect and evaluate the cost drivers of a certain case study. To obtain uncertainty and achieve a more practical model, this study suggested using a fuzzy theory with Delphi methods and with AHP. The screened parameters can be used to develop a precise parametric cost model for FCIPs as a future research work.

# CHAPTER 5

# QUANTITATIVE APPROACH

## 5.1 Introduction

The objective of the chapter is to identify the effective predictors among the complete set of predictors. This could be achieved by deleting both irrelevant predictors (i.e. variables not affecting the dependent variable). Therefore, key predictors selection methodology is based on a trinity of selection methodology:

1. Statistical tests (for example, F, chi-square and t-tests, and significance testing);
2. Statistical criteria (for example, R-squared, adjusted R-squared, Mallows' C p and MSE);
3. Statistical stopping rules (for example, P -values flags for variable entry / deletion / staying in a model) (Ratner, 2010).

As illustrated in the previous chapter, step 1 and 2 of the study methodology include literature survey to review previous practices for key parameters identification, collecting data of FCIPs, respectively. Step 3 is to conduct four statistical methods to analysis data and identify cost drivers. These methods are Exploratory Factor Analysis (EFA), Regression methods, correlation matrixes and hybrid methods. Finally, Step 4 compares results of statistical methods to select the best logic set of key cost drivers suitable for the conceptual stage of FCIPs. After collecting relevant data which represents all variables, statistical methods can be used to analysis data and screen such variables. All methods are developed using software SPSS 19 for Windows.

## 5.2 Data Collection

A total of 111 historical cases of FCIPs are randomly collected between 2010 and 2015. Table 5.1 illustrates a descriptive statistics of collected data where mean and standard deviation are calculated for each variable where 17 variables are named from $P_1$ to $P_{17}$. These variables can be collected based only on contracts information and construction site records as the quantitative data. The PVC pipe diameters are ranging from 225 mm to 350 mm. To collect data in one parameter, Equation (5.1) used to prepare equivalent diameter ($P_4$). Appendix C (collected data snap shot) illustrates a snap shot of all collected data.

$$\text{Equivalent Diameter} = \frac{\sum_i^n \text{Diameter} * \text{Length}}{\sum_i^n \text{Length}} \qquad (5.1)$$

## Table 5.1. Descriptive Statistics for training data.

| ID | Variables | Unit | Minimum value | Maximum value |
|---|---|---|---|---|
| $P_1$ | Area served | hectare | 19 | 100 |
| $P_2$ | Average area served sections | hectare | 2.65 | 13.1 |
| $P_3$ | Total length of pipeline | m | 119 | 1832 |
| $P_4$ | Equivalent Diameter | mm | 225 | 313.4 |
| $P_5$ | Duration (working days) | day | 58 | 122.5 |
| $P_6$ | Irrigation valves number | unit | 3 | 27 |
| $P_7$ | Air and pressure relief valves number | unit | 1 | 7 |
| $P_8$ | sump (its diameter 1.7) | unit | 0 | 1 |
| $P_9$ | Pump house (its size 3m*4m) | unit | 0 | 1 |
| $P_{10}$ | Max discharge | liter/sec | 40 | 120 |
| $P_{11}$ | Electrical pump discharge | liter/sec | 40 | 120 |
| $P_{12}$ | Diesel pump discharge | liter/sec | 40 | 120 |
| $P_{13}$ | Orientation | ----- | 0 | 3 |
| $P_{14}$ | Construction year | year | 2010 | 2015 |
| $P_{15}$ | Rice | ----- | 0 | 1 |
| $P_{16}$ | Intake existence | unit | 0 | 1 |
| $P_{17}$ | Ganabiaa canal | ------ | 0 | 1 |

Table 5.1 presents the collected parameters of FCIPs where $P_1$ represents the area served by the improved field canal, $P_2$ represents the average area of area served sections where the total area served are divided into sections based on the number of irrigation valves, $P_3$ represents the total length of PVC pipeline, $P_4$ represents the equivalent diameter of the improved field canal where this value can be calculated by Equation (5.1), $P_5$ represents the total construction duration as working days. $P_6$ represents the number of irrigation valves used during construction of FCIP. $P_7$ represents the number of air and pressure relief valves.

There are two sizes of sump structures used in FCIP, these sizes are 1.7 m and 1.9 m. $P_8$ represents the existence of sump with diameter of 1.7 m by binary code where 0 represents sump with diameter of 1.9 m and 1 represents sump with diameter of 1.7 m., $P_9$ represents the existence of pump house with size of 3m * 4m by binary code where 0 represents pump house with size of 3m * 3m and 1 represents pump house with size of 3m * 4m.

$P_{10}$ represents the maximum discharge that can be pumped through the pipeline. $P_{11}$, $P_{12}$ represents the discharge of electrical and diesel pump used in FCIP respectively. $P_{13}$ represents the orientation of FCIP divided into four cases. These cases are no intersection (code = 0), intersecting with drain (code = 1), intersecting with road (code = 2) and intersecting with both road and drain (code = 3). This case exists when pipeline intersect with drain, this requires an aqueduct to pass irrigation water over this drain, whereas the pipeline intersect with road this requires a boring excavation under road to avoid blocking the intersecting road. $P_{14}$ represents the construction year of FCIP ranging between 2010 and 2015. The area served can be cultivated of rice or of any other crop, rice crop needs to be submerged by water and that causes problems during soil excavation process. Therefore, $P_{15}$ represents the existence of rice crop or not. Similarly, Some FCIPs have intakes and the others has no intakes, $P_{16}$ represents the existence of intake structure. There are two types of FCIP, these types are Ganabiaa canal or perpendicular canal. In Ganabiaa, field canal is parallel to branch canal whereas the other type of FCIP where field canal is perpendicular to the branch canal. $P_{17}$ represent that case.

## 5.3 Exploratory Factor Analysis (EFA)

Factor analysis is a statistical method to covert correlated variables to a lower number of variables called factors. This method can be used to screen data to identify and categorize key parameters. Exploratory factor analysis (EFA) and Confirmatory factor analysis (CFA) are two types of factor analysis. Using (EFA), relations among items and groups are identified where no prior assumptions about factors relationships are proposed. Confirmatory factor analysis (CFA) is a more complex technique where uses structural equation modeling to evaluate relationships between observed variables and unobserved variables (Polit DF Beck CT, 2012).

There are many types of factoring such as Principal Component Analysis (PCA), Canonical factor analysis and Image factoring etc. This study uses (PCA). PCA is a factor extraction method where factor weights are computed to extract the maximum possible variance. Subsequently, the factor model must be rotated for analysis (Polit DF Beck CT, 2012). The aim of PCA is reducing variable where no assumptions are proposed whereas factor analysis requires assumptions to be used. PCA is related to (EFA) where PCA is a version of (EFA). The advantage of EFA is to combine two or more variables into a single factor that reduces the number of variables. However, factor analysis cannot causality to interpret data factored.

EFA is conducted by using (PCA) where the objective of PCA is to reduce a lot of variables to few variables and to understand the structure of a set of variables (Field, 2009). The following question should be answered to conduct EFA:

(1) How large the sample needs to be?
(2) Is there multicollinearity or singularity?
(3) What is the method of data extraction?
(4) What is the number of factors to retain?
(5) What is the method of factor rotation?
(6) Choosing between factor analysis and principal components analysis?
, all these questions will be answered in the following sections.

**5.3.1 Sample size**

       Factors obtained from small datasets do not generalize as well as those derived from larger samples. Some researchers have suggested using the ratio of sample size to the number of variables as a criterion. The following Table 5.2 summarized the main finding of the involved researchers

**Table 5.2. The review of sample size requirement.**

| | Author / Reference | Review of Findings |
|---|---|---|
| 1 | Nunnally (1978) | required sample size is 10 times as many participants as variables |
| 2 | Kass and Tinsley (1979) | required sample size between 5 and 10 participants per variable up to a total of 300 cases |
| 3 | Tabachnick and Fidell (2007) | Required sample size is at least 300 cases for factor analysis. 50 observations are very poor, 100 are poor, 200 are fair, 300 are good, 500 are very good and 1000 or more is excellent. |
| 4 | Comrey and Lee (1992) | The sample size can be classified to 300 as a good sample size, 100 as poor and 1000 as excellent. |
| 5 | Guadagnoli and Velicer (1988) | The sample size depends on the factor loading where factors with 10 or more loadings greater than 0.40 are reliable if the sample size is greater than 150, and factors with a few low loadings should be at least 300 or more and suggested a minimum sample size of 100 - 200 observations where the ratio of cases to variables is 4.9 |
| 6 | Allyn, Zhang, and Hong (1999) | The minimum sample size or sample to variable ratio depends on the design of the study where a sample of 300 or more will probably provide a stable factor solution. |
| 7 | (Kaiser, 1970) , Kaiser (1974), (Hutcheson & Sofroniou, 1999) | Another manner is to use the Kaiser–Meyer–Olkin measure of sampling adequacy (KMO) (Kaiser, 1970). The KMO is calculated for individual and multiple variables based on the ratio of the squared correlation between variables to the squared partial correlation between variables. Kaiser (1974) recommends that values greater than 0.5 are barely acceptable (values below this should lead you to either collect more data or rethink which variables to include). Moreover, values between 0.5 and 0.7 are mediocre, values between 0.7 and 0.8 are good, values between 0.8 and 0.9 are great and values above 0.9 are superb. |
| 8 | (Kline, 1999) | the absolute minimum sample size required (e.g., 100 participants; Kline, 1999) |

In the current study, there are 17 variables and the collected data set are 111 historical cases, the ratio between cases to variables is (6.5). According to (Kline, 1999); Guadagnoli and Velicer (1988), this ratio is initially acceptable. According to Tabachnick and Fidell (2007) and Comrey and Lee (1992), this sample size is classified as a poor sample. Furthermore, the Kaiser–Meyer–Olkin measure of sampling adequacy (KMO) is conducted in the later section (Kaiser, 1970).

### 5.3.2 Correlation among variables

The first iteration to check correlation and avoid multicollinearity and singularity (Tabachnick and Fidell, 2007). Multicollinearity is that variables are correlated too highly whereas singularity is that variables are perfectly correlated. It is used to describe variables that are perfectly correlated (it means the correlation coefficient is 1 or -1). There are two methods for assessing multicollinearity or singularity:

1) The first method is conducted by scanning the correlation matrix for all independent variables to eliminate variables with correlation coefficients greater than 0.90 (Field, 2009) or correlation coefficients greater than 0.80 (Rockwell, 1975).

2) The second method is to scan the determinant of the correlation matrix. Multicollinearity or singularity may be in existence, if the determinant of the correlation matrix is less than 0.00001. One simple heuristic is that the determinant of the correlation matrix should be greater than 0.00001 (Field, 2009). IF the visual inspection reveals no substantial number of correlations greater than 0.3, PCA probably is not appropriate. Also, any variables that correlate with no others ($r = 0$) should be eliminated (Field, 2009).

The present study has the following iterations:

Iteration 1: remove any variable higher than 0.9 with all independent variables to avoid singularity and multicollinearity (17 variables).

Iteration 2: the determinant of the correlation matrix is equal to 0.000 which is less than 0.00001. It implies that there is a problem of multicollinearity. By trial and error, it is found that (P10), (P11), (P12) and (P13) caused multicollinearity. Therefore, these factors have been deleted. Accordingly, the remaining variables are 13 variables.

Iteration 3: EFA is repeated for the third time after removing these parameters. The determinant of the correlation matrix is equal to 0.001, which it is greater than 0.00001.

Iteration 4: the Anti-Image correlation matrix that contains Measures of Sampling Adequacy (MSA) is examined. All diagonal elements should be greater than 0.5 whereas the off-diagonal elements should all be very small (close to zero) in a good model (Field, 2009). The scan of Anti-image correlation matrix diagonal elements greater than 0.5 except three variables P2, P15, and P16 which has values less than 0.5 equals to 0.459, 0.178, 0.383 respectively as illustrated in Table 5.3. The remaining variables are now 10 variables.

**Table *5*.3. SPSS Anti-image Correlation.**

| ID | P1 | P2 | P3 | P4 | P5 | P14 | P15 | P16 | P17 |
|----|----|----|----|----|----|-----|-----|-----|-----|
| **P1** | .805 | -.180 | .075 | -.610 | -.140 | -.111 | .065 | -.166 | -.002 |
| **P2** | -.180 | **.459** | -.002 | -.119 | -.056 | -.064 | .104 | -.048 | -.135 |
| **P3** | .075 | -.002 | .619 | -.031 | -.882 | .079 | .809 | -.178 | -.148 |
| **P4** | -.610 | -.119 | -.031 | .808 | -.002 | .066 | .044 | .186 | -.096 |
| **P5** | -.140 | -.056 | -.882 | -.002 | .584 | -.055 | -.912 | .185 | .150 |
| **P6** | -.242 | .403 | -.030 | .077 | -.086 | .027 | .120 | -.012 | -.072 |
| **P7** | .015 | .114 | -.144 | .010 | -.134 | -.120 | .178 | -.073 | .046 |
| **P8** | -.241 | .060 | -.155 | -.052 | .084 | .031 | -.089 | .008 | -.071 |
| **P9** | -.115 | .044 | .081 | -.092 | -.149 | .178 | .088 | -.001 | .018 |
| **P14** | -.111 | -.064 | .079 | .066 | -.055 | .531 | .070 | .197 | -.109 |
| **P15** | .065 | .104 | .809 | .044 | -.912 | .070 | **.176** | -.202 | -.154 |
| **P16** | -.166 | -.048 | -.178 | .186 | .185 | .197 | -.202 | **.383** | -.130 |
| **P17** | -.002 | -.135 | -.148 | -.096 | .150 | -.109 | -.154 | -.130 | .644 |

### 5.3.3 Kaiser-Meyer-Olken measure of sampling adequacy

Kaiser-Meyer-Olken (KMO) is another measure to compute the degree of inter-correlations among variables. The KMO statistic varies between 0 and 1(Kaiser, 1974). A value of 0 shows that the sum of partial correlations is large relative to the sum of correlations, which means that there is diffusion in the pattern of correlations.

Therefore, factor analysis is likely to be inappropriate. A value close to 1 shows that patterns of correlations are relatively compact and that factor analysis should yield reliable factors. In the present study, KMO measure of sampling adequacy is 0.69 which is classified as mediocre.

### 5.3.4 Bartlett's test

Bartlett's test can be used to test the adequacy of the correlation matrix. It tests the null hypothesis that the correlation matrix is an identity matrix where all the diagonal values are equal to 1 and all off-diagonal values are equal to 0. A significant test indicates that the correlation matrix is not an identity matrix where a significance value less than 0.05 and null hypothesis can be rejected (Field, 2009). The significance value (p-value) = 0.000 where less than significance level. Therefore, it indicates that correlations between variables are sufficiently large for Factor Analysis.

### 5.3.5 Factor extraction by using principal component analysis (PCA)

Factor (component) extraction is the second step in conducting EFA to determine the smallest number of components that can be used to best represent interrelations among a set of variables (Tabachnick and Fidell, 2007). As shown in Table 5.4, communalities for retained variables after extraction are more than 0.5 which show that these variables are reflected well by extracted factors. Accordingly, the factor analysis is reliable (Field, 2009). The Kaiser criterion  stated that if the number of variables is less than 30, the average communality is more than 0.7 or when the number of variables is greater than 250, the mean communality is near or greater than 0.6 (Stevens, 2002). Based on this criterion, only 6 parameters have been retained. If the 0.7 is considered as a threshold, then the parameters will be P1, P3, P4, P5, P6, and P7 as shown in Table 5.4.

**Table 5.4. Communalities for each parameters.**

| ID | Extraction |
|---|---|
| P1 | 0.821 |
| P3 | 0.846 |
| P4 | 0.778 |
| P5 | 0.933 |
| P6 | 0.700 |
| P7 | 0.826 |
| P8 | 0.527 |
| P9 | 0.630 |
| P14 | 0.616 |
| **Average communalities** | **0.714** |

This result can be confirmed by retaining all components with eigenvalues more than 1 that contains five components. Table 5.5 illustrates initial eigenvalues with an eigenvalue of one or more are retained where that contains five components and percent of variance before and after the rotation. Table 5.6 illustrates the components with each parameter. Finally, Table 5.6 shows Rotated Component Matrix where the highest loading parameters for the first component are P3, P1, P5, P4, P6, and P7 respectively.

**Table 5.5. Total variance explained.**

| Component | Initial Eigenvalues | | | Extraction Sums of Squared Loadings | |
|---|---|---|---|---|---|
| | Total | % of Variance | Cumulative % | Total | % of Variance |
| 1 | 4.628 | 35.601 | 35.601 | 4.628 | 35.601 |
| 2 | 1.545 | 11.887 | 47.488 | 1.545 | 11.887 |
| 3 | 1.307 | 10.051 | 57.539 | 1.307 | 10.051 |
| 4 | 1.083 | 8.331 | 65.870 | 1.083 | 8.331 |
| 5 | 1.006 | 7.735 | 73.605 | 1.006 | 7.735 |

**Table 5.6. Component matrix.**

| ID | Component | | | | |
|---|---|---|---|---|---|
| | 1 | 2 | 3 | 4 | 5 |
| P3 | .847 | | -.197 | | .292 |
| P1 | .844 | .219 | .123 | | -.213 |
| P5 | .821 | -.228 | .135 | .361 | .241 |
| P4 | .737 | .355 | | | -.315 |
| P6 | .728 | -.248 | -.217 | -.128 | -.201 |
| P7 | .565 | .110 | -.424 | | .561 |

## 5.4 Regression methods

Regression analysis can be used for both cost drivers selection and cost prediction modeling. The current study focused on cost driver's selection. Therefore, the forward, backward, stepwise methods is applied as following:

### 5.4.1 Forward method

Forward selection initiates with no variables in the model where each added variable is tested by a comparison criterion to improve the model performance. If the independent variable significantly improves the ability of the model to predict the dependent variable, then this predictor is retained in the model and the computer searches for a second independent variable (Field, 2009). Results are illustrated in Table 5.7 where model 1 included a single variable (P3) and its correlation factor is (0.85).

**Table 5.7. Forward method results.**

| Model | Independent Variable | R | R Square | Adjusted R Square |
|-------|---------------------|------|------|------|
| **1** | P3 | 0.85 | 0.73 | 0.72 |
| **2** | P3, P14 | 0.89 | 0.80 | 0.79 |
| **3** | P3, P14,  P10 | 0.92 | 0.84 | 0.84 |
| **4** | P3, P14,  P10, P11 | 0.93 | 0.87 | 0.86 |
| **5** | P3, P14,  P10, P11, P9 | 0.94 | 0.89 | 0.89 |
| **6** | P3, P14,  P10, P11, P9, P5 | 0.95 | 0.90 | 0.90 |
| **7** | P3, P14,  P10, P11, P9, P5, P8 | 0.95 | 0.91 | 0.90 |
| **8** | P3, P14,  P10, P11, P9, P5, P8, P6 | 0.96 | 0.92 | 0.91 |

## 5.4.2 Backward elimination method

The backward method is the opposite of the forward method. In this method, all input  independent variables are initially selected, and then the most unimportant independent variable are eliminated one-by-one based on  the significance value of the t-test for each  variable. The contribution of the remaining variable is then reassessed (Field, 2009). The results are  illustrated in Table 5.8 where model 1 includes 16 variables and its correlation factor is (0.96).

**Table 5.8 Backward elimination method results.**

| Model | Independent Variable | R | R Square | Adjusted R Square |
|-------|---------------------|------|------|------|
| 1 | P17,  P7, P14, P15, P2, P16, P13, P8, P9, P6, P4, P3, P12, P1, P5,  P10 | 0.96 | 0.93 | 0.92 |
| 2 | P17,  P7, P14, P15, P2, P16, P13, P8, P9, P6, P3, P12, P1, P5, P10 | 0.96 | 0.93 | 0.92 |
| 3 | P17,  P7, P14, P15, P2, P13, P8, P9, P6, P3, P12, P1, P5, P10 | 0.96 | 0.93 | 0.92 |
| 4 | P17,  P7, yearP14, P15, P2, P8, P9, P6, P3, P12, P1, P5,P10 | 0.96 | 0.93 | 0.92 |
| 5 | P17,  P7, P14, P15, P2, P8, P9, P6, P3, P12, P5,  P10 | 0.96 | 0.93 | 0.92 |
| 6 | P17,  P7, P14, P15, P2, P8, P9, P6, P3, P12,  P10 | 0.96 | 0.93 | 0.92 |
| 7 | P7, P14, P15, P2, P8, P9, P6, P3, P12, P10 | 0.96 | 0.93 | 0.92 |

### 5.4.3 Stepwise Method

Stepwise selection is an extension of the forward selection approach, where input variables may be removed at any subsequent iteration (Field, 2009). The results are illustrated in Table 5.9 where model 1 includes a single variable (P3) and its correlation factor is (0.85).

**Table 5.9. Stepwise Method results.**

| Model | Independent Variable | R | R Square | Adjusted R Square |
|---|---|---|---|---|
| 1 | P3 | 0.85 | 0.73 | 0.72 |
| 2 | P3, P14 | 0.89 | 0.80 | 0.79 |
| 3 | P3, P14, P10 | 0.92 | 0.84 | 0.84 |
| 4 | P3, P14, P10, P11 | 0.93 | 0.87 | 0.86 |
| 5 | P3, P14, P10, P11, P9 | 0.94 | 0.89 | 0.89 |
| 6 | P3, P14, P10, P11, P9, P5 | 0.95 | 0.90 | 0.90 |
| 7 | P3, P14, P10, P11, P9, P5, P8 | 0.95 | 0.91 | 0.90 |
| 8 | P3, P14, P10, P11, P9, P5, P8, P6 | 0.96 | 0.92 | 0.91 |

## 5.5 Correlation

The relation among all variables are showed in the correlation matrix, the aim is to screen variable based only on the correlation matrix. Therefore, all independent variables that are highly correlated with each other will be eliminated (R>= 0.8) and all dependent variables that are low correlated with the dependent variable (R <=0.3) will be eliminated by using both person and spearman matrices.

### 5.5.1 Pearson Correlation

Pearson Correlation is a measure of the linear correlation between two variables, giving a value between +1 and −1 where 1 is the positive correlation, 0 is no correlation, and −1 is the negative correlation. It is developed by Karl Pearson as a measure of the degree of linear dependence between two variables (Field, 2009). After the first scan of correlation matrix, it found that:

First, the Correlation among independent variables, (P10), (P11), (P12) and (P13) is highly correlated with (P1) where correlation factor is approximately 0.86 for them.

Second, Correlation among independent variables and the dependent variable, (P14), Rice (P15), (P16) and (P17) are low correlated with the dependent variable (the cost of FCIP) where there is no relation between them. The correlation coefficient are 0.071, 0.206, 0.036, 0.104 and 0.19 respectively.

Therefore, these variables are eliminated and correlation matrix scanned for the second time where all correlations are significant at P=0.01 (level 2-tailed). Selected variables are P1, P3, P4, P5, P6, P7, P8 and P9.

## 5.5.2 Spearman correlation

Spearman correlation is a nonparametric measure of statistical dependence between two variables using a monotonic function using a monotonic function. A perfect Spearman correlation of +1 or −1 occurs when each of the variables is a perfect monotone function of the other. First, the Correlation among independent variables, (P10), (P11) and (P12) is highly correlated with (P1) where correlation factor is approximately 0.86 for them. Second, Correlation among independent variables and the dependent variable, (P14), (P15), and (P16) and (P17) are low correlated with the dependent variable (the cost of FCIP) where there is no relation between them. The correlation coefficient are 0.12, 0.18, 0.05, 0.14 and 0.19 respectively. Therefore, these variables are eliminated and correlation matrix scanned for the second time, all correlations are significant at P=0.01 (level 2-tailed). The selected variables are (P1), (P3), (P4), (P5), (P6), (P7), (P8) and (P9) and (P13).

## 5.6 Hybrid Method

A hybrid model is to merge two methods as one method to obtain better results. The first model is to conduct Person correlation where all independent variables are high correlated (R>=0.8) is eliminated and all independent variables low correlated (R =< 0.3) with dependent variable are eliminated, and then to conduct stepwise method to identify the final selection of variables. The results show in Table 5.10.

**Table *5*.10. The results of the first iteration of hybrid model (1).**

| Hybrid Model 1 | Independent Variable | R | R Square | Adjusted R Square |
|---|---|---|---|---|
| 1 | P3 | 0.85 | 0.73 | 0.72 |
| 2 | P3, P1 | 0.88 | 0.77 | 0.76 |
| 3 | P3, P1, P6 | 0.88 | 0.78 | 0.77 |
| 4 | P3, P1, P6, P7 | 0.89 | 0.79 | 0.78 |

The second hybrid model is to conduct the first approach without deleting independent variables low correlated (R =< 0.3) with the dependent variable as shown in Table 5.11.

**Table 5.11. The results of the first iteration of hybrid model (2).**

| Hybrid Model (2) | Independent Variable | R | R Square | Adjusted R Square |
|---|---|---|---|---|
| 1 | P3 | 0.85 | 0.73 | 0.72 |
| 2 | P3, P14 | 0.89 | 0.80 | 0.79 |
| 3 | P3, P14, P6 | 0.92 | 0.84 | 0.83 |
| 4 | P3, P14, P6, P1 | 0.93 | 0.86 | 0.85 |
| 5 | P3, P14, P6, P1, P9 | 0.93 | 0.87 | 0.86 |

## 5.7 Discussion of results

In the current study, the correlation coefficient among dependent variable and the independent variable is used as the benchmark to compare the results of variable extraction methods. Table 5.12 and Fig. 5.1 summarized the results of methods where correlation is shown against the number of variables.

**Table 5.12. Results of all methods.**

| Method | Select Variables | R as a bench mark | Number of variables |
|---|---|---|---|
| EFA | P3, P1, P5, P4,P6 and P7 | 0.89 | 6 |
| Forward Method | Table 5.7 | | |
| Backward Method | Table 5.8 | | |
| Stepwise Method | Table 5.9 | | |
| Pearson Correlation | P1 , P3, P4, P5 , P6, P7, P8 and P9 | 0.89 | 8 |
| Spearman Correlation | P1 , P3, P4, P5, P6, P7, P8, P9 and P13 | 0.89 | 9 |
| hybrid model 1 | Table 5.10 | | |
| hybrid model 2 | Table 5.11 | | |

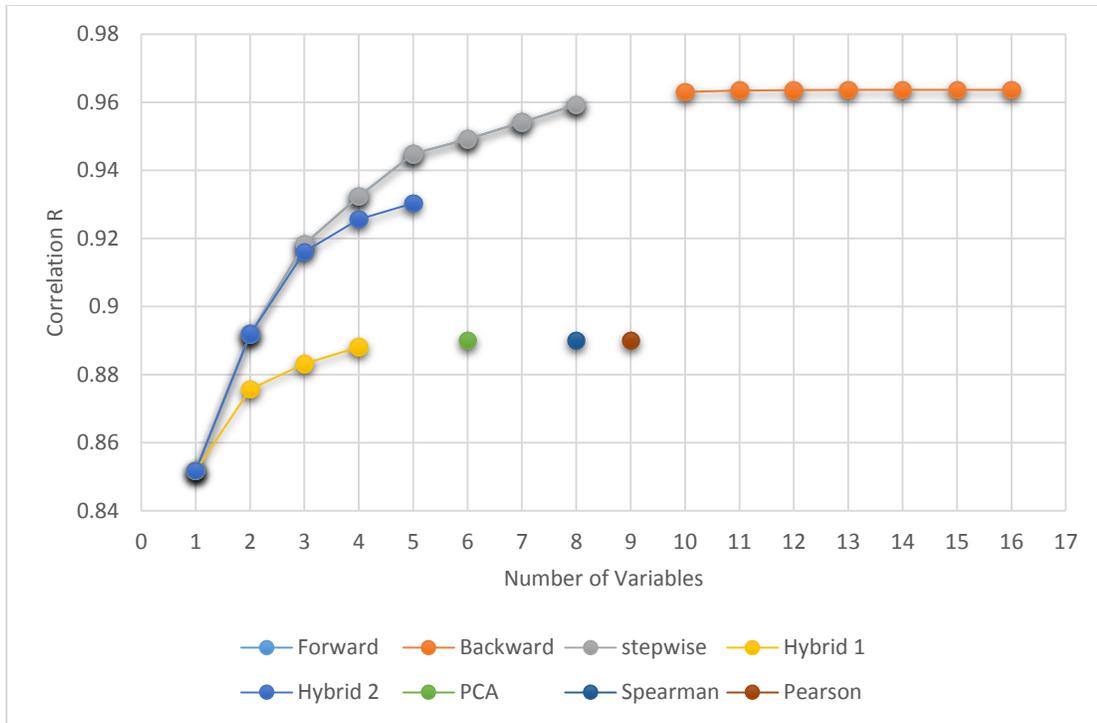

**Fig. 5.1 the plotted results for each method.**

Fig. 5.1 can be used by a model developer to choose key cost drivers based on the following two criterion: First, the fewer number of variables that can represent the highest correlation with the outcome (cost of FCIP). Second, the availability of data at the conceptual stage where this chart provides a set of alternatives of variables to give the same accuracy with the outcome. For example, the model developer wants to develop a model with the highest accuracy, Fig. 5.1 suggests to use backward elimination method to provide high correlation (R=0.96). However, the number of the required variables may be ten variables and that may not be available at the conceptual stage.

The second logic trial is to use the fewer variables (assume four variable), Fig.5.1 suggests the two methods with approximately the same correlation (Stepwise and Hybrid model 2). By looking at the corresponding Table 5.9 and Table 5.11, Stepwise method variables are (P3), (P14), (P10), and (P11) whereas Hybrid model 2 are (P3), (P14), (P6), and (P1). At this phase the model developer has two options to develop the proposed model, the choice will depend on the second criteria (availability of data at the conceptual stage). The final selection is based on the hybrid model 2. Accordingly, the four key cost drivers are (P3), (P14), (P6), and (P1).

## 5.8 Limitations

According to EFA model, this model needs a sufficient data set to successfully be conducted. Furthermore, if observed variables are highly similar to each other, factor analysis will identify a single factor to them. .Moreover, this method is a complicated statistical method that requires good understanding to extract results and set its parameters.

According to regression-based models, statistically, there are several points of criticism. Wilkinson and Dallal (1981) indicated that testes are biased where there is a difference in significance level in the F-procedure as a test of forward regression method. To avoid model over fitting and over simplified model, the expert judgment may be needed to validate the selected variables and the model instead of the validation data set (Flom and Cassell, 2007).

According to correlation models, correlation cannot imply causation where a correlation between two variables is not a sufficient condition to identify a causal relationship. A correlation coefficient is not sufficient to identify the dependence structure between random variables (Mahdavi Damghani B., 2013). The strength of a linear relationship between two variables can be identified by the Pearson correlation coefficient, however its value generally does not completely characterize their relationship (Mahdavi Damghani, Babak, 2012). Therefore, there is a need for an intelligent technique to explain the causation of any relation among the studied variables.

## 5.9 Conclusion

Developing a precise parametric cost model mainly depends on the key cost drivers of the project at early stages of the project life cycle. Therefore, this study presents several techniques to identify these cost drivers. The contribution of this chapter is providing more than quantitative approaches to identify key cost drivers based on statistical methods such as EFA, regression methods, and correlation matrix. These statistical methods can be combined to develop a hybrid model to have the best subset of key cost drivers. The final key cost drivers are total length (P3), year (P14), number of irrigation valves (P6) and area served (P1). These parameters are extracted by hybrid model 2 where Pearson correlation matrix scanning and stepwise method are used to filter independent variables. Accordingly, the next chapter to illustrate how these drivers can be applied to develop a precise cost model.

# CHAPTER 6

# MODEL DEVELOPMENT

## 6.1 Introduction

The purpose of this chapter is to develop a reliable cost estimating model to be used in early cost estimation. This research consists of six steps, the first step is to review the past literature. The second step includes data collection of real historical construction FCIPs whereas the third step includes a model development. The fourth step is to select the most accurate model based on the coefficient of determination ($R^2$) and Mean Absolute Percentage Error (MAPE). The fifth step focused on model validity by comparing the model results of 33 cases to compute MAPE for the selected models. The sixth step is to conduct sensitivity analysis to determine the contribution of each key parameter on the total cost of FCIPs.

## 6.2 Data Collection

Once the inputs (most important variables) and the output (preliminary construction cost of FCIP) are identified, relevant data are collected for each historical case to develop the parametric cost model. The quantity and quality of the historical cases are necessary for the conceptual estimation that affect the performance of the model (Bode, 2000). Neural networks models need many historical data to give a good performance. Consequently, more collected data means better generalization. Therefore, 144 FCIPs between 2010 and 2015 have been collected. This collected sample has been divided to training sample (111 cases) and testing sample (33) for validation purposes for the selected model. These data have been collected from irrigation improvement project sector of Egyptian ministry of water resources and irrigation. Appendix D (Data for key cost drivers) illustrates the used training data (111 cases) and validation data (33 cases).

## 6.3 Multiple Regression Analysis (MRA)

A parametric cost estimating model consists of one or more functions and cost-estimating relationships, between the cost as the dependent variable and the cost-governing factors as the independent variables. Traditionally, cost-estimating relationships are developed by applying regression analysis to historical project information. The main idea of regression analysis is to fit the given data while minimizing the sum of squared error and maximizing the coefficient of determination ($R^2$). There is always a problem of determining the class of relations between parameters and project costs (Hegazy and Ayed, 1998). The next step is to apply regression analysis on the only four key parameters: total PVC pipeline length, command area, construction years, the number of irrigation valves. The $R^2$ values are used as a model accuracy criterion.

### 6.3.1 Sample Size

Miles and Shevlin (2001) has classified the required samples based on its effect for small, medium and large effect depending on the number of predictors.

Green (1991) stated two rules for the minimum acceptable sample size, the first based on the overall fit of the regression model (i.e. Test the $R^2$), and the second based on individual predictors within the model (i.e. test b-values of the model). Green (1991) has recommended a minimum sample size of 50 + 8k, where k is the number of predictors. For example, with four predictors, the required sample size would be 50 + 32 = 82. For the individual predictors, it suggested a minimum sample size of 104 + k. For example, with 4 predictors, the required sample size would be 104 + 4 = 108 and in this situation, Green recommended to calculate both of the minimum sample sizes and use the one that has the largest value. Therefore, in the present case study, the collected sample is 111 cases more than 108 cases that would be sufficiently acceptable to develop the regression model. Appendix D (Data for key cost drivers) illustrates the used training data (111 cases) and validation data (33 cases).

### 6.3.2 Regression models using transformed data

Many researchers conducted transformed models where these models produce more accurate results than standard linear regression. (Stoy et al, 2008) used semilog model to predict the cost of residential construction where the MAPE for the semilog model (9.6%) was more than linear regression model (9.7%). The previous result proved that semilog models may produce a more accurate model that plain regression model. However, this is not a rule, in other words, plain regression models may produce more accurate and simple models than transformed models.

It is vital to select the appropriate response variable at the commencement of model development which leads to a reliable and stable model in the conclusion Lowe et al. (2006). EMSLEY et al, (2002) have conducted two approaches for modelling, predicting cost per m$^2$ and log of cost per m$^2$. The results had similar performance with small differences and predicting the cost per m$^2$ tends to develop a model with a higher $R^2$ value than the cost per m$^2$ terms. However, the log model yields lower values of MAPE. The function of the log model is to minimize proportional differences, whereas the untransformed cost per m$^2$ model minimizes the square of the error on the cost per m$^2$.

Stoy et al, (2012) have performed a semilog regression model to develop cost models for German residential building projects. The most significant variables for the cost of external walls were determined by backward regression method. These predictors were compactness, the percentage of openings and height of the building. The detection of multicollinearity and singularity problems were investigated. The proposed model was 7.55% prediction accuracy for the selected population.

Lowe et al, (2006) have developed linear regression models and neural network model to predict the construction cost of buildings based on 286 past cases of data collected in the United Kingdom. The raw cost was rejected as a suitable dependent variable and models were developed for three alternatives: cost/m2, the log of cost variable, and the log of cost/m$^2$. Both forward and backward stepwise regression analyses were applied to develop a total of six models. The best regression model was the log of cost backward model which gave an R$^2$ of 0.661 and a MAPE of 19.3%. The best neural network model was one which uses all variables; this gave an R$^2$ value of 0.789 and a MAPE of 16.6%. Love et al, (2005) have developed a Logarithmic regression model to examine the project time–cost relationship by using project scope factors as predictors for 161 building construction in 2000. The project type was included in the analysis by using a dummy variable system. Projects in various Australian States have performed a transformed regression model (semilog) to estimate a building cost index based on historical construction projects in several markets where the key parameters were the number of stories, absolute size, a number of units, frame type, and year of construction (Wheaton and Simonton, 2007).

(Williams, 2002) has developed the neural network and regression models to estimate the completed cost of competitively bid highway projects constructed by the New Jersey Department of Transportation. A natural log transformation of the data was performed to improve the linear relationship between the low bid and completed cost. The stepwise regression procedure was used to select predictors. Radial basis neural networks were also constructed to predict the final cost the best performing regression model produced more accurate predictions to the best performing neural network model. This study also used hybrid models where regression model prediction have been used as an input to a neural network.

**Table *6*.1. Transformed regression models.**

| Model | Method | Transformation | $R^2$ | $R^2$ adjusted | F value | MAPE |
|-------|--------|----------------|-------|----------------|---------|------|
| **1** | Standard Linear regression | None | 0.857 | 0.851 | 158.5 | 9.13% |
| **2** | Quadratic Model | Dependent Variable =Sqrt(y) | 0.863 | 0.857 | 167.1 | 9.13% |
| **3** | Reciprocal Model | Dependent Variable =1/y | 0.803 | 0.796 | 108 | 11.20% |
| **4** | Semilog Model | Dependent Variable =LIN(y) | 0.857 | 0.851 | 158.5 | 9.30% |
| **5** | Power Model | Dependent Variable =(y)$^2$ | 0.814 | 0.801 | 115 | 11.79% |

Table 6.1 Summaries Transformed regression models. The regression model was developed using the software SPSS 19. The accuracy of the model is tested from two perspectives: MAPE (Equation.6.1) and coefficients of determination ($R^2$).

$$\text{MAPE} = \frac{1}{n} \sum_{i=1}^{n} \frac{|\text{actual i} - \text{predicted i}|}{\text{predicted i}} \times 100 \qquad (6.1)$$

This study has applied five models: standard linear regression, quadratic model, reciprocal model, semilog model and power model where the most accurate model is Quadratic Model. Quadratic Model is a dependent variable transformation by taking the square root (sqrt). Additionally, this model showed $R^2$ of 0.863, $R^2$ adjusted of 0.857 and a MAPE of 9.13%. Accordingly, the quality of the developed quadratic regression model can be classified as good. The shape of probability distribution plays a vital role in statistical modeling (Tabachnick and Fidell, 2007). Normality assumption has to be fulfilled to perform linear models where this enables error terms to be distributed normally (Tabachnick and Fidell, 2007). The histogram for the total cost of FCIP is positively skewed as shown in Fig.6.1. Root square transformation is employed for the dependent variable that results in inducing symmetry and reducing skewness.

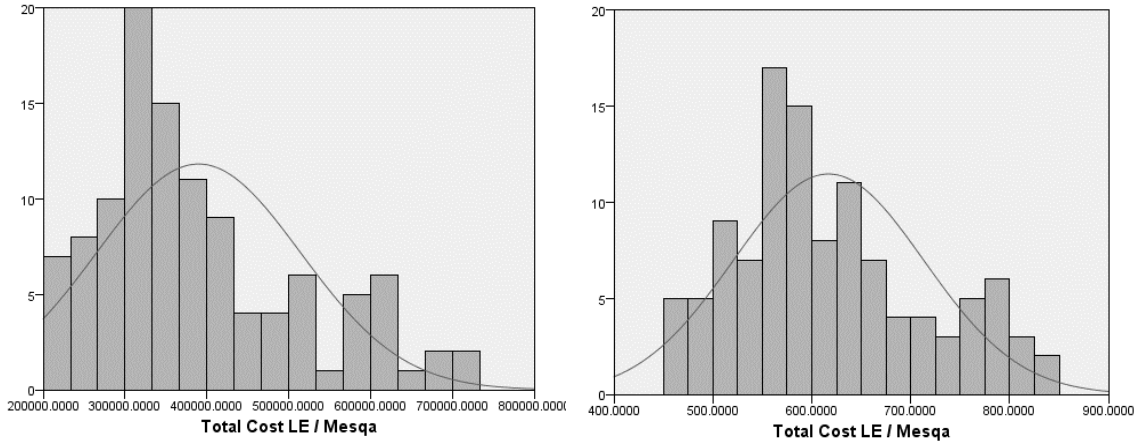

**Fig.6.1 The left chart is the probability distribution for untransformed regression model (standard linear regression). The right chart is the probability distribution for quadratic model regression model (dependent variable = sqrt(y)).**

### 6.3.3 Deleting Outliers

According to deleting outliers, Cook's distance is a measure of the overall influence of a case on the regression model where Cook and Weisberg (1982) suggested that values greater than 1 may be cause for concern. If Cook's distance is < 1, there is no need to delete that case because it does not have a large effect on regression analysis (Stevens, 2002). The current regression analysis, there is no need to delete any case where the Cook's distance is 0.249< 1.

### 6.3.4 Multicollinearity and singularity

Multicollinearity is a strong correlation between two or more predictors in a regression model. Multicollinearity and singularity are problems that occur when variables are too highly correlated. Multicollinearity is that variables are correlated too highly and singularity is that variables are perfectly correlated. Multicollinearity means that the b-values are less trustworthy (Field, 2009). Rockwell, (1975) stated that variables are high correlated at (r > 0.8). In the present study, by scanning a correlation matrix of all variables, no correlated variables are above 0.80 or 0.90. The variance inflation factor (VIF) indicates whether a variable has a strong linear relationship with the other variable(s) (Field, 2009). Myers (1990) suggested that a value of 10 is a good value to cause for concern. If average VIF is greater than 1, then multicollinearity may be biasing the regression model (Bowerman & O'Connell, 1990). Menard (1995) suggested that the tolerance statistic below 0.2 are worthy of concern. In the current study after calculating the variance inflation factor (VIF), the values of tolerance are 0.576, 0.567, 0.607 and 0.977 where no multicollinearity occurs.

### 6.3.5 Durbin and Watson Test

To avoid biased regression model, the assumption of homoscedasticity should be met. Homoscedasticity is the same variance of the residual terms where the residual variance should be constant. Heteroscedasticity is high unequal residual variances the residual terms should be uncorrelated (Field, 2009). To test these assumptions, the Durbin–Watson test is applied to test for serial correlations between errors, the test statistic can vary between 0 and 4 where the value of 2 meaning that the residuals are uncorrelated (Field, 2009). Values less than 1 or greater than 3 are definitely cause for concern (Durbin and Watson's, 1951). In the current study, Durbin-Watson is 2.224 where this value is causing no problem.

### 6.3.6 Quantification of causal relationships (model causality)

**Table *6*.2. Coefficient Table of model 1 where dependent variable is FCIP cost**

| Parameters | Unstandardized Coefficients | Std. Error | Standardized Coefficients | t | Sig. | Collinearity Statistics | VIF |
|---|---|---|---|---|---|---|---|
| | B | | Beta | | | Tolerance | |
| (Constant) | -37032.81 | 4851.21 | | -7.63 | 0.00 | | |
| Area served(P1) | 0.93 | 0.25 | 0.18 | 3.72 | 0.00 | 0.57 | 1.76 |
| Total length (P3) | 0.17 | 0.01 | 0.66 | 14.01 | 0.00 | 0.58 | 1.73 |
| Irrigation valves number(P6) | 5.27 | 1.26 | 0.19 | 4.17 | 0.00 | 0.61 | 1.65 |
| year(P14) | 18.59 | 2.41 | 0.28 | 7.72 | 0.00 | 0.98 | 1.02 |

As shown Table 6.2, the regression model (model 1) can incorporate four independent variables: Area served (P1); Total length (P3); Irrigation valves number (P6); year (P14). This quadratic regression model can be represented by the following equation (Equation.6.2):

$$(Y)^{0.5} = -37032.81 + 2.21X_1 + 0.1691X_2 + 2.265X_3 + 18.594X_4 \qquad (6.2)$$

Where $y$: FCIP cost LE / mesqa;

$X_1$ (hectare): Area served (P1);

$X_2$ (meter): Total length (P3);

$X_3$ (unit): Irrigation valves number (P6);

$X_4$ : year (P14).

The causal relationships of each independent variable of the proposed model are described in the following section. It begins with (P1) whose direct impact on FCIP construction costs is represented in Fig.6.2. (P1) also exhibits a positive cost impact. That means the square root (Sqrt) of FCIP construction costs rise with an increasing area served based on Equation.6.2 where an additional increase by one hectare in the area served rises the square root of FCIP construction costs by 2.21 LE.

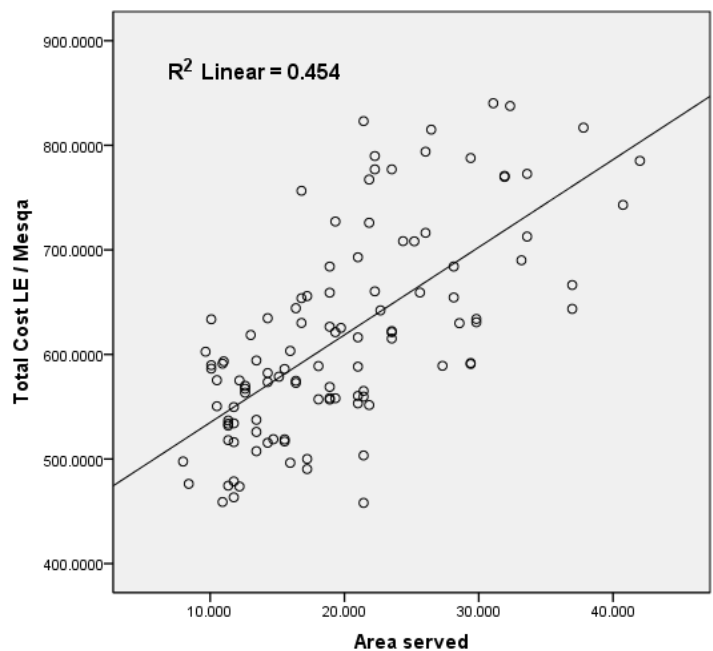

**Fig. 6.2. Total construction costs of FCIP and area served (R = 0.454).**

According to (P3), Fig.6.3 also exhibits a positive cost impact. That means the square root (Sqrt) of FCIP construction costs rise with an increasing total length based on Equation 6.2, an additional increase by one meter in the total length of PVC pipeline rises the Sqrt FCIP construction costs by 0.1691 LE.

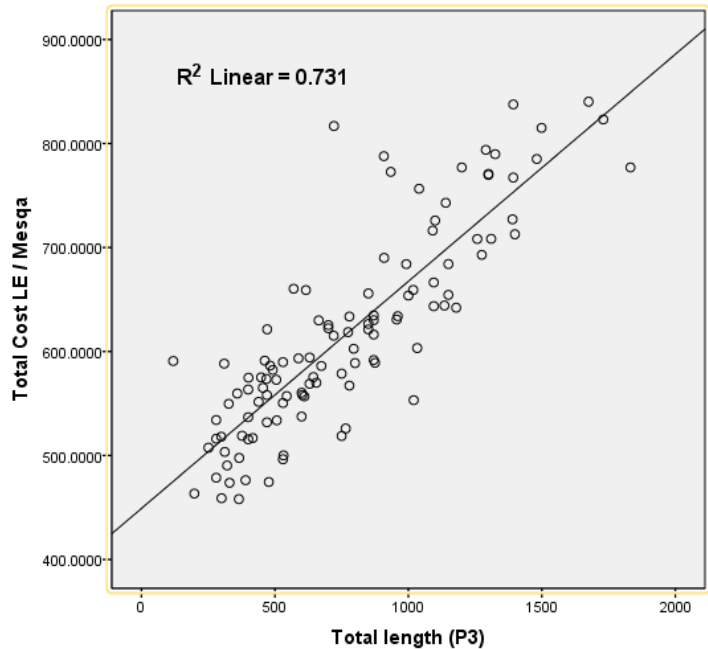

**Fig. 6.3. Total construction costs of FCIP and total length (R = 0.649).**

According (P6), Fig. 6.4 also exhibits a positive cost impact. That means the square root (sqrt) of FCIP construction costs rise with an increasing number of Irrigation valves. Based on Equation.6.2, an additional increase by one valve rises the square root of FCIP construction costs by 2.265 LE.

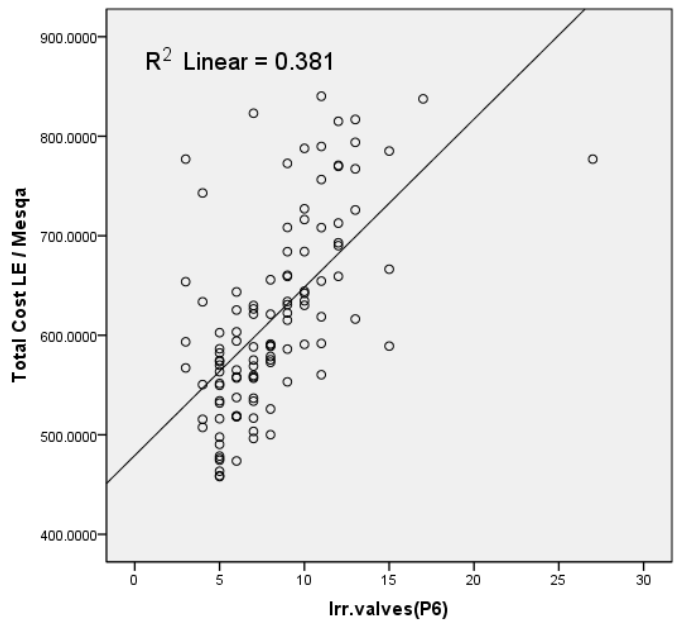

**Fig. 6.4. Total costs of FCIP and irrigation valves number (R = 0.381).**

According to (P14), Fig. 6.5 also exhibits a positive cost impact. That means the square root (Sqrt) of FCIP construction costs rise with an increasing construction year. Based on Equation.6.2, an additional increase by one year rises the square root of FCIP construction costs by 18.594.

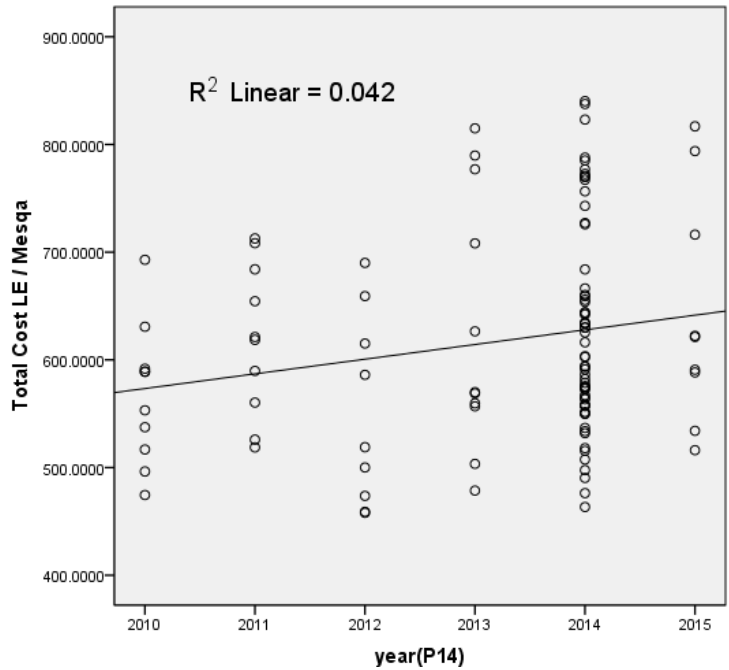

**Fig. 6.5. Total construction costs of FCIP and construction year (R = 0.042).**

## 6.4 Artificial Neural Network (ANNs) Model

ANNs is a computational method based on human brain working. ANNs consists of a group of nodes arranged in three following layers called an input layer, a hidden layer, and an output layer. Input layer nodes are used to receive input parameters of the model, hidden nodes are used to connect and develop the relation between input and output layer. The output layer is used to produce the final result of the model which is the conceptual cost of FCIPs in the current study. Each node in the hidden and the output layer calculates a sum product of the coming nodes with its corresponding weights and sends the result to the following node in the next layer. The major advantage of ANNs is their ability to fit nonlinear data, learn from past data and to generalize that knowledge to similar cases (Hegazy and Ayed, 1998). Using an appropriately configured NN model and a sufficient set of historical data, ANNs model would be able to arrive at the accurate prediction of the cost of a new construction FCIPs.

Williams (2002) has applied ANNs by various combination of input transformations to develop a cost model for Predicting completed project cost using bidding data. Table.6.3 shows three neural network models that are studied, and their performance in predicting the total cost of FCIP. The first model is untransformed model whereas the second model is transformed by the square root of the completed project cost. The third model is transformed by the natural log of the completed project cost. The results of MAPE showed that the ability of three models for prediction where the results are similar.

**Table.*6*.3. the three neural network models.**

| Model | Transformation | MAPE |
|---|---|---|
| **Model one** | None | 9.27% |
| **Model two** | Dependent Variable =Sqrt(y) | 9.20% |
| **Model three** | Dependent Variable =LIN(y) | 10.23% |

After many experiments by SPSS.19 software, this study has performed an ANNs model with structure (4-5-0-1) where four represents number of inputs (four key parameters), five represents the number of hidden nodes in the first hidden layer, zero means no second hidden layer used and one represents one node to produce the total cost of the FCIPs as illustrated in Fig.6.6. This model has a MAPE (9.20%). the type of training is batch, the learning algorithm is scaled conjugate gradient and the activation function is hyperbolic tangent. Appendix D1 (Training data) illustrates the used training data.

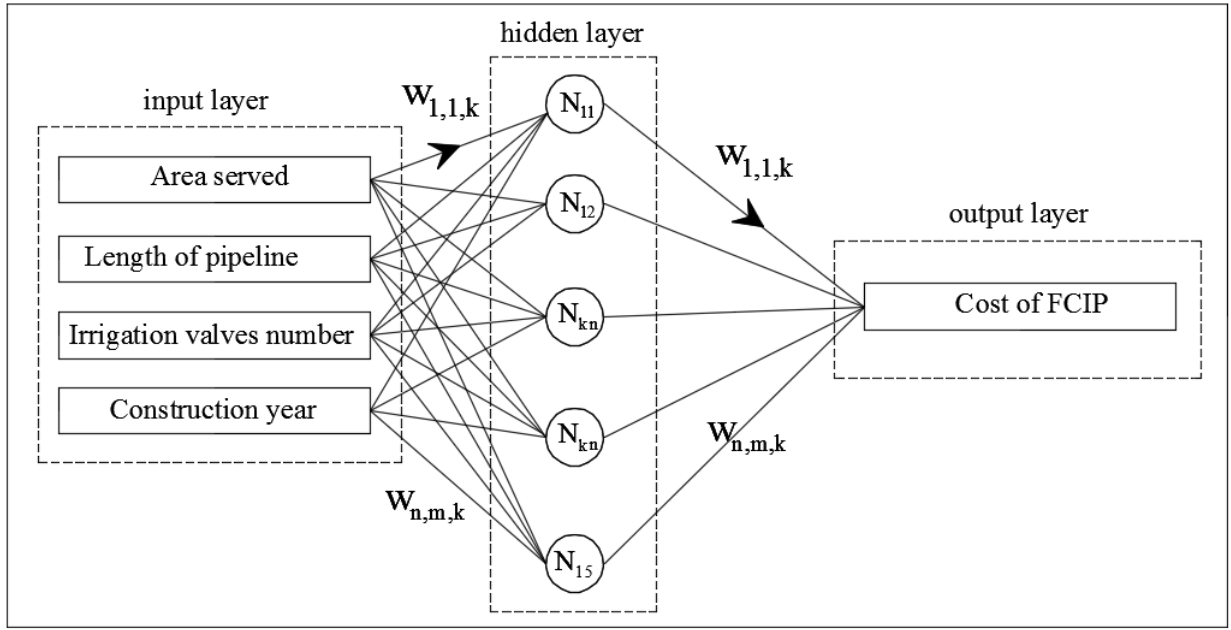

**Fig. 6.6. The structure of ANNs model.**

## 6.5 CBR model

Case-based reasoning (CBR) is a sustained learning and incremental approach that solves problems by searching the most similar past case and reusing it for the new problem situation (Aamodt and Plaza, 1994). Therefore, CBR mimics a human problem solving (Ross, 1989; Kolodner, 1992). CBR is a cyclic process learning from past cases to solve a new case. The main processes of CBR are retrieving, reusing, revising and retaining. Retrieving process is solving a new case by retrieving the past cases. The case can be defined by key attributes. Such attributes are used to retrieve the most similar case, whereas, reuse process is utilizing the new case information to solve the problem. Revise process is evaluating the suggested solution for the problem. Finally, retain process is to update the stored past cases with such new case by incorporating the new case to the existing case-base (Aamodt and Plaza, 1994).

| Case Base Reasoning (CBR) model | | | | | | | | | | | | | | |
|---|---|---|---|---|---|---|---|---|---|---|---|---|---|---|
| Case Attributes | | | | | | Attribute Similarity (AS) | | | | Attribute Weights (AW) | | | | Case similarity |
| SN | Area surved(P1) | Total length (P3) | Irr.valve s(P6) | year(P14) | Total Cost LE / Mesqa | AS (P1) | AS (P1) | AS (P1) | AS (P1) | W (P1) | W (P1) | W (P1) | W (P1) | CS |
| New cas | 24.00 | 779.00 | 4.00 | 2014.0 | 303024.42 | | | | | | | | | 0.93 |
| 1 | 25.00 | 530.00 | 4.00 | 2014.00 | 303024.42 | 0.96 | 0.68 | 1.00 | 1.00 | 0.20 | 0.20 | 0.20 | 0.40 | 0.93 |
| 2 | 30.00 | 779.00 | 3.00 | 2014.00 | 321716.52 | 0.80 | 1.00 | 0.75 | 1.00 | 0.20 | 0.20 | 0.20 | 0.40 | 0.91 |
| 31 | 30.00 | 655.00 | 5.00 | 2013.00 | 324865.50 | 0.80 | 0.84 | 0.80 | 1.00 | 0.20 | 0.20 | 0.20 | 0.40 | 0.89 |
| 51 | 26.20 | 588.00 | 3.00 | 2014.00 | 352094.48 | 0.92 | 0.75 | 0.75 | 1.00 | 0.20 | 0.20 | 0.20 | 0.40 | 0.88 |
| 3 | 24.00 | 482.00 | 5.00 | 2014.00 | 343826.61 | 1.00 | 0.62 | 0.80 | 1.00 | 0.20 | 0.20 | 0.20 | 0.40 | 0.88 |
| 28 | 27.00 | 477.00 | 5.00 | 2010.00 | 225198.20 | 0.89 | 0.61 | 0.80 | 1.00 | 0.20 | 0.20 | 0.20 | 0.40 | 0.86 |
| 50 | 27.00 | 470.00 | 5.00 | 2014.00 | 282934.27 | 0.89 | 0.60 | 0.80 | 1.00 | 0.20 | 0.20 | 0.20 | 0.40 | 0.86 |
| 6 | 25.00 | 644.00 | 8.00 | 2014.00 | 331075.61 | 0.96 | 0.83 | 0.50 | 1.00 | 0.20 | 0.20 | 0.20 | 0.40 | 0.86 |
| 2 | 37.00 | 750.00 | 6.00 | 2011.00 | 269103.14 | 0.65 | 0.96 | 0.67 | 1.00 | 0.20 | 0.20 | 0.20 | 0.40 | 0.85 |
| 33 | 32.00 | 765.00 | 8.00 | 2011.00 | 276584.04 | 0.75 | 0.98 | 0.50 | 1.00 | 0.20 | 0.20 | 0.20 | 0.40 | 0.85 |
| 30 | 32.00 | 630.00 | 6.00 | 2014.00 | 353089.18 | 0.75 | 0.81 | 0.67 | 1.00 | 0.20 | 0.20 | 0.20 | 0.40 | 0.85 |
| 55 | 34.00 | 400.00 | 4.00 | 2014.00 | 265774.75 | 0.71 | 0.51 | 1.00 | 1.00 | 0.20 | 0.20 | 0.20 | 0.40 | 0.84 |
| 72 | 32.00 | 600.00 | 6.00 | 2010.00 | 288882.06 | 0.75 | 0.77 | 0.67 | 1.00 | 0.20 | 0.20 | 0.20 | 0.40 | 0.84 |
| 37 | 20.00 | 390.00 | 5.00 | 2014.00 | 226870.10 | 0.83 | 0.50 | 0.80 | 1.00 | 0.20 | 0.20 | 0.20 | 0.40 | 0.83 |
| 5 | 40.00 | 1000.00 | 3.00 | 2014.00 | 427421.32 | 0.60 | 0.78 | 0.75 | 1.00 | 0.20 | 0.20 | 0.20 | 0.40 | 0.83 |
| 14 | 31.00 | 774.00 | 11.00 | 2011.00 | 382644.52 | 0.77 | 0.99 | 0.36 | 1.00 | 0.20 | 0.20 | 0.20 | 0.40 | 0.83 |
| 29 | 30.00 | 400.00 | 5.00 | 2014.00 | 317470.28 | 0.80 | 0.51 | 0.80 | 1.00 | 0.20 | 0.20 | 0.20 | 0.40 | 0.82 |
| 73 | 34.00 | 468.00 | 5.00 | 2014.00 | 329058.72 | 0.71 | 0.60 | 0.80 | 1.00 | 0.20 | 0.20 | 0.20 | 0.40 | 0.82 |

**Fig. 6.7. CBR model for cost prediction of FCIP.**

In the present study, a CBR is developed to predict the cost of FCIP based on similarity attribute of the entered case comparable with the stored cases. As illustrated in Fig.6.7, the user or cost engineer enters the case attributes (P1, P3, P6 and P14). Once attributes are entered, attributes similarities (AS) can be computed based on (equation (6.3); (Kim & Kang, 2004)) and case similarity (CS) can be computed by (equation (6.4); (Perera and Waston, 1998).) depend on AS and attribute weights (AW). AW are selected by expert to emphasis the existence and importance of the case attributes. After validation process, the CBR model produces 17.3% MAPE which is acceptable accuracy.

$$AS = \frac{Min(AV\ new - case,\ AV\ retrived - case)}{Max(AV\ new - case,\ AV\ retrived - case)} \qquad (6.3)$$

Where AS = Attribute Similarity, $AV_{new\text{-}case}$ = Attribute Value of new entered Case, $AV_{retrieved\text{-}case}$ = Attribute Value of retrieved case.

$$CS = \frac{\sum_{i=1}^{n}(AS_i\ *\ AW_i)}{\sum_{i=1}^{n}(AW_i)} \qquad (6.4)$$

Where CS = Case Similarity, AS = Attribute Similarity, AW = Attribute Weight.

## 6.7 Model selection and Validation

By reviewing results in Table.6.1, Table.6.3 and the developed CBR model, the most accurate model is regression quadratic model (Dependent Variable =Sqrt(y)) where correlation coefficient is 0.86 and MAPE is 9.12 % for training sets. The next step is to validate that model for this study, 33 cases are extracted from the FCIPs historical data for validation. The MAPE for validation sets is 7.82 % where 20% is an accepted MAPE for the conceptual cost estimate (Peurifoy and Oberlender, 2002). Therefore, it is concluded that the regression model is suited to the present case study with acceptable MAPE for both training and validation. Appendix D2 (Validation data) illustrates the used validation data.

## 6.8 Sensitivity Analysis

Sensitivity analysis is a method that discovers the cause and effect relationship between input and output variables of the proposed model. A sensitivity analysis is then carried out in order to assess the contribution of each parameter to model's performance. As illustrated in Fig.6.7, the sensitivity analysis graphs indicate that PVC pipeline length parameter has the highest impact on the final cost of FCIPs. Area served, construction year and the number of irrigation valves have the high significant on the total cost of FCIPs where the irrigation valves number has higher significant than construction year and command area. The number of irrigation valves is more affecting the FCIPs cost more than the command area. Area served has a relatively weak impact, which may be due to the presence of the pipeline total length parameter.

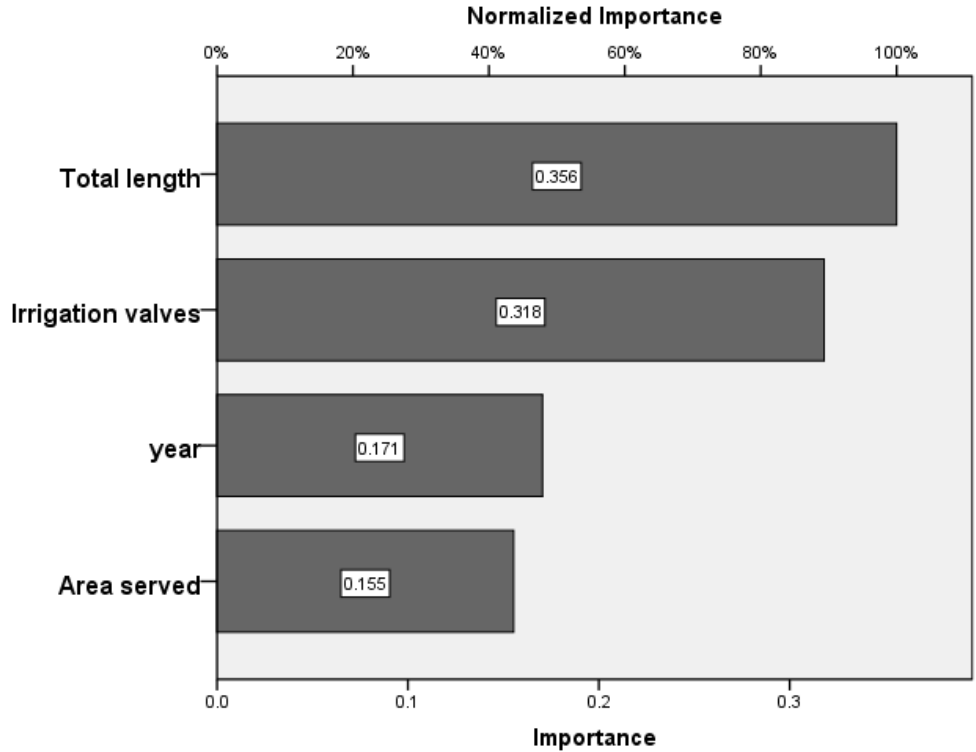

**Fig.6.7 Independent variable importance for key cost drivers by SPSS.**

## 6.9 Project data input screen for the model

To facilitate implementation process and use of the developed model, a Microsoft Excel 2013 and Excel Visual Basic are used that are easy to use, flexible and powerful to create a user-friendly interface. Appendix E (Excel VBA Code) illustrates the used code to develop the application. As illustrated in Fig 6.10, a real case study will be discussed in the next section. This user interface is the way that the model accepts new instructions by the user and presents the results.

In addition to the model, the inflation rate for the time is also added to the developed model, as using cost information of a previous project to predict the cost in the future will not be reliable unless an adjustment is made proportional to the difference in time. The future cost estimation is calculated as follows Equation (6.5) where the inflation rate can be obtained via World Bank web site:

**Future cost (LE/FCIPs) =Predicted cost (LE / FCIPs) × (1+i)$^{n}$**     (6.5)

Where:
i: The average inflation rate for the period (2015 to the future year).

n: The number of years from 2015 to the future time.

A sensitivity analysis application has been incorporated in the model for key parameters manipulation to obtain uncertainty case when the user have no certain information about a defined parameter. This application produces 30 random scenarios where the checked parameters are varied randomly in the range of 25% below or above its initial value and bounded by the maximum and minimum limits.

For example, the total pipeline length is checked as shown in Fig.6.9, as a result, an automatically sensitivity analysis results have been produced as shown in Fig.6.10.This approach has been followed by (Hegazy and Ayed, 1998).

**Fig. 6.9. Project data input screen for the parametric model by Visual basic.**

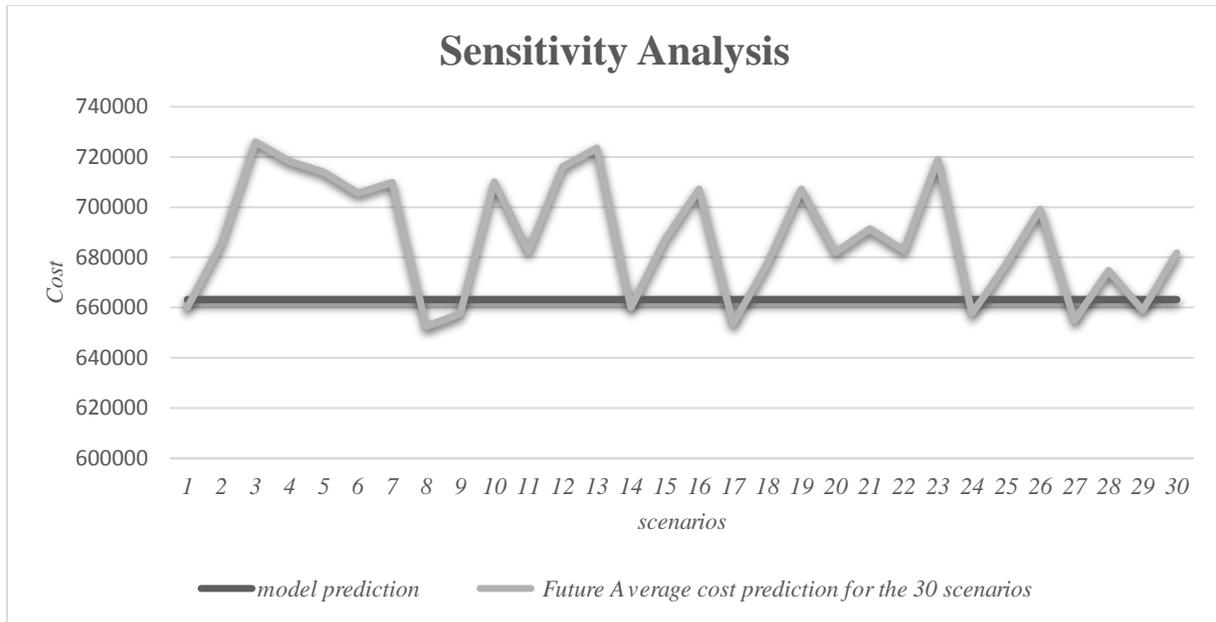

**Fig. 6.10. a sensitivity analysis application by MS Excel spreadsheet.**

## 6.10 A real case study in Egypt

To explain the usage of the developed model, a real case study in Egypt is selected to run the model. The case study is presented in Fig.1.1 in chapter 1 where the command area is 19.6 hectares, the total PVC length is 453 m, the irrigation valves number is 6 valves and the selected future year for the prediction is by 2020. Frist, the user can enter values of key parameters as illustrated in Fig.6.8 into the input screen, then enter the future year of construction. If the selected year is after 2015, inflation rate must be entered to calculate the future cost based on the Equation.6.3.

After click on "predict the total cost of FCIP" button, two values is displayed where "663209" represents the total predicted cost of FCIP and "33837" is the total cost divided by the command area value in hectares. Second, to implement uncertainty where the user is hesitating of a certain parameter, the user can check on a parameter or more than one parameter. In the present case study, "Total pipeline length" is checked to apply uncertainty. Accordingly, the average of the 30 scenarios (691725LE / FCIP) and standard deviation (18631 LE / FCIP) is automatically calculated as shown in Fig.6.8 and is represented in a sensitivity analysis chart as illustrated in Fig. 6.10.

## 6.11 Conclusion

This chapter aims to develop a reliable and practical model for conceptual cost estimating that can be used by organizations involved in the planning and construction of irrigation improvement projects such as FCIPs. The research has used four parameters as key parameters that have the most influence on the costs of constructing FCIPs. Data are transformed to produce various regression models where these models are compared based on MAPE. The ANNs model is designed with four nodes in the input layer while the output layer consists of one node representing the total cost of FCIPs. After multiple regression models and ANNs models are developed, the best model is the quadratic transformed model where dependent variable is transformed by the square root. The MAPE is 9.12% and 7.82% for training and validation respectively, and the correlation coefficient is (0.86). To facilitate the usage of the model, a user friendly input screen has been developed to receive inputs from the user and to maintain uncertainty and model manipulation, a sensitivity analysis application has been incorporated in the developed model.

# CHAPTER 7

# AUTOMATED FUZZY RULES GENERATION MODEL

## 7.1 Introduction

Fuzzy systems have the ability to model numerous applications and to solve many kinds of problems with uncertainty nature such as cost prediction modeling. However, traditional fuzzy modeling cannot capture any king of learning or adoption which formulates a problem in fuzzy rules generation. Therefore, hybrid fuzzy models can be conducted to automatically generate fuzzy rules and optimally adjust membership functions (MFs). This study has reviewed two types of hybrid fuzzy models: neuro-fuzzy and evolutionally fuzzy modeling. Moreover, a case study is applied to compare the accuracy and performance of traditional fuzzy model and hybrid fuzzy model for cost prediction where the results show a superior performance of hybrid fuzzy model than traditional fuzzy model.

At the conceptual stage of the project, cost prediction is a critical process where crucial decisions about the project depend on it and limited information about the project is available (Hegazy and Ayed, 1998). Conceptual cost estimate occurs at 0% to 2% of project completion for conceptual screening. Capacity factored model, analog model and parametric model are conducted to perform such conceptual estimate where its accuracy varies from -50% to +100% (AACE, 2004). Similarly, study cost estimate occurs at 1% to 15% of project completion for feasibility study based on parametric model where its accuracy varies from -30% to +50% (AACE, 2004). Parametric cost modeling is creating a model based on key cost drivers extracted from experts' experience or the collected past cases by conducting statistical analyses such as regression models, artificial neural networks(ANNs) and fuzzy logic (FL) model (Dell'Isola, 2002). The scope of this study is FL modeling and hybrid fuzzy modeling for parametric cost estimate.

Fuzzy logic (FL) and expert systems are widely modeling techniques used for engineering modeling where fuzzy modeling represents a promising trend for many engineering aspects such as cost prediction. FL depends on Fuzzy set theory to provide uncertainty nature to the studied case and to deal with the imprecision existed in the studied system. A FL model provides a flexible approach to solve the imprecise nature of many variables affecting on the project where the measurement of the actual data for such variables may be not available.

The objective of this study is to criticize the traditional fuzzy modeling (TFM) and to highlight the application of the hybrid fuzzy model (HFM). The TFM has many limitations such as developing fuzzy rules and determining the MFs. Therefore, this study firstly reviewed the TFMs and its application in construction cost modeling. Secondly, the study has reviewed the HFMs and its applications in construction cost modeling. Thirdly, a case study has been developed based on TFM and HFM to evaluate the performance of the both techniques.

## 7.2. Traditional Fuzzy logic model

FL is to model human reasoning taking uncertainties possibilities into account where incompleteness, randomness and ignorance of data are represented in the model (Zadeh, 1965, 1973). Moreover, FL incorporates human experiential knowledge with nonlinearity and uncertainty with reasoning and inference by semantic or linguistic terms. For example, 1 stands for true, 0 stands for false what if 1/2 is existed? The answer may be that 1/2 stands for a third truth value for 'possible'. Accordingly, there are an infinite values between zero and one or true or false can be represented by fuzzy theory as MFs. Therefore, the MF ranges between zero and one and thus, all human reasoning can be converted to fuzziness terms where all truths and falseness can be partially approximated by partial truth (Siddique and Adeli, 2013; Zadeh, 1965, 1973).

No strict rule exists to define a MF where its choice is inherently problem-dependent. Triangular, trapezoidal, Gaussian and bell-shaped functions are the most common used MFs in the FL. The shape of MF depending on the parameters of the MF used, which greatly influences the performance of a fuzzy system. There are different approaches to construct MFs, such as heuristic selection (the most widely used), clustering approach, C-means clustering approach, adaptive vector quantization and self-organizing map. The shape of MF greatly impacts the performance of a fuzzy model Chi et al. (1996).

To define membership functions, concepts exist such as support set, core, singleton, cross points (or crossover points), peak point, symmetric or asymmetric membership function, left and right width. These concepts can represent mathematically as following Equation {7.1, 7.2, and 7.3} and illustrated in Fig.7.1

$$\text{Support (A)} = \{x \mid \mu A(x) > 0 \text{ and } x \in X\} \qquad (7.1)$$

$$\text{Core (A)} = \{x \mid \mu A(x) = 1 \text{ and } x \in X\} \qquad (7.2)$$

**Crossover (A) = {x |µA(x) = 0.5}**                    **(7.3)**

Where (A) is a fuzzy set in X that is represented by a MF µA (x). Such MF is related to each point in X where x in R belongs to [0, 1]. µA (x) at x characterizes the grade of membership of x in A and  µA(x) ∈ [0, 1]. On the other hand, in classical set theory, the MF is represented only two values: 0 and 1, i.e., either µA(x) =1 or µA(x) = 0 where this is written as µA(x) ∈ {0, 1}. Fig (7.1) illustrates two trapezoidal MFs A = {a1, a2, a3, a4}, and B = {b1, b2, b3, b4}.

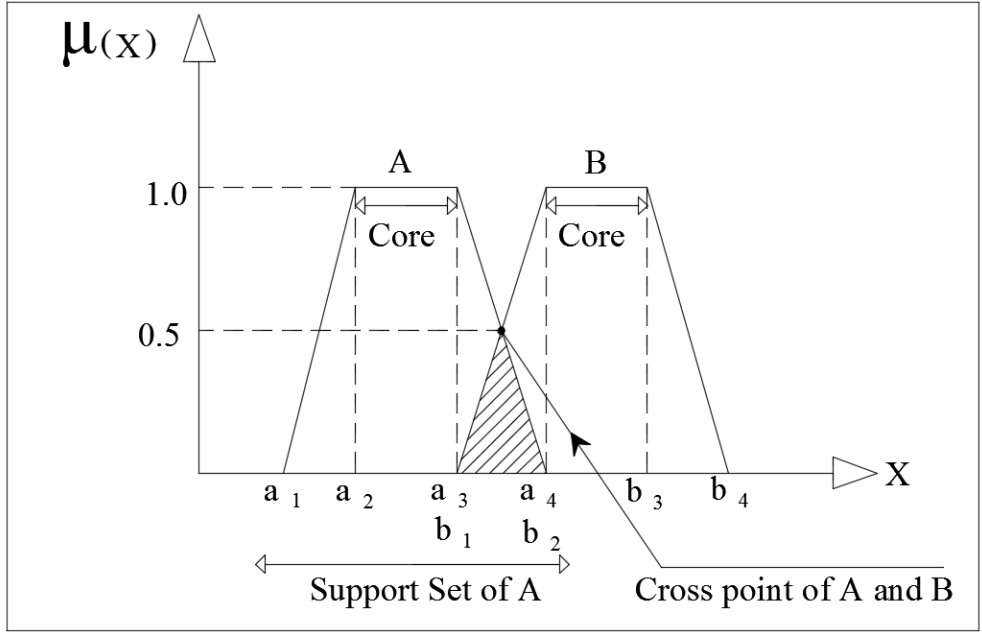

**Fig 7.1. The feature of MF (Siddique and Adeli, 2013).**

### 7.2.1 Fuzzy sets operations

Operations on fuzzy sets are union of fuzzy sets, intersection of fuzzy sets, and complement of fuzzy set and α-cut of a fuzzy set. The union of two fuzzy sets A and B with MFs µA and µB, respectively, is a fuzzy set Z, denoted Z = A ∪ B, with the membership function µZ. There are two definitions for the union operation: the max membership function and the product rule, as defined in equations (7.3) and (7.4):

**The max membership function: µZ(x) = max [µA(x), µB(x)]**          **(7.4)**

**The product rule: µZ(x) = µA(x) + µB(x) − µA(x)µB(x)**                    **(7.5)**

Where x is an element in the universe of discourse X. The intersection of two fuzzy sets A and B with MFs µA and µB, respectively, is a fuzzy set C, denoted C = A ∩

B, with MF μC defined using the min MF or the product rule as equations (7.6) and (7.7):

**The min membership function: μC(x) = min [μA(x), μB(x)]** **(7.6)**

**The product rule: μC(x) = μA(x) ∗ μB(x)** **(7.7)**

The complement of a fuzzy set A with membership function μA is a fuzzy set, denoted ∼A, with MF μ∼A defined as Equation (8)

**μA(x) = 1 − μA(x)** **(7.8)**

α-cut of a fuzzy set is a subset of X consisting of all the elements in X defined by Equation(7.9).

**Aα = {x |μAα (x) ≥ α and x ∈ X}** **(7.9)**

### 7.2.2 Linguistic variables and IF-Then fuzzy rules

Linguistic variables are labels of fuzzy subsets whose values are words or sentences (Zadie, 1976). Such linguistic terms mean approximation of system features which cannot be represented precisely by quantitative terms. For example, project cost is a linguistic variable if its values are high cost, medium cost, low cost, etc. For example, "high cost" is a linguistic term of the project cost compared with an exact numeric term of the project cost such as 'project cost is 33 million dollars'. If–Then rule statements are utilized to formulate the conditional statements that develop FL rules base system.  A single fuzzy If– Then rule can be represented as the following:

**If** <fuzzy proposition (x is A₁)> **Then** <fuzzy proposition (y is B₂)>

Where x is an input parameter and $A_1$ is a MF of x, and y is an output parameter and $B_2$ is a MF of y. Rule-based systems are systems that have more than one rule to represent human logic and experience to the developed system. Aggregation of rules is the process of developing the overall consequent from the individual consequents added by each rule (Siddique and Adeli, 2013).

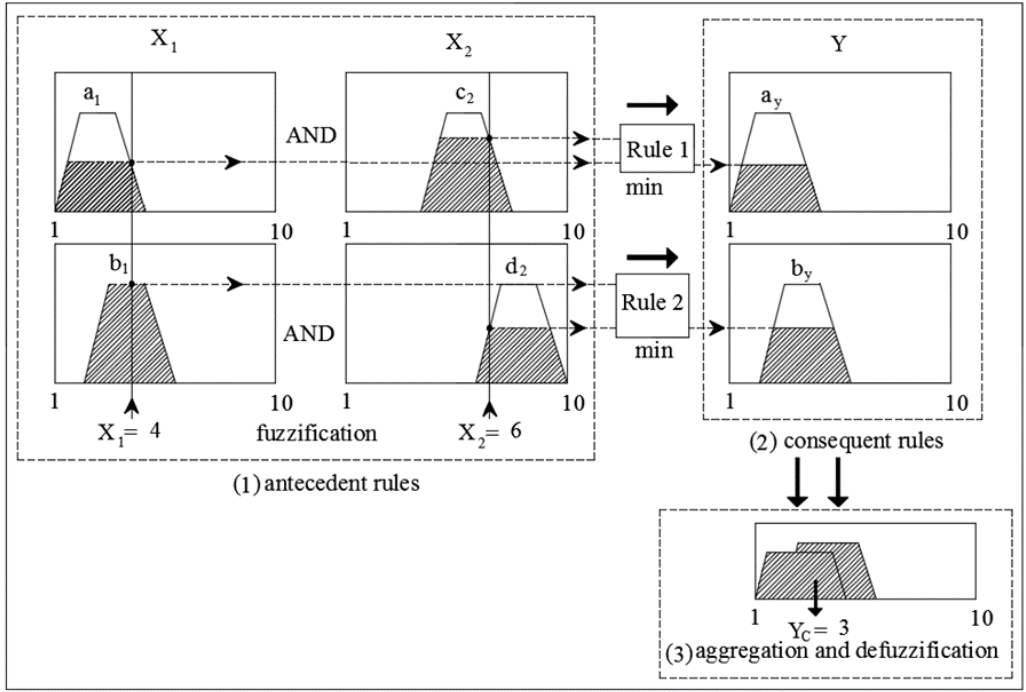

**Fig. 7.2. Fuzzy rules firing (Siddique and Adeli, 2013).**

As shown example in Fig. 7.2, there are two parameters $X_1$ and $X_2$ where

$\mu$ $X_1 = \{ a_1, b_1, c_1, d_1 \}$, $\mu$ $X_2 = \{ a_2, b_2, c_2, d_2 \}$, $\mu$ $Y = \{ a_y, b_y, c_y, d_y \}$ and the fuzzy system consists of two rules as following:

Rule 1: IF $x_1$ is small AND $x_2$ is medium THEN y is big.

Rule 2: IF $x_1$ is medium AND $x_2$ is big THEN y is small.

Where two inputs are used $\{X_1=4, X_2=6\}$. Such two inputs intersect with the antecedents MF of the two rules where two consequents rules are produced $\{R_1$ and $R_2\}$ based on minimum intersections. The consequent rules are aggregated based on maximum intersections where the final crisp value is 3. The aggregated output for $R_i$ rules are given by

Rule 1: $\mu$ $R_1$ = min [$\mu$ a1 (x1) and $\mu$ c2 (x₂)]
Rule 2: $\mu$ $R_2$ = min [$\mu$ b1 (x1) and $\mu$ d2 (x₂)]
Y: Fuzzification [max [$R_1$, $R_2$]

Fuzzification is converting a numeric value (or crisp value) into a fuzzy input. Conversely, defuzzification is the opposite process of fuzzification where the defuzzification is conversion of a fuzzy quantity into a crisp value. The shape of the MF plays a crucial role in fuzzification and defuzzification (Wang, 1997). Max-membership, centre of gravity, weighted average, mean-max, different defuzzification and centre of sums are different defuzzification methods (Runker, 1997).

Inference mechanism is the process of converting input space to output space. Three fuzzy inference mechanisms exist where the difference among them lies in the consequent parts of their fuzzy rules. These fuzzy inference mechanisms are Mamdani fuzzy inference (Mamdani and Assilian, 1974), Sugeno fuzzy inference (Takagi and Sugeno, 1985; Sugeno and Kang, 1988), and Tsukamoto fuzzy inference (Tsukamoto, 1979).

### 7.2.3 Traditional fuzzy cost model review

Many parametric cost models have been developed based on fuzzy theory to utilize uncertainty concepts for cost estimate. Based on 98 examples, (Ahiaga-Dagbui et al, 2013) have developed a cost model for water infrastructure projects where a combination of ANNs and fuzzy set theory are incorporated to develop more accurate model where MAPE was 0.8%. Based on the collected building projects, (Knight and Fayek, 2002) have developed a FL model for estimating design cost overruns on building projects with acceptable error and uncertainty considerations. (Chen and Chang, 2002) have built a FL model for wastewater treatment plants based on 48 historical cases. Based on 568 Towers, a four input fuzzy clustering model and sensitivity analysis are conducted for estimating telecommunication towers with acceptable MAPE (Marzouk and Alaraby, 2014).

Shaheen et al, (2007) have proposed the use of fuzzy numbers for cost range estimating and claimed the fuzzy numbers for fuzzy scheduling range assessment. (Yang and Xu, 2010) have developed a fuzzy model based on four inputs and one output where a set of IF-then rules, center of gravity fuzzufucation, the product inference engine and singleton fuzzifier are applied and the developed model has 3.2% error. (Shi et al, 2010) have applied index values for membership degree and exponential smoothing method to develop a construction cost model. (Shreenaath et al, 2015) have conducted a statistical fuzzy approach for prediction of construction cost overrun. Based on 60 respondents and relative important index (RII) scale, five factors are selected of 54 factors to be used as fuzzy model inputs. In addition, the model is validated by four case studies.

Papadopoulos et al, (2007) have compared linear regression model with fuzzy linear regression model for wastewater treatment plants in Greece where the results of both models are similar and reliable. (Marzouk and Amin, 2013) have developed an ANNs model for predicting construction materials prices, whereas a FL model is applied to determine the importance degree of each material for ANNs model. Such model has an acceptable accuracy in training and testing phases. A FL model is developed for satellite cost estimation. Such models works as a fuzzy expert tool for cost prediction based on two input parameters (Karatas and Ince, 2016).

A FL model is developed for building projects with acceptable error and good generalization based on 106 building projects in Gaza trip (El Sawalhi, 2012) .Moreover, FL theory is not the only technique to model the uncertainty, where Monti Carlo technique can be conducted to model uncertainty. (Moret and Einstein, 2016) have developed a Monti Carlo simulation model to model uncertainty in high-speed rail line projects, where three sources of uncertainty have been identified due to disruptive events, construction process variability and the correlations of repeated activities costs.

By surveying the past literature, the studies have developed fuzzy systems without mentioning the method of fuzzy rules generation or the fuzzy rules has been developed based on experts 'experience. Determining the fuzzy rules is the main limitation of the previous studies of fuzzy cost models. Therefore, a new trend evolves to solve this problem such as developing hybrid fuzzy modeling for the cost estimate as following.

## 7.3. Hybrid fuzzy models

### 7.3.1 Neuro-fuzzy model

ANN is utilized for solving prediction and classification problems based on given data, whereas FL is used for fuzzy prediction and fuzzy classification based on fuzzy rules. ANNs and FL can be incorporated to develop a hybrid model to find the parameters of a fuzzy system based on ANNs learning ability (Siddique and Adeli, 2013). Such a combination is used for building a rule base and determine MFs. Hybridization between neural networks and fuzzy systems can be cooperative neuro-fuzzy systems, concurrent neuro-fuzzy systems, and hybrid neuro-fuzzy systems. Two groups exist for such hybridization: cooperative FS-NN systems and cooperative NN-FS systems. This study will focus on cooperative NN-FS systems such as cooperative neuro-fuzzy systems and hybrid neuro-fuzzy systems (Siddique and Adeli, 2013).

In NN-FS cooperation is used to estimate different parameters of a fuzzy system such as rule base and MFs depend on available data, where ANNs works as a learning technique for these parameters. The objective of the NN is to optimize the performance of the fuzzy system. Takagi (1995) has applied NN and FL to customer product, where NNs have been applied to automatically design the MFs of fuzzy systems. (Yager, 1994) suggested a fuzzy production rules are driven by NN framework that system can develop the membership grades of the linguistic variables.

### 7.3.1.1 NN for estimation of MFs

Adeli and Hung (1995) proposed an algorithm to determine MFs using ANNs, where ANNs convert input patterns to output clusters. Initially, the algorithm starts with one output cluster based on all input patterns. By learning, all clusters are developed where these clusters shape the required MFs. The objective is to minimize the model performance based on the developed clusters that generate MFs. Fig(7.3) shows the neuro-fuzzy system for computing MFs. the objective is to produce MFs which minimize (E), where (E) can be calculated by Equation(7.10).

$$E = Y_{actual} - Y_{model} \qquad\qquad (7.10)$$

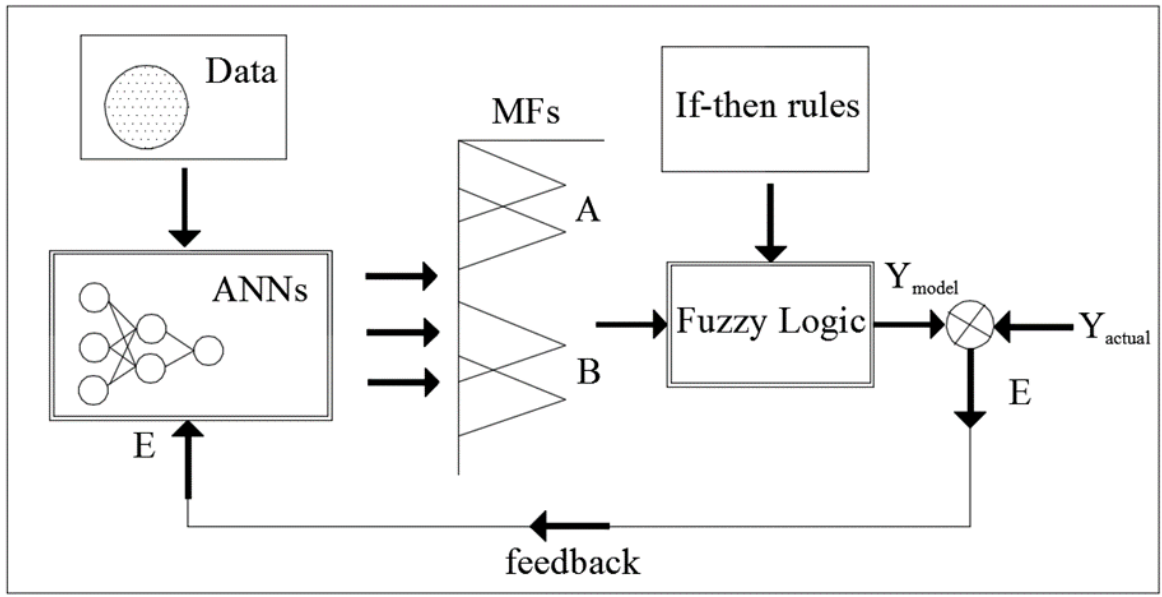

**Fig 7.3. NN for estimation of MFs (Siddique and Adeli, 2013).**

### 7.3.1.2 NN for fuzzy rules learning

ANNs solve the limitation of learning ability of the fuzzy systems. By ANNs, automatic generation of fuzzy rules can be developed based on the training data. Takagi and Hayashi (1991) have combined ANNs and fuzzy reasoning to design MFs where this approach have ability to learn and automatically determine the rules inference. The Takagi–Hayshi method consists of three steps:

1. By data clustering, convert the input data space into number of rules.
2. Define the MFs by applying NN for each rule.
3. Define the consequent value by developing NN model as the consequent function.

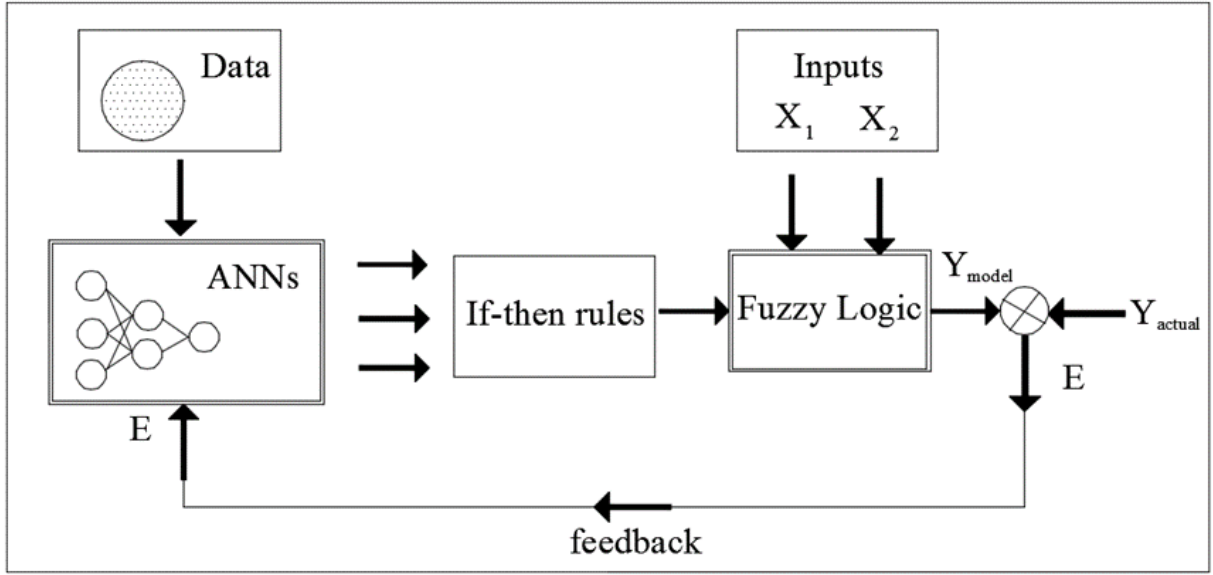

**Fig 7.4. NN for fuzzy rules learning (Siddique and Adeli, 2013).**

Fig (7.4) shows the Neuro-fuzzy system for computing fuzzy IF-Then rules. The objective is to produce fuzzy IF-Then rules which minimize E where E can be calculated by Equation (7.10). For example, If $(A_1, A_2, \ldots, A_n)$ are MFs for x where, $x = (x_1, x_2, \ldots, x_n)$ is the input vector and $Y_{model}$ is a ANNs model, the IF-then rule can be generated based on the following Equation (7.11).

**If $x_1$ is $A_1$ and $x_2$ is $A_2$, . . . , $x_n$ is $A_n$ Then y = f ($x_1, x_2, \ldots, x_n$)**     **(7.11)**

## 7.3.2 Evolutionary fuzzy Systems

Selecting the MF for each linguistic variable and developing fuzzy if-then rules are the main problems in fuzzy system (Cordon et al, 2001). Traditionally, an expert is consulted to develop such MFs and fuzzy rules or the fuzzy designer can use trial and error approach. However, such approach is time-consuming and does not guarantee the optimal MF and fuzzy rules. Moreover, the number of fuzzy if-then rules increase exponentially by increasing the number of inputs, linguistic variables or number of outputs. In addition, the experts cannot easily define all required fuzzy rules and the associated MFs. In many engineering problems, evolutionary algorithm (EA) optimization technique has been conducted to automatically develop fuzzy rules and MFs to improve the system performance (Chou, 2006; Kwon and Sudhoff, 2006). Hybridization of FL and genetic algorithm (GAs) can be genetic adaptive fuzzy systems and fuzzy adaptive genetic algorithms. EA fuzzy systems are discriminated along two main approaches: (i) Evolutionary tuning of fuzzy system, (ii) Evolutionary learning of fuzzy system.

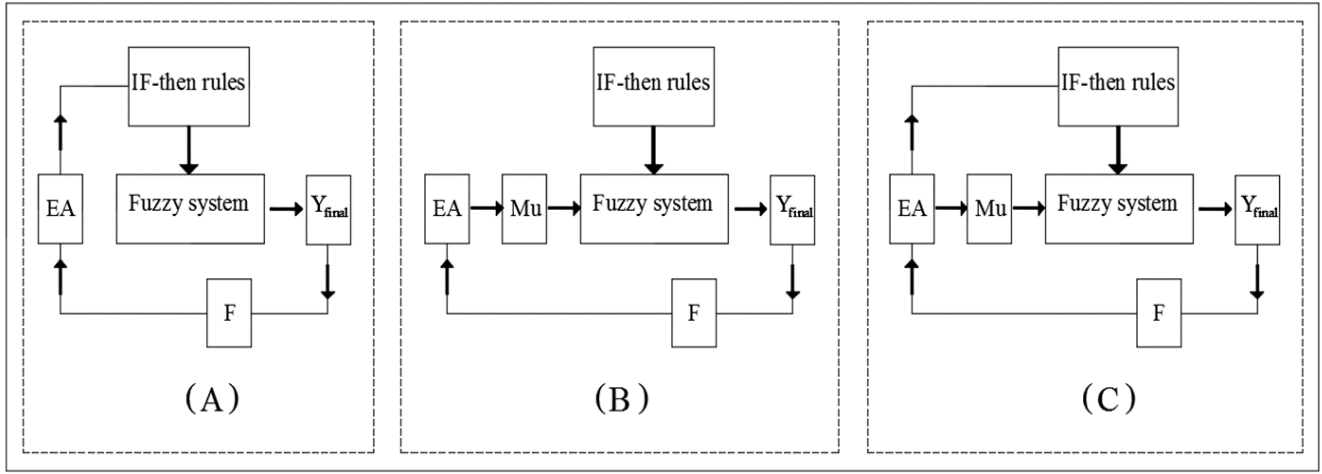

**Fig 7.5. Evolutionary Fuzzy Systems (Siddique and Adeli, 2013).**

As illustrated in Fig (5), there are three modules: A, B and C. Module (A) used EA to develop the optimal fuzzy rules, and model (B) used EA to determine the optimal MFs. whereas module (C) used EA for determining both MFs and Fuzzy rules. (Karr and Gentry, 1993) applied GA for tuning and computing MFs for FL controllers to improve the system performance. (Ishibuchi et al., 1995) developed a genetic fuzzy system for classification. The objective is to generate the maximum number of correct rules with the minimum number of associated rules. (Linkens and Nyongesa, 1995a, b) simplified and generated linguistic rules and fuzzy rules using GA. Fuzzy rules can be fixed and the MFs are tuned, on the other hand, fuzzy rules can be tuned while the MFs can be fixed. (Homaifar and McCormick, 1995) developed a genetic fuzzy system which can simultaneously generate both rule sets and MFs to eliminate the human need in fuzzy system design.

### 7.3.2.1 Tuning of MFs

MF is the basic concept of fuzzy system that is used for fuzzification process. Linear MFs are such as the triangular and trapezoidal MFs and differentiable MFs are such as Gaussian, sigmoidal and bell-shaped (Siddique and Adeli, 2013). No exact method exists to determine the shape of MFs. However, heuristic rules such as computational complexity and the function parameters are used as criterion to select MFs. According to computation complexity criterion, triangular MFs are the most economic. FL system is dependent on some parameters such as mapping of MF and fuzzy rules. The problem is that simple mapping of MFs cannot guarantee the highest performance of system. Therefore, MFs mapping parameters such as number of MFs, MFs overlapping and MF distribution need to be optimized (Kovacic and Bogdan, 2006). The objective function of the fuzzy system performance is minimizing sum of squared error or mean squared error.

## 7.3.2.2 Tuning of fuzzy rules base

The fuzzy rules base or if–then rules are the core of any fuzzy system that consists of a set of if–then rules. The performance of any fuzzy system mainly depends on the rule base which are if–then rules. The number of fuzzy rules grows exponentially with increasing input or output variables and linguistic terms associated for each variable. Accordingly, the experts face a problem to developing these fuzzy rules and determine its combinations.

Evolutionary learning is suitable technique as it can incorporate a priori knowledge to the developed system. The priori knowledge may be in the form of linguistic variables, MF parameters, and fuzzy rules. Learning capability is the main limitation of fuzzy rule-based system. To overcome this limitation, EA such as GA can be incorporated with fuzzy system to obtain learning capability (Bonarini and Trianni, 2001). Evolutionary learning can be merged in fuzzy system to optimize its parameters such as MF parameters, fuzzy rules and number of rules. Structure learning (i.e., rule base learning) and parameter learning (i.e., MF learning) are the two kinds of fuzzy system learning.

Two approaches exist to conduct evolutionary fuzzy system: Michigan approach and Pittsburgh approach (Casillas and Casillas, 2007). Michigan approach is to represent each chromosome as a single rule, whereas the rule set is the entire population. The objective of EA is to select the optimal subset of chromosomes that represents the optimal set of rules. The Michigan approach is outlined as follows:

1. Generate a random initial population of fuzzy if-then rules.
2. Select a sample from the developed population and evaluate the fitness of the rules of the selected sample.
3. Generate new individuals of fuzzy if-then rules by genetic operators.
4. Replace individuals with new individuals of the population.
5. Continue until no further improvement of system performance.

Similarly, Pittsburgh approach represents each chromosome as a set of fuzzy rules where the number of rules constant (Herrera, 2008).

### 7.3.2.3 Objective function

The fitness function is problem-dependent where the objective is to enhance the accuracy and quality of the system performance (Hatanaka et al., 2004). The fitness function can be formulated as the following Equation (7.12) where the objective is to maximize the fitness function and minimize MAPE and the number of rules.

$$\text{Max (F)} = \text{Min} \left( \frac{1}{\text{MAPE+Nr}} \right) \qquad (7.12)$$

Where: (F) is a fitness function and (Nr) is the number of rules and MAPE is as following:

$$\text{MAPE} = \frac{1}{n} \sum_{i=1}^{n} \frac{|(Y_{actual})_i - (Y_{model})_i|}{(Y_{model})_i} \times 100 \qquad (7.13)$$

Where (n) is the number of testing cases, (i) is the number of case and $Y_{model}$ is the outcome of model and $Y_{actual}$ is the actual outcome. Moreover, MAPE can be replaced by the root mean square error (MSE) as following:

$$\text{MSE} = \frac{1}{n} \sum_{i=1}^{n} [(Y_{actual})_i - (Y_{model})_i]^2 \qquad (7.14)$$

### 7.3.2.4 Hybrid fuzzy cost model

Hsiao et al, (2012) have established a Neuro-Fuzzy cost estimation model which is optimized by GA. Such model has accuracy better than the conventional cost method by approximately 20%. The model automatically optimizes the fuzzy rules and fuzzy MFs. Tokede et al, (2014) have built a Neuro-Fuzzy hybrid cost model based on 1600 water infrastructure projects in UK where max-product composition produces better results than the max-min composition. Zhai et al, (2012) have created an improved fuzzy system which is established based on fuzzy c-means (FCM) to solve the problem of fuzzy rules generation. Such model has produced better results for scientific cost prediction. Cheng et al, (2009) have incorporate computation intelligence models such as ANNs, FL and EA to make a hybrid model which improves the prediction accuracy. As a result, an evolutionary fuzzy neural model has been developed for conceptual cost estimation for building projects. Yu and Skibniewski, (2010) have developed an adaptive neuro-fuzzy model for cost estimation for residential construction projects. Such model is an integrated system with ratio estimation method and the adaptive neuro-fuzzy to obtain mining assessment knowledge that is not available in traditional approaches.

Cheng et al, (2009) have developed an evolutionary fuzzy hybrid neural network model for conceptual cost estimation. FL is used for fuzzification and defuzzification for inputs and outputs, respectively. GA is utilized for optimizing the parameter of the model such as NN layer connections and FL membership. Zhu et al, (2010) have conducted evolutionary fuzzy neural network model for cost estimation based on 18 examples and 2 examples for training and testing, respectively. GA is conducted for model optimization and to avoid sinking in local minimum results. Cheng and Roy, (2010) have developed a hybrid artificial intelligence (AI) system based on supportive vector machine (SVM), FL and GA for decision making construction management. The system has applied the FL to handle uncertainty to the system, SVM to map fuzzy inputs and outputs, and GA to optimize the FL and SVM parameters. The objective of such system is to produce accurate results with less human interventions, where MF shapes and distributions can be automatically mapped.

The current studies have developed cost estimate models based on hybrid fuzzy systems. The objective of the hybrid systems is to develop a reliable fuzzy models that have no limitations of traditional fuzzy model such as fuzzy rules generation and MFs tuning. GA or ANNs can be incorporated to fuzzy system to improve its performance, development and accuracy.

## 7. 4 Case study and discussion

Field Canals Improvement Project (FCIP) is a promising project to save fresh water in farm lands during irrigation operations (Radwan, 2013). The objective is to develop a parametric cost estimate model for FCIP for conceptual feasibility studies and cost estimation purposes. Therefore, a total of 111 historical cases are randomly collected from 2010 to 2015 to develop a data base for the cost model. Subsequently, the collected data is divided into two sets: training set (111 case, 77%) and validation set (33 case, 34%).

The first and most important step of parametric model development is to identify the key cost drivers of the case studied where the poor selection of cost drivers lead to poor performance and accuracy of the developed model. This study has evaluated the cost drivers affecting on the FCIPs based on both fuzzy Delphi method, fuzzy analytic hierarchy process where a total of 35 cost drivers are screened to four key cost drivers. Such key cost drivers are command area, PVC pipe line length, and construction year and inflation rate, and number of irrigation valves.

Once key cost drivers are identified, these cost drivers can be applied as inputs to the fuzzy model. Therefore, the following step is to fuzzification the four key cost drivers and identify their MFs as shown in Fig (7.6, A). The most critical stage is to develop fuzzy rules base. Traditionally, experts are consulted to give their experience to develop such rules. For example, the current case study consists of four key cost drivers, and assume that each cost drivers consists of seven MFs as shown in Fig (7.6, B). Accordingly, the number of possible rules may equal $7^4$ (2401 rules). Therefore, there is a need to automatically generate such rules to compete the fuzzy model successfully.

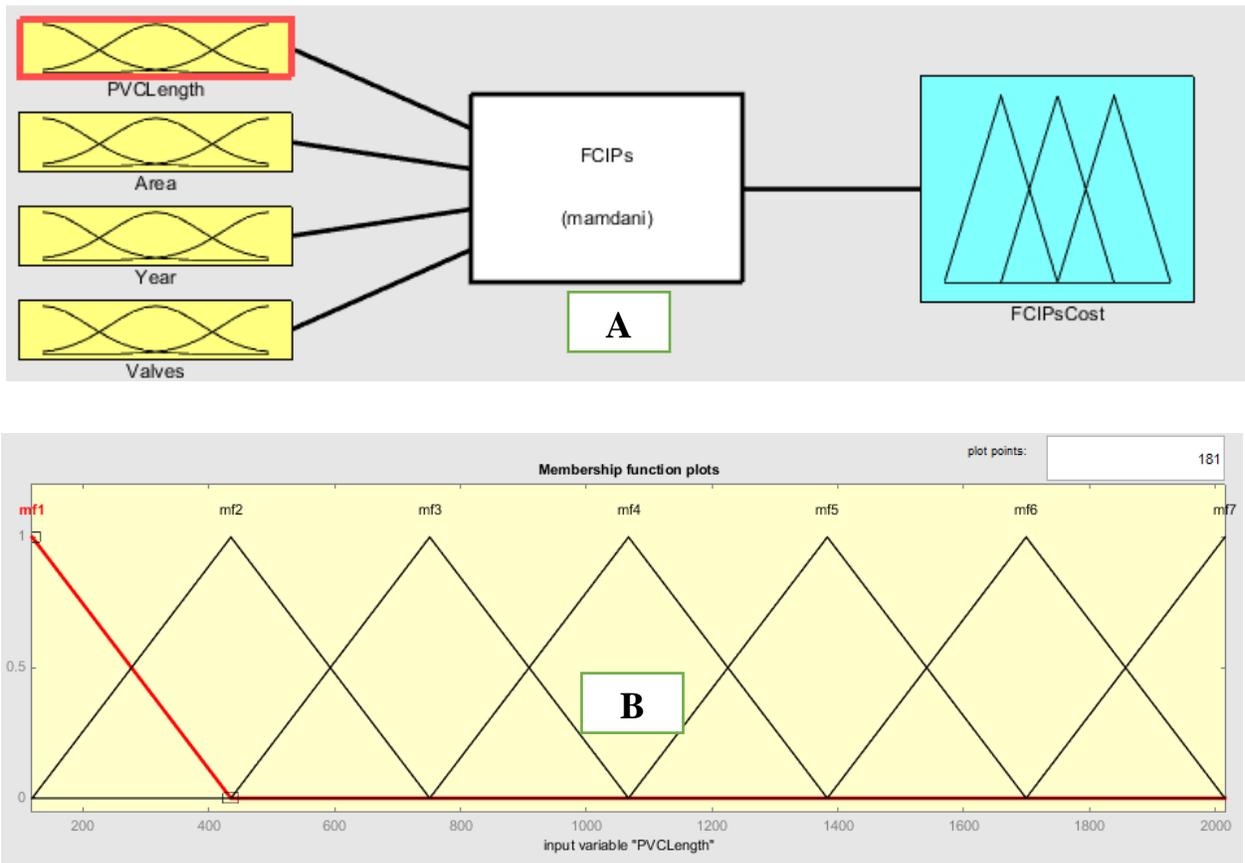

**Fig. 7.6 (A) Fuzzy system for FCIPs, and (B) MFs for PVC length parameters.**

### 7.4.1 Hybrid genetic fuzzy cost model

The study has applied GA to optimally select the fuzzy rules where 2401 rules represent the search space for GA. The number of generated rules by GA are 63 rules and the MAPE is 14.7%. On the other hand, a traditional fuzzy model has been built based on the experts 'experience where a total of 190 rules are generated to cover all the possible combinations of the fuzzy system and MAPE is 26.3 % based on the R programming (Appendix F). That results show that the rules generated by experts may have redundant rules which can be deleted to improve the model performance. Moreover, the expert's knowledge cannot cover all combination to represent all possible rules (2401 rules). In addition, the generation of the experts' rules are time and effort consuming process. However, GA approach provides fewer rules that optimally cover all the possible rules and provide the optimal accuracy and performance of the developed system. Therefore, this study recommends to develop an automated hybrid fuzzy rules models than traditional fuzzy models. In addition, this recommendation can be generalized not only for fuzzy cost estimation models but also for all fuzzy modeling in different applications. Accordingly, hybrid fuzzy modeling is a future research trend in engineering prediction and computation modeling.

## 7.5 Conclusion

The present study has discussed fuzzy modeling and its benefit to obtain uncertainty to the studied case. In addition, the study highlights the main problem for fuzzy modeling    which is fuzzy rules generation. The main limitation of the previous past literature for fuzzy cost modeling is the fuzzy rule generation method. This study has reviewed the hybrid fuzzy model methodologies to generate rules such as evolutionary fuzzy model and neuro fuzzy model. Moreover, a case study have been conducted to investigate the effectiveness of hybrid fuzzy modeling than traditional fuzzy modeling. The study recommendation emphasizes that hybrid fuzzy models such as genetic fuzzy model produces better results than traditional fuzzy models by generating the optimal fuzzy rules.

# CHAPTER 8

# CONCLUSIONS AND RECOMMENDATIONS

## 8.1 Conclusion

This study has developed a reliable parametric cost model for conceptual cost estimate of FCIPs in Egypt, through developing a model that is able to help parties involved in construction projects (owner, contractors, and others) in obtaining the total cost information at the early stages of project with limited available information. Accurate cost estimate means accurate decisions about the project management. Therefore, this study has analyzed the past cost modeling practices to provide a recent direction for construction cost modeling. The study shows that the computational intelligence techniques, artificial intelligence and machine learning have a powerful ability to develop the applicable and accurate cost predictive models. Moreover, cost modeling research area needs more studies to develop intelligent models which have the ability to interpret the resulting cost prediction and analysis the input model's parameters. In addition, this study has provided a list of recommendation and references for cost model developer to build a more practical parametric cost model. Moreover, the core trend of cost modeling is to computerize and automate the cost model where less human interventions required for operating such models with obtaining higher accuracy and optimal results.

This study has discussed the qualitative methods such as Delphi techniques and Fuzzy Analytical hierarchy Process (FAHP) to collect, rank and evaluate the cost drivers of the FCIPs. The current study has used two procedures where both Traditional Delphi Method (TDM) and Fuzzy Delphi Method (FDM) were used to collect and initially rank the cost drivers. Based on the second approach, The FAHP was used to finally rank the screened parameters by FDM. Out of 35 cost drivers, only four parameters were selected as final parameters. The contribution of this study was to find out and evaluate these parameters and to maintain the ability of FDT and FAHP to collect and evaluate the cost drivers of a certain case study. To obtain uncertainty and achieve a more practical model, this study suggested using a fuzzy theory with Delphi methods and with AHP. The screened parameters can be used to develop a precise parametric cost model for FCIPs as a future research work.

On the other hand, this study presents several quantitative techniques to identify these cost drivers. The contribution of this study is providing more than quantitative approaches to identify key cost drivers based on statistical methods such as EFA, regression methods, and correlation matrix. These statistical methods can be combined to develop a hybrid model to have the best subset of key cost drivers. The final key cost drivers are total length (P3), year (P14), number of irrigation valves (P6) and area served (P1). These parameters are extracted by hybrid model 2 where Pearson correlation matrix scanning and stepwise method are used to filter the independent variables.

The research has used four parameters as key parameters that have the most influence on the costs of constructing FCIPs. Data are transformed to produce various regression models where these models are compared based on MAPE. The ANNs model is designed with four nodes in the input layer while the output layer consists of one node representing the total cost of FCIPs. After multiple regression models and ANNs models are developed, the best model is the quadratic transformed model where dependent variable is transformed by the square root. The MAPE is 9.12% and 7.82% for training and validation respectively, and the correlation coefficient is (0.86). To facilitate the usage of the model, a friendly input screen has been developed to receive inputs from the user and to maintain uncertainty and model manipulation, a sensitivity analysis application has been incorporated in the developed model.

## 8.2 Research Recommendations

Based on the survey literatures as Table.2.4, the maximum error extracted is 28.4% and the minimal error extracted is 0.7%, and the average error obtained is 9%. This study has performed a survey and analysis for construction cost modeling development. The study is presented as two main parts. The first part is presenting the most common modeling techniques used for cost models. Whereas, the second part is presenting the review of the current state for cost model development.

The first part is explaining statistical methods such as MRA and intelligent methods such as FL, ANNs, EC, CBR, and hybrid models. The second part is reviewing the model development as summarized in Table.2.4 where modeling techniques, construction project, parameters used, sample size and model accuracy have been extracted and summarized. The following points summarize the recommendations and future trends:

I.    This study recommended using fuzzy approaches such as FDM and FAHP than traditional methods such as TDM and AHP, as the fuzzy approaches produce better reliable performance.

II. This study recommended applying both qualitative and quantities approaches to obtain the most reliable cost drives. Such procedure can be called a hybrid approach for cost drivers' identification. The limitation of the hybrid approach is that both experts' and historical cases are required to be operated on.

III. For future studies, it is recommended to start with asking experts to identify key cost drivers that should be collected to develop the proposed model.

IV. The conceptual cost estimate is conducted under uncertainty. Therefore, this study recommended using fuzzy theory such as Fuzzy Logic and to develop a hybrid model based on Fuzzy Logic to obtain uncertainty nature for the developed model and produce a more reliable performance.

V. It is recommended to develop more than one model to ensure the resulting estimates of cost where the same collected data can be used for developing more than one model such as regression model, ANNs, FL or CBR model. As a result, the researcher can compare the results and set a bench mark to select the most accurate model. In addition, the comparisons of the developed models enhance the quality of cost estimate and the decision based on it (Amason, 1996).

## 8.3 General Recommendations

I. The Genetic algorithm is a powerful tool to select the optimal set of the cost drivers where the prediction error is minimized.

II. It is recommended to study data mining techniques such as factor analysis technique to extract key cost drivers based on quantitative data.

III. Researches should be aware of statistical soft wares such as SPSS and MatLab and programming languages such as python and R to develop automated systems

IV. CI models such as ANNs, FL and Gas are used widely for hybrid model development. Moreover, ML techniques can be efficiently conducted for the parametric cost modeling. Therefore, the cost modeling researcher should firstly study ML, CI, and artificial intelligence (AI) techniques.

V. This study recommends establishing a database for every construction project and such projects to be open source to be used for research development.

Therefore, Government and engineering associations are recommended to establish a database of historical constructed projects to develop cost estimation models. To improve a precession of the developed model, it is recommended to obtain more training data from newly projects. Such projects should to be an open source to be used for researches development.

## 8.4 Recommendations for Future Research trends

I. Computational models and information systems have been applied in business and construction industry to effectively improve the job efficiency (Davis, 1993). Therefore, the hybrid model represents the current trend of parametric cost modeling to improve the model performance and accuracy where the limitations of each technique can be avoided. The objective is to develop computerized automated systems with less interventions of humans to save time, effort and avoid human error for cost estimate. Moreover, computer technologies have a great ability to deal with vast data and complicated computations.

II. Hybrid model can be incorporated to CBR to enhance the performance of CBR such as applying GA and decision tree to optimize attributes weights and applying regression analysis for revising process.

III. CBR represents an increasing importance of ML tools and data mining techniques for knowledge acquisition, prediction and decision making. Specifically, CBR efficiently deals with vast data and has the ability to update case-base for future problem-solving. Moreover, Finding similarities and similar cases improve the reliability and confidence in the output.

IV. Almost of studies focuses on building types of construction projects, a need exists to apply cost models widely for different kinds of construction projects to help project managers and cost estimate engineers.

V. There is a need to develop a model that has the ability to give justification on the model's results, and to give answers and interpretations for the predicted cost. That may require a higher level of AI and may represent the future trend of cost modeling. Moreover, such concept may be generalized for any prediction model. The objective is to avoid the estimator's biases, warn the user to the input parameters of the model, and to avoid the limitation of the black box nature.

However, the study has main limitations where this study has not discussed all models such as SVM and probabilistic models such as Monte Carlo simulation, and decision tree. Almost of studies focuses on building types of construction projects, a need exists to apply cost models widely for different kinds of construction projects. In addition, more reviewed studied means more generalization and better quality of the results.

# 9. References

A.Aamodt, E. Plaza (1994); Case-Based Reasoning: Foundational Issues, Methodological Variations, and System Approaches. AI Communications. IOS Press, Vol. 7: 1, pp. 39-59.

A.Amason, (1996), Distinguishing the effects of functional and dysfunctional conflict on strategic decision making: resolving a paradox for top management teams, Academy of Management Journal 39 123–148.

A.Jrade, (2000), a conceptual cost estimating computer system for building projects. Master Thesis, Department of Building Civil & Environmental Engineering, Concordia University, Montreal, Quebec, Canada.

AACE International recommended practices. (2004). AACE International, Morgantown, W.V.

Abdal-Hadi, M., 2010. Factors Affecting Accuracy of Pre-tender Cost Estimate in Gaza Strip., Gaza strip. Master thesis in construction management, The Islamic University of Gaza Strip.

Adeli, H., and Wu, M. (1998). "Regularization Neural Network for Construction Cost Estimation." Journal of Construction Engineering and Management, 124(1), 18–24.

Ahiaga-Dagbui DD, Tokede O, Smith SD and Wamuziri, S. (2013) "a neuro-fuzzy hybrid model for predicting final cost of water infrastructure projects". Procs 29th Annual ARCOM Conference, 2-4, Reading, UK, Association of Researchers in Construction Management, 181-190.

Akintoye, A. (2000). "Analysis of factors influencing project cost estimating practice." Construction Management and Economics, 18(1), 77–89.

Allyn & Zhang, S., & Hong, S. (1999). Sample size in factor analysis. Psychological Methods, 4(1), 84–99.

Alroomi, A., Jeong, D. H. S., and Oberlender, G. D. (2012). "Analysis of Cost-Estimating Competencies Using Criticality Matrix and Factor Analysis." Journal of Construction Engineering and Management J. Constr. Eng. Manage., 138(11), 1270–1280.

Al-Thunaian, S., (1996). Cost estimation practices for buildings by A/E firms in the eastern province, Saudi Arabia. Unpublished master thesis in construction engineering and management, Dhahran, Saudi Arabia. Master thesis in construction engineering and management. King Fahd University of Petroleum and Minerals.

An, S.-H., Kim, G.-H., and Kang, K.-I. (2007). "A case-based reasoning cost estimating model using experience by analytic hierarchy process." Building and Environment, 42(7), 2573–2579.

An, S.-H., Park, U.-Y., Kang, K.-I., Cho, M.-Y., and Cho, H.-H. (2007). "Application of Support Vector Machines in Assessing Conceptual Cost Estimates." Journal of Computing in Civil Engineering, 21(4), 259–264.

Attalla, M., and Hegazy, T. (2003). "Predicting Cost Deviation in Reconstruction Projects: Artificial Neural Networks versus Regression." Journal of Construction Engineering and Management, 129(4), 405–411.

B.H Ross, (1989), some psychological results on case-based reasoning. Case-Based Reasoning Workshop, DARPA 1989. Pensacola Beach. Morgan Kaufmann, 1989. pp. 144-147.

Bayram, S., Ocal, M. E., Oral, E. L., and Atis, C. D. (2015). "Comparison of multi-layer perceptron (MLP) and radial basis function (RBF) for construction cost estimation: the case of Turkey." Journal of Civil Engineering and Management, 22(4), 480–490.

Bertram D. Likert (2017) Scales are the meaning of life. Available from: http://poincare.matf.bg.ac.rs/~kristina/topic-dane-likert.pdf. Accessed February 20, 2017.

Bezdek, J.C. (1994) what is computational intelligence? In Computational Intelligence Imitating Life, J.M. Zurada, R.J. Marks II and C.J. Robinson (eds), IEEE Press, New York, pp. 1–12.

Bode, J. (2000). "Neural networks for cost estimation: Simulations and pilot application." International Journal of Production Research, 38(6), 1231–1254.

Bowerman, B. L., & O'Connell, R. T. (1990). Linear statistical models: An applied approach (2nd ed.). Belmont, CA: Duxbury. (This text is only for the mathematically minded or postgraduate students but provides an extremely thorough exposition of regression analysis.

Buckley, J.J., (1985). Fuzzy hierarchical analysis. Fuzzy Sets and Systems, 34, 187-195.

Caputo, A. & Pelagagge, P.,(2008). Parametric and neural methods for cost estimation of process vessels. Int. J. Production Economics, Volume 112, p. 934–954.

Casillas, J., Carse, B., and Bull, L. (2007). "Fuzzy-XCS: A Michigan Genetic Fuzzy System." IEEE Transactions on Fuzzy Systems, 15(4), 536–550.
Cattell, R. B. (1966). The scree test for the number of factors. Multivariate Behavioral Research, 1, 245–276.

Cavalieri S., Maccarrone P., Pinto R., 2004. Parametric vs. neural network models for the estimation of production costs: a case study in the automotive industry. Int J Prod Econ 91, 165-177.

Chen, H. (2002). "A comparative analysis of methods to represent uncertainty in estimating the cost of constructing wastewater treatment plants." Journal of Environmental Management, 65(4), 383–409.

Cheng, M.-Y., Tsai, H.-C., and Hsieh, W.-S. (2009). "Web-based conceptual cost estimates for construction projects using Evolutionary Fuzzy Neural Inference Model." Automation in Construction, 18(2), 164–172.

Cheng, M.-Y., and Roy, A. F. (2010). "Evolutionary fuzzy decision model for construction management using support vector machine." Expert Systems with Applications, 37(8), 6061–6069.

Cheng, M.-Y., Hoang, N.-D., and Wu, Y.-W. (2013). "Hybrid intelligence approach based on LS-SVM and Differential Evolution for construction cost index estimation: A Taiwan case study." Automation in Construction, 35, 306–313.

Cheng, M.-Y., and Hoang, N.-D. (2014). "Interval estimation of construction cost at completion using least squares support vector machine." Journal of Civil Engineering and Management, 20(2), 223–236.

Chi, Z.,Yan, H. and Phan, T. (1996) Fuzzy Algorithms:With Applications to Image Processing and Pattern Recognition,World Scientific, Singapore.

Choi, S., Ko, K. and Hong, D. (2001) A multilayer feedforward neural network having N/4 nodes in two hidden layers, Proceedings of the IEEE International Joint Conference on Neural Networks,Washington, DC, Vol. 3, pp. 1675–1680.

Choi, S., Kim, D. Y., Han, S. H., and Kwak, Y. H. (2014). "Conceptual Cost-Prediction Model for Public Road Planning via Rough Set Theory and Case-Based Reasoning." Journal of Construction Engineering and Management, 140(1), 04013026.

Choon, T. & Ali, K. N., (2008). A review of potential areas of construction cost estimating and identification of research gaps. Journal Alam Bina, 11(2), pp. 61-72.
Chou, J.-S. (2009). "Web-based CBR system applied to early cost budgeting for pavement maintenance project." Expert Systems with Applications, 36(2), 2947–2960.

Chou, J.-S., Lin, C.-W., Pham, A.-D., and Shao, J.-Y. (2015). "Optimized artificial intelligence models for predicting project award price." Automation in Construction, 54, 106–115.

Comrey, A. L., & Lee, H. B. (1992). A first course in factor analysis (2nd ed.). Hillsdale, NJ: Erlbaum

Cook, R. D., & Weisberg, S. (1982).Residuals and influence in regression. New York: Chapman& Hall.

Cordon, O., Herrera, F., Hoffmann, F. and Magdalena, L. (2001) Genetic Fuzzy Systems: Evolutionary Tuning and Learning of Fuzzy Knowledge Bases, World Scientific, Singapore.

Darwin, C. (1859) The Origin of Species by Means of Natural Selection or the Preservation of Favoured Races in the Struggle for Life, Mentor Reprint 1958, New York.

Doğan, S. Z., Arditi, D., and Günaydin, H. M. (2008). "Using Decision Trees for Determining Attribute Weights in a Case-Based Model of Early Cost Prediction." Journal of Construction Engineering and Management, 134(2), 146–152.

Draper, N. R., and Smith, H. (1998). Applied regression analysis. Wiley, New York.

Duran, O., Rodriguez, N. & Consalter, L.,(2009). Neural networks for cost estimation of shell and tube heat exchangers. Expert Systems with Applications, Volume 36, p. 7435–7440.

Durbin, J., & Watson, G. S. (1951).Testing for serial correlation in least squares regression, II.Biometrika, 30, 159–178.

Dursun, O., and Stoy, C. (2016). "Conceptual Estimation of Construction Costs Using the Multistep Ahead Approach." Journal of Construction Engineering and Management, 142(9), 04016038.

Dysert, L. R. (2006) Is "estimate accuracy" an oxymoron?. AACE International Transactions EST.01: EST01.1 - 01.5.

Dziuban, Charles D.; Shirkey, Edwin C. (1974), Psychological Bulletin, Vol 81(6), 358-361.http://dx.doi.org/10.1037/h0036316.

Dell'Isola M.D., 2002, Architect's Essentials of Cost Management, Wiley & Sons, Inc., New York, NY.

ElSawy.I, Hosny.H, Abdel Razek.M,(2011), A Neural Network Model for Construction Projects Site Overhead Cost Estimating in Egypt, IJCSI International Journal of Computer Science Issues, Vol. 8, Issue 3, No. 1, May 2011 ISSN (Online): 1694-0814

El Sawalhi,N.I. (2012), "modeling the Parametric Construction Project Cost Estimate using Fuzzy Logic", International Journal of Emerging Technology and Advanced Engineering, Website: www.ijetae.com (ISSN 2250-2459, Volume 2, Issue 40).

El-Sawalhi, N. I., and Shehatto, O. (2014). "A Neural Network Model for Building Construction Projects Cost Estimating." Journal of Construction Engineering and Project Management, 4(4), 9–16.

Elbeltagi.E, Hosny.O, Abdel-Razek.R, El-Fitory.A (2014), Conceptual Cost Estimate of Libyan Highway Projects Using Artificial Neural Network, Int. Journal of Engineering Research and Applications www.ijera.com ISSN: 2248-9622, Vol. 4, Issue 8( Version 5), pp.56-66"

Elfaki, A. O., Alatawi, S., and Abushandi, E. (2014). "Using Intelligent Techniques in Construction Project Cost Estimation: 10-Year Survey." Advances in Civil Engineering, 2014, 1–11.

Elmousalami, H. H., Elyamany, A. H., and Ibrahim, A. H. (2017). "Evaluation of Cost Drivers for Field Canals Improvement Projects." Water Resources Management.

El-Sawah H, Moselhi O (2014). Comparative study in the use of neural networks for order of magnitude cost estimating in construction, ITcon Vol. 19, pg. 462-473, http://www.itcon.org/2014/27

Emami, M.R., Turksen, I.B. and Goldberg, A.A. (1998) Development of a systematic methodology of fuzzy logic modelling, IEEE Transactions on Fuzzy Systems, 6(3), 346-361.

Emsley, M. W., Lowe, D. J., Duff, A. R., Harding, A., and Hickson, A. (2002). "Data modelling and the application of a neural network approach to the prediction of total construction costs." Construction Management and Economics, 20(6), 465–472.

Engelbrecht, A.P. (2002) Computational Intelligence: An Introduction, John Wiley & Sons, New York.

Erensal, Y. C., Öncan, T., and Demircan, M. L. (2006). "Determining key capabilities in technology management using fuzzy analytic hierarchy process: A case study of Turkey." Information Sciences, 176(18), 2755–2770.

F.D. Davis, (1993), User acceptance of information technology: system characteristics, user perceptions, and behavioral impacts, International Journal of Man–Machine Studies 38 475–487.

Field, A., (2009). Discovering Statistics Using SPSS for Windows. Sage Publications, London e Thousand Oaks e New Delhi

Flom, P. L. and Cassell, D. L. (2007) "Stopping stepwise: Why stepwise and similar selection methods are bad, and what you should use," NESUG 2007.

Green, S. B. (1991). How many subjects does it take to do a regression analysis? Multivariate Behavioral Research, 26,499–510.

Guadagnoli, E., & Velicer, W. F. (1988). Relation of sample size to the stability of component patterns. Psychological Bulletin, 103(2), 265–275.

Günaydın, H. M., and Doğan, S. Z. (2004). "A neural network approach for early cost estimation of structural systems of buildings." International Journal of Project Management, 22(7), 595–602.

GUYON, I., and ELISSEEFF, A. (2003). "An Introduction to Variable and Feature Selection." Journal of Machine Learning Research 3 1157-1182.

Hatanaka, T., Kawaguchi, Y. and Uosaki, K. (2004) Nonlinear system identification based on evolutionary fuzzy modelling, IEEE Congress on Evolutionary Computing, 1, 646–651.

Hays, W. L. (1983). "Using Multivariate Statistics." Psyccritiques, 28(8).

Hegazy, T., and Ayed, A. (1998). "Neural Network Model for Parametric Cost Estimation of Highway Projects." Journal of Construction Engineering and Management J. Constr. Eng. Manage., 124(3), 210–218.

Hegazy, T. (2002). Computer-based construction project management. Prentice Hall.

Herrera, F. (2008). "Genetic fuzzy systems: taxonomy, current research trends and prospects." Evolutionary Intelligence, 1(1), 27–46.

Hinze, J.,(1999). Construction Planning and Scheduling. Columbus, Ohio: Prentice Hall.

Holland, J.H. (1975) Adaptation in Natural and Artificial Systems, University Michigan Press, Ann Arbor, MI.

Homaifar, A. and McCormick, E. (1995) Simultaneous design of membership functions and rule sets for fuzzy controllers using genetic algorithms, IEEE Transactions on Fuzzy Systems, 3(2), 129–139.

Hopfield, J.J. (1982), neural networks and physical systems with emergent collective computational abilities, Proceedings of National Academy of Sciences, 79, 2554–2558.

Hsiao, F.-Y., Wang, S.-H., Wang, W.-C., Wen, C.-P., and Yu, W.-D. (2012). "Neuro-Fuzzy Cost Estimation Model Enhanced by Fast Messy Genetic Algorithms for Semiconductor Hookup Construction." Computer-Aided Civil and Infrastructure Engineering, 27(10), 764–781.

Hsu Chia-Chien, Sandford Brian A. (2007)."The Delphi Technique:Making Sense Of Consensus,Practical Assessment, Research & Evaluation, Vol 12, No 10 2"

Hsu, Y.-L., Lee, C.-H., and Kreng, V. (2010). "The application of Fuzzy Delphi Method and Fuzzy AHP in lubricant regenerative technology selection." Expert Systems with Applications, 37(1), 419–425.

Huang, S.-C. and Huang, Y.-F. (1991) Bounds on the number of hidden neurons in multilayer neurons, IEEE

Humphreys, K., 2004. Project and cost engineers. 4th ed. s.l.:Marcel Dekker.

Hutcheson, G., & Sofroniou, N. (1999). The multivariate social scientist. London: Sage. Chapter 4.

Ishibuchi, H., Nozaki, K., Yamamoto, N. and Tanaka, H. (1995) Selecting fuzzy if–then rules for classification problems using genetic algorithms, IEEE Transactions on Fuzzy Systems, 3, 260–270.

Ishikawa, A., Amagasa, M., Shiga, T., Tomizawa, G., Tatsuta, R., and Mieno, H. (1993). "The max-min Delphi method and fuzzy Delphi method via fuzzy integration." Fuzzy Sets and Systems, 55(3), 241–253.

Ji, S.-H., Park, M., and Lee, H.-S. (2012). "Case Adaptation Method of Case-Based Reasoning for Construction Cost Estimation in Korea." Journal of Construction Engineering and Management, 138(1), 43–52.

Jin, R., Cho, K., Hyun, C., and Son, M. (2012). "MRA-based revised CBR model for cost prediction in the early stage of construction projects." Expert Systems with Applications, 39(5), 5214–5222.

Jolliffe, I. T. (1986). Principal component analysis. New York: Springer.

Ju Kim K, Kim K. (2010), Preliminary cost estimation model using case-based reasoning and genetic algorithms. Journal of Computing in Civil Engineering. 24(6):499–505.

Phaobunjong. K, (2002) Parametric cost estimating model for conceptual cost estimating of building construction projects, Ph.D. Dissertation, University of Texas, and Austin, TX.

Kaiser, H. F. (1960). The application of electronic computers to factor analysis. Educational and Psychological Measurement, 20, 141–151.

Kaiser, H.F., (1970). A second generation little jiffy. Psychometrika 35, 401e415.

Kaiser, H.F., (1974). An index of factorial simplicity. Psychometrika 39, 31e36.

Kan, P. (2002). "Parametric Cost Estimating Model for Conceptual Cost Estimating of Building Construction Projects." Ph.D. Thesis. University of Texas, Austin, USA.

Karatas, Y., and Ince, F. (2016). "Feature article: Fuzzy expert tool for small satellite cost estimation." IEEE Aerospace and Electronic Systems Magazine, 31(5), 28–35.

Karr, C.L. and Gentry, E.J. (1993) Fuzzy control of pH using genetic algorithms, IEEE Transactions on Fuzzy Systems, 1(1), 46–53.

Kass, R.A., Tinsley, H.E.A., (1979). Factor analysis. J. Leis. Res. 11, 120e138.

Kim, G.-H., An, S.-H., and Kang, K.-I. (2004). "Comparison of construction cost estimating models based on regression analysis, neural networks, and case-based reasoning." Building and Environment, 39(10), 1235–1242.

Kim, G. H., Seo, D. S., and Kang, K. I. (2005). "Hybrid Models of Neural Networks and Genetic Algorithms for Predicting Preliminary Cost Estimates." Journal of Computing in Civil Engineering J. Comput. Civ. Eng., 19(2), 208–211.

Kim, K. J., and Kim, K. (2010). "Preliminary Cost Estimation Model Using Case-Based Reasoning and Genetic Algorithms." Journal of Computing in Civil Engineering, 24(6), 499–505.

Kim, S. (2013). "Hybrid forecasting system based on case-based reasoning and analytic hierarchy process for cost estimation." Journal of Civil Engineering and Management, 19(1), 86–96.

Kline, P. (1999). The handbook of psychological testing (2nd ed.). London: Routledge.

Klir George J. & Yuan Bo (1995). Fuzzy Sets and Fuzzy Logic Theory and Applications.

Knight, K., and Fayek, A. R. (2002). "Use of Fuzzy Logic for Predicting Design Cost Overruns on Building Projects." Journal of Construction Engineering and Management, 128(6), 503–512.

Kohavi, R. and John, G. (1997) Wrappers for feature subset selection, Artificial Intelligence, 97(1/2), 273–324.

Kolodner.J.L, (1992), An Introduction to Case-Based Reasoning, Artificial Intelligence Review 6, 3--34, College of Computing, Georgia Institute of Technology, Atlanta, GA 30332-0280, U.S.A.

Kovacic, Z. and Bogdan, S. (2006) Fuzzy Controller Design: Theory and Application, CRC Press, Boca Raton, FL.

Laarhoven, P. J. M., & Pedrycz, W. (1983). A fuzzy extension of Sati's priority theory. Fuzzy Sets and System, 11, 229–241.

Leng, K. C., (2005). Principles of knowledge transfer in cost estimating conceptual model, Malaysia: University Teknologi Malaysia.

Linkens, D.A. and Nyongesa, H.O. (1995a) Genetic algorithms for fuzzy control, Part 1: Offline system development and application, IEE Proceedings of Control Theory and Application, 142(3), 161–176.

Linkens, D.A. and Nyongesa, H.O. (1995b) Genetic algorithms for fuzzy control, Part 2: Online system development and application, IEE Proceedings of Control Theory and Application, 142(3), 177–185.

Liu, W.-K. (2013). "Application of the Fuzzy Delphi Method and the Fuzzy Analytic Hierarchy Process for the Managerial Competence of Multinational Corporation Executives."

Loucks, D. P., Beek, E. van., Stedinger, J. R., Dijkman, J. P. M., and Villars, M. T. (2005). Water resources systems planning and management: an introduction to methods, models and applications. UNESCO, Paris.

Love, P. E. D., Tse, R. Y. C., and Edwards, D. J. (2005). "Time–Cost Relationships in Australian Building Construction Projects." Journal of Construction Engineering and Management J. Constr. Eng. Manage., 131(2), 187–194.

Lowe, D. J., Emsley, M. W., and Harding, A. (2006). "Predicting Construction Cost Using Multiple Regression Techniques." Journal of Construction Engineering and Management J. Constr. Eng. Manage., 132(7), 750–758.

Ma, L., Shen, S., Zhang, J., Huang, Y., and Shi, F. (2010). "Application of fuzzy analytic hierarchy process model on determination of optimized pile-type." Frontiers of Architecture and Civil Engineering in China Front. Archit. Civ. Eng. China, 4(2), 252–257.

Mahdavi Damghani, Babak (2012). "The Misleading Value of Measured Correlation". Wilmott. 2012 (1): 64–73.doi:10.1002/wilm.10167.

Mahdavi Damghani B. (2013). "The Non-Misleading Value of Inferred Correlation: An Introduction to the Cointelation Model". Wilmott Magazine.doi:10.1002/wilm.10252.

Mamdani, E.H. and Assilian, S. (1974) Application of fuzzy algorithms for control of simple dynamic plant, Proceedings of IEEE, 121, 1585–1588."

Manoliadis, O. G., Pantouvakis, J. P., and Christodoulou, S. E. (2009). "Improving qualifications-based selection by use of the fuzzy Delphi method." Construction Management and Economics, 27(4), 373–384.

Marzouk, M. M., and Ahmed, R. M. (2011). "A case-based reasoning approach for estimating the costs of pump station projects." Journal of Advanced Research, 2(4), 289–295.

Marzouk, M., and Amin, A. (2013). "Predicting Construction Materials Prices Using Fuzzy Logic and Neural Networks." Journal of Construction Engineering and Management, 139(9), 1190–1198.

Marzouk, M., and Alaraby, M. (2014). "Predicting telecommunication tower costs using fuzzy subtractive clustering." Journal of Civil Engineering and Management, 21(1), 67–74.

Marzouk, M., and Elkadi, M. (2016). "Estimating water treatment plants costs using factor analysis and artificial neural networks." Journal of Cleaner Production, 112, 4540–4549.

McCulloch, W.S. and Pitts, W.H. (1943) A logical calculus of the ideas imminent in nervous activity, Bulletin of Mathematical Biophysics, 5, 115–133.

Menard, S. (1995). Applied logistic regression analysis. Sage university paper series on quantitative applications in the social sciences, 07–106. Thousand Oaks, CA: Sage.

Miles, J. N. V., & Shevlin, M. (2001). Applying regression and correlation: a guide for students and researchers. London: Sage. (This is an extremely readable text that covers regression in loads of detail but with minimum pain – highly recommended.

Ministry of Public Works and Water Resources, (1998). Egypt's irrigation improvement program. rep., 1–118, ministry of public works and water resources, US Agency for International Development, Agricultural Policy Reform Program, Egypt.

Moret, Y., and Einstein, H. H. (2016). "Construction Cost and Duration Uncertainty Model: Application to High-Speed Rail Line Project." Journal of Construction Engineering and Management, 142(10), 05016010.

Moselhi, O., and Hegazy, T. (1993). "Markup estimation using neural network methodology." Computing Systems in Engineering, 4(2-3), 135–145.

Myers, R. (1990). Classical and modern regression with applications (2nd ed.). Boston, MA: Duxbury.

Nunnally, J.C., (1978). Psychometric Theory. McGraw-Hill, New York.

Ostwald, P., (2001). Construction cost analysis and estimating. s.l.:Upper Saddle River., N.J. : Prentice Hall. Overview of applications. European Journal of Operational.

Ozdemir, M., Embrechts, M.J., Arciniegas, F., Breneman, C.M., Lockwood, L. and Bennett, K.P. (2001) Feature selection for in-silico drug design using genetic algorithms and neural networks. IEEE Mountain Workshop on Soft Computing in Industrial Applications, Blacksburg, VA, pp. 53–57.

Pan, N.-F. (2008). "Fuzzy AHP approach for selecting the suitable bridge construction method." Automation in Construction, 17(8), 958–965.

Papadopoulos, B., Tsagarakis, K. P., and Yannopoulos, A. (2007). "Cost and Land Functions for Wastewater Treatment Projects: Typical Simple Linear Regression versus Fuzzy Linear Regression." Journal of Environmental Engineering, 133(6), 581–586.

Park, H.-S., and Kwon, S. (2011). "Factor analysis of construction practices for infrastructure projects in Korea." KSCE Journal of Civil Engineering KSCE J Civ Eng, 15(3), 439–445.

Perera, S., & Watson, I. (1998). Collaborative case-based estimating and design. Advances in Engineering Software, 29(10), 801–808.

Petroutsatou, K., Georgopoulos, E., Lambropoulos, S., and Pantouvakis, J. P., (2012) "Early cost estimating of road tunnel construction using neural networks," Journal of Construction Engineering and Management, Vol. 138(6), pp. 679–687.

Petruseva.S, Phil Sherrod, Pancovska.V.Z , Petrovski.A , (2016), Predicting Bidding Price in Construction using Support Vector Machine, TEM Journal 5(2) 143–151.

Peurifoy, R.L. and Oberlender, G.D. (2002), "Estimating Construction Costs", 5th Edition, McGraw-Hill, New York.

Polit DF Beck CT (2012). Nursing Research: Generating and Assessing Evidence for Nursing Practice, 9th ed. Philadelphia, USA: Wolters Klower Health, Lippincott Williams & Wilkins.

Radwan,H.G. ( 2013). "Sensitivity Analysis of Head Loss Equations on the Design of Improved Irrigation On-Farm System in Egypt", International Journal of Advancements in Research & Technology, Volume 2, Issue1.

Ranasinghe, M. (2000). "Impact of correlation and induced correlation on the estimation of project cost of buildings." Construction Management and Economics, 18(4), 395–406.

Ratner, B. (2010). "Variable selection methods in regression: Ignorable problem, outing notable solution." J Target Meas Anal Mark Journal of Targeting, Measurement and Analysis for Marketing, 18(1), 65–75.

Rockwell, R. C. (1975). Assessment of multicollinearity: The Haitovsky test of the determinant.Sociological Methods and Research, 3(4), 308–320.

Runker, T.A. (1997) Selection of appropriate deffuzification methods using application specific properties, IEEE Transactions on Fuzzy Systems, 5(1), 72–79."

Rutkowski, L. (2005) New Soft Computing Techniques for System Modeling, Pattern Classification and Image.

S.-H. Ji, M. Park, H.-S. Lee, (2011) Cost estimation model for building projects using casebased reasoning, Can. J. Civ. Eng. 38 (5) (2011) 570–581.

Saaty, T. L. (1980). The analytic hierarchy process: planning, priority setting. New York: McGraw Hill International Book.

Saaty, T. L. (1994). "How to Make a Decision: The Analytic Hierarchy Process."Interfaces, 24(6), 19–43.

Sabol, L., (2008), "Challenges in cost estimating with building information modeling". Design + Construction Strategies, LLC, 11 Dupont Circle, Suite 601 Washington, DC 20036.USA.

Salem, A., Elbeltagi, E., Abdel-Razek, R. (2008) "Predicting Conceptual Cost Of Libyan Highway Projects Using Artific ial Neura l Netwo rk", Thesis(MSc),

Department Of Construction And Building, Arab Academy For Sciences, Technology And Maritime Transport.

Shaheen, A. A., Fayek, A. R., and Abourizk, S. M. (2007). "Fuzzy Numbers in Cost Range Estimating." Journal of Construction Engineering and Management, 133(4), 325–334.

Shehatto.M.O and EL-Sawalhi.N (2013) "Cost estimation for building construction projects in Gaza Strip using Artificial Neural Network (ANN)", M.Sc. Thesis, The Islamic University – Gaza.

Shi, H., Song, J., and Zhang, X. (2010). "The method and application for estimating construction project costs." 2010 IEEE International Conference on Advanced Management Science (ICAMS 2010).

Siddique, N., and Adeli, H. (2013). Computational intelligence: synergies of fuzzy logic, neural networks and evolutionary computing. John Wiley & Sons, Chichester, West Sussex.

Siedlecki, W. and Sklansky, J. (1988) on automatic feature selection, International Journal of Pattern Recognition and Artificial Intelligence, 2(2), 197–220.

Siedlecki, W. and Sklansky, J. (1989) A note on genetic algorithms for large-scale feature selection, Pattern Recognition

Son, H., Kim, C., and Kim, C. (2012). "Hybrid principal component analysis and support vector machine model for predicting the cost performance of commercial building projects using pre-project planning variables." Automation in Construction, 27, 60–66.

Soto, B. G. C. A. D. D., and Adey, B. T. (2015). "Investigation of the Case-based Reasoning Retrieval Process to Estimate Resources in Construction Projects." Procedia Engineering, 123, 169–181.

Srichetta, P., and Thurachon, W. (2012). "Applying Fuzzy Analytic Hierarchy Process to Evaluate and Select Product of Notebook Computers." International Journal of Modeling and Optimization, Vol. 2(2) pp. 168–173.

Stevens, J. P. (2002). Applied multivariate statistics for the social sciences (4th ed.). Hillsdale,NJ: Erlbaum.

Stewart, Rodney D. (1991) Cost Estimating, 2nd ed., John Wiley & Sons, Inc., New York.

Stoy, C., and Schalcher, H.-R. (2007). "Residential Building Projects: Building Cost Indicators and Drivers." Journal of Construction Engineering and Management J. Constr. Eng. Manage., 133(2), 139–145.

Stoy, C., Pollalis, S., and Schalcher, H.-R. (2008). "Drivers for Cost Estimating in Early Design: Case Study of Residential Construction." Journal of Construction Engineering and Management J. Constr. Eng. Manage., 134(1), 32–39.

Stoy, C., Pollalis, S., and Dursun, O. (2012). "A concept for developing construction element cost models for German residential building projects." IJPOM International Journal of Project Organisation and Management, 4(1), 38.

Sugeno, M. and Kang, G.T. (1988) Structure identification of fuzzy model, Fuzzy Sets and Systems, 28, 15–33.

Tabachnick, B.G. and Fidell, L.S. (2007) Using Multivariate Statistics, Pearson/Allyn & Bacon, Boston.

Takagi, T. and Sugeno, M. (1985) Fuzzy identification of systems and its application to modeling and control, IEEE Transactions on Systems, Man and Cybernetics, 15, 116–132.

Takagi, H. and Hayashi, I. (1991) NN-driven fuzzy reasoning, International Journal of Approximate Reasoning, 5(3), 191–212.

Takagi, H. (1995) Applications of neural networks and fuzzy logic to consumer products. Industrial Applications of Fuzzy Control and Intelligent Systems, J. Yen, R. Langari and L. Zadeh (eds), IEEE Press, Piscataway, NJ, pp.

Thomas L. Saaty, (2008), Decision making with the analytic hierarchy process, Int. J. Services Sciences, Vol. 1, No. 1, 2008

Tokede, O., Ahiaga-Dagbui, D., Smith, S., and Wamuziri, S. (2014). "Mapping Relational Efficiency in Neuro-Fuzzy Hybrid Cost Models." Construction Research Congress. Transactions on Neural Networks, 2(1), 47–55.

Trefor P. Williams (2002) Predicting completed project cost using bidding data, Construction Management and Economics, 20:3, 225-235, DOI: 10.1080/01446190110112838

Tsukamoto, Y. (1979) An approach to fuzzy reasoning method. In Advances in Fuzzy Set Theory and Applications, M.M. Gupta, R.K. Ragade and R. Yager (eds), North-Holland, Amsterdam, pp. 137–149.

Vaidya, O.S., Kumar, S., (2006). Analytic hierarchy process: Analytic hierarchy process: An overview of applications, European Journal of Operational Research, Volume 169, Issue 1, 16 February 2006, Pages 1-29

Wang, L.-X. (1997) Adaptive Fuzzy Systems and Control: Design and Stability Analysis, Prentice-Hall, Englewood

Wang, L.-X. (1997) Adaptive Fuzzy Systems and Control: Design and Stability Analysis, Prentice-Hall, Englewood

Wang, Y.-R., Yu, C.-Y., and Chan, H.-H. (2012). "Predicting construction cost and schedule success using artificial neural networks ensemble and support vector machines classification models." International Journal of Project Management, 30(4), 470–478.

Wheaton, William C. and Simonton, William E., (2007), The Secular and Cyclic Behavior of True Construction Costs. Journal of Real Estate Research Vol. 29, No. 1. Available at SSRN: https://ssrn.com/abstract=979008

Wilkinson, L., & Dallal, G.E. (1981). Tests of significance in forward selection regression with an F-to enter stopping rule. Technometrics, 23, 377–380.

Williams, T. P. (2002). "Predicting completed project cost using bidding data." Construction Management and Economics, 20(3), 225–235.

Wilmot, C. G., and Mei, B. (2005). "Neural Network Modeling of Highway Construction Costs." Journal of Construction Engineering and Management, 131(7), 765–771.

Woldesenbet, A., and Jeong, "david" H. S. (2012). "Historical Data Driven and Component Based Prediction Models for Predicting Preliminary Engineering Costs of Roadway Projects." Construction Research Congress 2012.

Xu, M., Xu, B., Zhou, L., and Wu, L. (2015). "Construction Project Cost Prediction Based on Genetic Algorithm and Least Squares Support Vector Machine." Proceedings of the 5th International Conference on Civil Engineering and Transportation 2015.

Yang, I.-T. (2005). "Simulation-based estimation for correlated cost elements."International Journal of Project Management, 23(4), 275–282.

Yang, J. and Honavar, V. (1998) Feature subset selection using a genetic algorithm, IEEE Intelligent Systems, 13(2), 44–49.

Yang, S.-S., and Xu, J. (2010). "The application of fuzzy system method to the cost estimation of construction works." 2010 International Conference on Machine Learning and Cybernetics.

Yu, W.-D., and Skibniewski, M. J. (2010). "Integrating Neurofuzzy System with Conceptual Cost Estimation to Discover Cost-Related Knowledge from Residential Construction Projects." Journal of Computing in Civil Engineering, 24(1), 35–44.

Zadeh, L.A. (1965) Fuzzy sets, Information and Control, 8(3), 338–353.

Zadeh, L.A. (1973) Outline of a new approach to the analysis of complex systems and decision process, IEEE Transactions on System, Man and Cybernetics, 3, 28–44.

Zadeh, L.A. (1976), the concept of linguistic variable and its application to approximate reasoning – III, Information.

Zadeh, L.A. (1994) Fuzzy logic, neural networks and soft computing, Communications of the ACM, 37, 77–84.

Zhai, K., Jiang, N., and Pedrycz, W. (2012). "Cost prediction method based on an improved fuzzy model." The International Journal of Advanced Manufacturing Technology, 65(5-8), 1045–1053.

Zhu, W.-J., Feng, W.-F., and Zhou, Y.-G. (2010). "The Application of Genetic Fuzzy Neural Network in Project Cost Estimate." 2010 International Conference on E-Product E-Service and E-Entertainment.

# Appendices

# Appendix A: Field survey module

Please, based on Likert scale (5 points), select the most appropriate rate for each of the following parameters to evaluate each parameter affecting on the cost of FCIPs

Likert scale:

| Strongly Disagree | Disagree | Undecided | Agree | Strongly Agree |
|---|---|---|---|---|
| (1) | (2) | (3) | (4) | (5) |

| ID | Parameters categories | Parameters | Degree of Importance | | | | | Notes |
|---|---|---|---|---|---|---|---|---|
| | | | 1 | 2 | 3 | 4 | 5 | |
| P1 | Civil | Command Area (hectare) | | | | | | |
| P2 | Civil | PVC Length (m) | | | | | | |
| P3 | Civil | Construction year and inflation rate | | | | | | |
| P4 | Civil | Mesqa discharge ( capacity ) | | | | | | |
| P5 | Mechanical | Number of Irrigation Valves ( alfa-alfa valve ) | | | | | | |
| P6 | Civil | Consultant performance and errors in design | | | | | | |
| P7 | Electrical | Number of electrical pumps | | | | | | |
| P8 | Civil | PVC pipe diameter | | | | | | |
| P9 | Location | Orientation of mesqa ( intersecting with drains or roads or both) | | | | | | |
| P10 | Mechanical | Electrical and diesel pumps discharge | | | | | | |
| P11 | Civil | PC Intake , steel gate and  Pitching with cement mortar | | | | | | |
| P12 | Location | Type of mesqa (  Parallel to branch canal (Gannabya) , Perpendicular on branch canal) | | | | | | |
| P13 | miscellaneous | Farmers Objections | | | | | | |
| P14 | Electrical | Electrical consumption board type | | | | | | |
| P15 | Location | location of governorate (Al Sharqia , Dakahlia , …) | | | | | | |

| P16 | Civil | Pump house size  3m*3m  or 3m*4m | | | | | | |
|-----|-------|-----------------------------------|--|--|--|--|--|--|
| P17 | miscellaneous | cement  price | | | | | | |
| P18 | Mechanical | Head of electrical and diesel pumps | | | | | | |
| P19 | miscellaneous | Farmers adjustments | | | | | | |
| P20 | Civil | Sand filling | | | | | | |
| P21 | Civil | Sump size | | | | | | |
| P22 | Civil | Contractor performance and bad construction works | | | | | | |
| P23 | miscellaneous | pump price | | | | | | |
| P24 | Civil | Crops on submerged soils ( Rice) and its season (May to July) | | | | | | |
| P25 | miscellaneous | pipe price | | | | | | |
| P26 | Location | Topography and land levels of command area | | | | | | |
| P27 | Civil | Construction durations | | | | | | |
| P28 | Civil | Pumping and suction  pipes | | | | | | |
| P29 | Mechanical | Steel  mechanical connections | | | | | | |
| P30 | Civil | Difference between land and water levels | | | | | | |
| P31 | miscellaneous | steel price | | | | | | |
| P32 | Civil | Number of PVC branches | | | | | | |
| P33 | miscellaneous | Cash for damaged crops | | | | | | |
| P34 | Mechanical | Air / Pressure relief valve | | | | | | |
| P35 | miscellaneous | Crops on unsubmerged soils (wheat, corn, cotton, etc.) | | | | | | |

# Appendix B: Delphi Rounds

| | Respondents (Ri) | | | | | | | | | | | | | | | | |
|---|---|---|---|---|---|---|---|---|---|---|---|---|---|---|---|---|---|
| ID | R1 | R2 | R3 | R4 | R5 | R6 | R7 | R8 | R9 | R10 | R11 | R12 | R13 | R14 | R15 | Mean | SE |
| P1 | 5 | 5 | 5 | 5 | 5 | 5 | 4 | 5 | 5 | 5 | 5 | 5 | 5 | 5 | 5 | 4.97 | 0 |
| P2 | 4 | 4 | 4 | 4 | 4 | 5 | 4 | 4 | 5 | 5 | 5 | 4 | 5 | 5 | 5 | 4.43 | 0.1 |
| P3 | 5 | 5 | 4 | 4 | 4 | 5 | 5 | 5 | 5 | 5 | 5 | 3 | 4 | 3 | 2 | 4.33 | 0.2 |
| P4 | 4 | 5 | 4 | 5 | 4 | 2 | 4 | 4 | 5 | 4 | 5 | 5 | 4 | 3 | 4 | 4.33 | 0.2 |
| P5 | 4 | 5 | 5 | 5 | 5 | 5 | 4 | 5 | 2 | 3 | 3 | 5 | 4 | 4 | 2 | 4.2 | 0.2 |
| P6 | 5 | 5 | 5 | 5 | 5 | 4 | 3 | 4 | 4 | 4 | 5 | 4 | 3 | 4 | 2 | 4.13 | 0.2 |
| P7 | 5 | 5 | 5 | 5 | 5 | 4 | 3 | 4 | 4 | 4 | 5 | 4 | 3 | 4 | 2 | 4.1 | 0.2 |
| P8 | 5 | 5 | 5 | 5 | 5 | 2 | 2 | 4 | 5 | 5 | 4 | 4 | 5 | 1 | 4 | 3.9 | 0.2 |
| P9 | 4 | 4 | 3 | 4 | 3 | 4 | 2 | 4 | 5 | 3 | 5 | 4 | 4 | 4 | 4 | 3.8 | 0.2 |
| P10 | 5 | 3 | 4 | 5 | 5 | 1 | 4 | 3 | 4 | 4 | 4 | 2 | 4 | 5 | 5 | 3.57 | 0.1 |
| P11 | 5 | 4 | 4 | 4 | 4 | 3 | 2 | 4 | 4 | 3 | 5 | 4 | 3 | 3 | 4 | 3.4 | 0.2 |
| P12 | 5 | 5 | 3 | 2 | 3 | 1 | 4 | 3 | 4 | 4 | 4 | 1 | 4 | 4 | 5 | 3.07 | 0.2 |
| P13 | 5 | 5 | 3 | 2 | 3 | 1 | 4 | 3 | 4 | 4 | 4 | 1 | 4 | 4 | 5 | 2.93 | 0.2 |
| P14 | 4 | 3 | 3 | 4 | 4 | 4 | 2 | 4 | 4 | 3 | 4 | 3 | 2 | 4 | 4 | 2.87 | 0.2 |
| P15 | 3 | 4 | 4 | 4 | 4 | 3 | 2 | 4 | 4 | 1 | 3 | 4 | 3 | 1 | 1 | 2.63 | 0.2 |
| P16 | 1 | 3 | 3 | 1 | 1 | 4 | 2 | 1 | 3 | 5 | 4 | 5 | 4 | 5 | 2 | 2.6 | 0.2 |
| P17 | 5 | 4 | 4 | 5 | 4 | 3 | 3 | 2 | 2 | 1 | 3 | 1 | 1 | 4 | 2 | 2.53 | 0.2 |
| P18 | 4 | 4 | 4 | 5 | 3 | 1 | 2 | 3 | 3 | 4 | 3 | 3 | 2 | 2 | 2 | 2.53 | 0.2 |
| P19 | 2 | 4 | 2 | 4 | 4 | 1 | 2 | 5 | 3 | 3 | 2 | 3 | 2 | 2 | 2 | 2.5 | 0.2 |
| P20 | 3 | 4 | 2 | 2 | 3 | 2 | 4 | 1 | 3 | 1 | 2 | 2 | 3 | 4 | 4 | 2.4 | 0.2 |
| P21 | 3 | 4 | 2 | 2 | 3 | 4 | 2 | 4 | 2 | 4 | 2 | 1 | 2 | 2 | 2 | 2.37 | 0.2 |
| P22 | 2 | 2 | 2 | 2 | 3 | 3 | 2 | 4 | 2 | 5 | 2 | 5 | 2 | 1 | 1 | 2.2 | 0.2 |
| P23 | 2 | 3 | 4 | 3 | 3 | 4 | 2 | 4 | 2 | 1 | 2 | 2 | 2 | 2 | 2 | 2.13 | 0.2 |
| P24 | 2 | 2 | 2 | 2 | 2 | 3 | 4 | 4 | 2 | 4 | 2 | 4 | 2 | 1 | 2 | 2.1 | 0.2 |
| P25 | 2 | 2 | 2 | 5 | 2 | 2 | 2 | 3 | 2 | 2 | 2 | 4 | 2 | 4 | 2 | 2.1 | 0.2 |
| P26 | 2 | 3 | 3 | 3 | 5 | 1 | 2 | 3 | 2 | 2 | 2 | 2 | 2 | 2 | 2 | 2.1 | 0.1 |
| P27 | 3 | 3 | 2 | 3 | 3 | 1 | 2 | 4 | 2 | 4 | 2 | 1 | 2 | 2 | 1 | 2.07 | 0.2 |
| P28 | 1 | 3 | 4 | 3 | 4 | 3 | 1 | 3 | 1 | 2 | 1 | 2 | 1 | 5 | 1 | 2.07 | 0.2 |
| P29 | 4 | 1 | 3 | 2 | 2 | 1 | 2 | 4 | 1 | 1 | 2 | 5 | 3 | 1 | 1 | 2 | 0.2 |
| P30 | 2 | 1 | 1 | 1 | 3 | 4 | 3 | 1 | 3 | 4 | 3 | 2 | 2 | 1 | 2 | 1.9 | 0.1 |
| P31 | 4 | 1 | 1 | 1 | 2 | 1 | 2 | 4 | 2 | 1 | 2 | 4 | 4 | 1 | 1 | 1.8 | 0.2 |
| P32 | 2 | 3 | 1 | 2 | 4 | 1 | 2 | 1 | 2 | 3 | 4 | 1 | 2 | 1 | 2 | 1.8 | 0.2 |
| P33 | 2 | 2 | 2 | 2 | 1 | 1 | 3 | 4 | 1 | 3 | 3 | 1 | 2 | 1 | 2 | 1.73 | 0.1 |
| P34 | 4 | 2 | 2 | 2 | 1 | 2 | 1 | 2 | 1 | 4 | 1 | 1 | 2 | 1 | 2 | 1.6 | 0.2 |
| P35 | 2 | 1 | 1 | 1 | 4 | 2 | 1 | 3 | 2 | 2 | 1 | 2 | 2 | 1 | 2 | 1.5 | 0.1 |

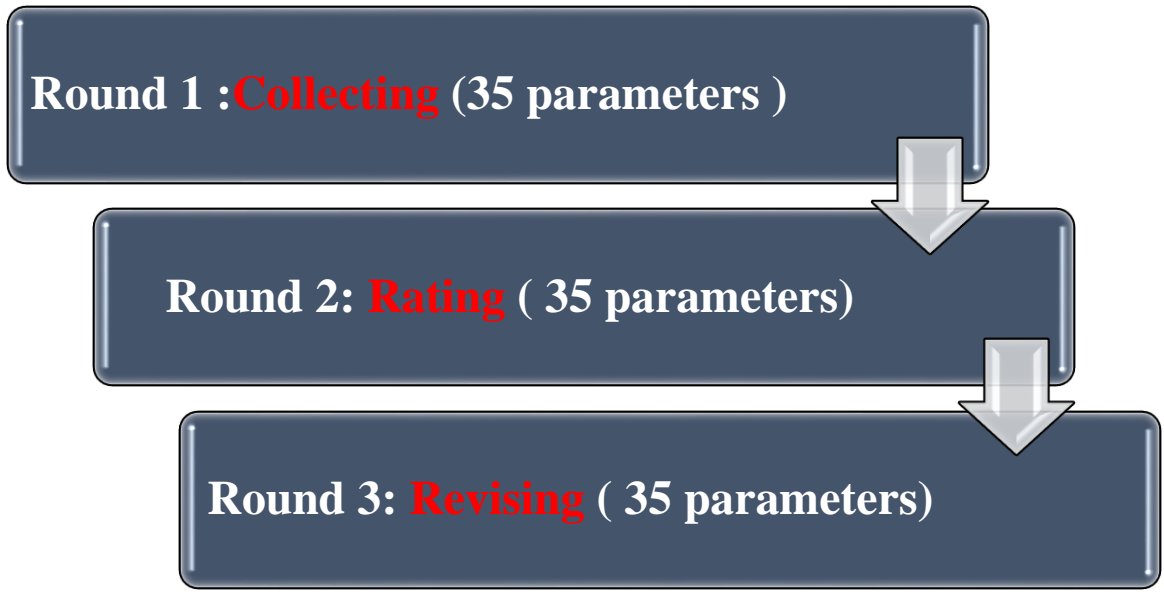

Fig. Delphi rounds.

**Appendix C: Collected data snap shot**

| ID | Area surved(P1) | Average area sections(P2) | Total length (P3) | Equavilant Diameters(P4) | Duration (P5) | Irr.valves(P6) |
|-----|-----------------|---------------------------|-------------------|--------------------------|---------------|----------------|
| M1 | 19 | 4.4 | 366 | 225 | 62 | 5 |
| M2 | 20 | 4.5 | 390 | 225 | 62 | 5 |
| M3 | 23 | 11.85 | 795 | 239 | 70 | 5 |
| M4 | 24 | 12 | 779 | 248 | 70 | 4 |
| M5 | 24 | 3.8 | 530 | 227 | 77 | 8 |
| M6 | 24 | 6 | 482 | 244 | 64 | 5 |
| M7 | 25 | 2.65 | 644 | 225 | 68 | 8 |
| M8 | 25 | 3 | 530 | 225 | 65 | 4 |
| M10 | 26 | 4.5 | 300 | 225 | 67 | 5 |
| M9 | 26 | 3.6 | 462 | 276 | 64 | 8 |
| M11 | 26.2 | 13.1 | 588 | 225 | 66 | 3 |
| M12 | 27 | 4.3 | 507 | 225 | 65 | 7 |
| M13 | 27 | 6 | 477 | 225 | 64 | 5 |

# Appendix D: Data for key cost drivers

## D.1 Training data

| ID | Area served(P1) | Total length (P3) | Irrigation valves(P6) | year(P14) | Total Cost LE / Mesqa |
|---|---|---|---|---|---|
| M1 | 19 | 366 | 5 | 2014 | 247632 |
| M2 | 20 | 390 | 5 | 2014 | 226870 |
| M3 | 23 | 795 | 5 | 2014 | 363098 |
| M4 | 24 | 779 | 4 | 2014 | 401454 |
| M5 | 24 | 530 | 8 | 2011 | 347855 |
| M6 | 24 | 482 | 5 | 2014 | 343827 |
| M7 | 25 | 644 | 8 | 2014 | 331076 |
| M8 | 25 | 530 | 4 | 2014 | 303024 |
| M10 | 26 | 300 | 5 | 2012 | 210616 |
| M9 | 26 | 462 | 8 | 2014 | 349523 |
| M11 | 26.2 | 588 | 3 | 2014 | 352094 |
| M12 | 27 | 507 | 7 | 2014 | 284942 |
| M13 | 27 | 477 | 5 | 2010 | 225198 |
| M14 | 27 | 470 | 5 | 2014 | 282934 |
| M15 | 27 | 400 | 7 | 2014 | 288227 |
| M16 | 27 | 299 | 6 | 2014 | 268368 |
| M17 | 28 | 327 | 5 | 2014 | 302181 |
| M18 | 28 | 280 | 5 | 2013 | 229054 |
| M19 | 28 | 280 | 5 | 2015 | 266404 |
| M20 | 28 | 280 | 5 | 2015 | 285292 |
| M21 | 28 | 198 | 5 | 2014 | 214734 |
| M22 | 29 | 448 | 7 | 2014 | 330764 |
| M23 | 29 | 330 | 6 | 2012 | 224449 |
| M24 | 30 | 779 | 3 | 2014 | 321717 |
| M25 | 30 | 655 | 5 | 2013 | 324866 |
| M26 | 30 | 400 | 5 | 2014 | 317470 |
| M27 | 31 | 774 | 11 | 2011 | 382645 |
| M28 | 32 | 765 | 8 | 2011 | 276584 |
| M29 | 32 | 630 | 6 | 2014 | 353089 |
| M30 | 32 | 600 | 6 | 2010 | 288882 |
| M31 | 32 | 251 | 4 | 2014 | 257528 |
| M32 | 34 | 870 | 10 | 2014 | 402796 |
| M33 | 34 | 492 | 5 | 2014 | 339042 |
| M34 | 34 | 468 | 5 | 2014 | 329059 |
| M35 | 34 | 400 | 4 | 2014 | 265775 |
| M36 | 35 | 377 | 6 | 2012 | 269297 |

**D.1: Training data**

| ID | Area served(P1) | Total length (P3) | Irrigation valves(P6) | year(P14) | Total Cost LE / Mesqa |
|----|-----------------|-------------------|-----------------------|-----------|------------------------|
| M37 | 36 | 750 | 8 | 2014 | 334916 |
| M38 | 37 | 750 | 6 | 2011 | 269103 |
| M39 | 37 | 674 | 9 | 2012 | 343534 |
| M40 | 37 | 417 | 7 | 2010 | 267104 |
| M41 | 38 | 1033 | 6 | 2014 | 364033 |
| M42 | 38 | 530 | 7 | 2010 | 246388 |
| M43 | 39 | 1135 | 10 | 2014 | 415018 |
| M44 | 39 | 505 | 8 | 2014 | 328105 |
| M45 | 39 | 401 | 5 | 2014 | 330347 |
| M46 | 40 | 1040 | 11 | 2014 | 572213 |
| M47 | 40 | 1000 | 3 | 2014 | 427421 |
| M48 | 40 | 870 | 10 | 2014 | 397017 |
| M49 | 41 | 850 | 8 | 2014 | 430067 |
| M50 | 41 | 532 | 8 | 2012 | 250097 |
| M51 | 41 | 321 | 5 | 2014 | 240469 |
| M52 | 43 | 800 | 8 | 2010 | 346772 |
| M53 | 43 | 544 | 6 | 2014 | 310335 |
| M54 | 45 | 992 | 9 | 2014 | 467948 |
| M55 | 45 | 850 | 7 | 2013 | 392516 |
| M56 | 45 | 630 | 7 | 2013 | 323702 |
| M57 | 45 | 616 | 9 | 2014 | 434407 |
| M58 | 45 | 610 | 7 | 2013 | 310025 |
| M59 | 45 | 603 | 6 | 2014 | 311572 |
| M60 | 46 | 1390 | 10 | 2014 | 528668 |
| M61 | 46 | 471 | 7 | 2015 | 386043 |
| M62 | 46 | 470 | 7 | 2014 | 311473 |
| M63 | 47 | 700 | 6 | 2014 | 391094 |
| M64 | 50 | 1275 | 12 | 2010 | 480177 |
| M65 | 50 | 1020 | 9 | 2010 | 306053 |
| M66 | 50 | 870 | 13 | 2014 | 379835 |
| M67 | 50 | 600 | 11 | 2011 | 313996 |
| M68 | 50 | 310 | 7 | 2015 | 346180 |
| M69 | 51 | 1730 | 7 | 2014 | 677433 |
| M70 | 51 | 455 | 6 | 2014 | 319335 |
| M71 | 51 | 365 | 5 | 2012 | 209795 |
| M72 | 51 | 359 | 7 | 2013 | 313156 |
| M73 | 51 | 312 | 7 | 2013 | 253404 |
| M74 | 52 | 1393 | 13 | 2014 | 588634 |

**D.1: Training data**

| ID | Area served(P1) | Total length (P3) | Irrigation valves(P6) | year(P14) | Total Cost LE / Mesqa |
|---|---|---|---|---|---|
| M75 | 52 | 1100 | 13 | 2014 | 526920 |
| M76 | 52 | 439 | 5 | 2014 | 304224 |
| M77 | 53 | 1325 | 11 | 2013 | 623674 |
| M78 | 53 | 1200 | 27 | 2013 | 603681 |
| M79 | 53 | 570 | 9 | 2014 | 436038 |
| M80 | 54 | 1179 | 10 | 2014 | 412340 |
| M81 | 56 | 1832 | 3 | 2014 | 603733 |
| M82 | 56 | 850 | 8 | 2011 | 385871 |
| M83 | 56 | 720 | 9 | 2012 | 378414 |
| M84 | 56 | 700 | 9 | 2015 | 387368 |
| M85 | 58 | 1310 | 9 | 2011 | 501687 |
| M86 | 60 | 1258 | 11 | 2013 | 501560 |
| M87 | 61 | 1018 | 12 | 2012 | 434545 |
| M88 | 62 | 1290 | 13 | 2015 | 630254 |
| M89 | 62 | 1091 | 10 | 2015 | 513023 |
| M90 | 63 | 1499 | 12 | 2013 | 664207 |
| M91 | 65 | 875 | 15 | 2010 | 347098 |
| M92 | 67 | 1150 | 10 | 2011 | 468006 |
| M93 | 67 | 1150 | 11 | 2011 | 428314 |
| M94 | 68 | 664 | 7 | 2014 | 396737 |
| M95 | 70 | 908 | 10 | 2014 | 620577 |
| M96 | 70 | 870 | 11 | 2010 | 350214 |
| M97 | 70 | 119 | 10 | 2015 | 349062 |
| M98 | 71 | 960 | 9 | 2014 | 401864 |
| M99 | 71 | 956 | 9 | 2010 | 397841 |
| M100 | 74 | 1675 | 11 | 2014 | 705904 |
| M101 | 76 | 1300 | 12 | 2014 | 593994 |
| M102 | 76 | 1300 | 12 | 2014 | 592582 |
| M103 | 77 | 1393 | 17 | 2014 | 701507 |
| M104 | 79 | 909 | 12 | 2012 | 476151 |
| M105 | 80 | 1399 | 12 | 2011 | 507948 |
| M106 | 80 | 934 | 9 | 2014 | 597031 |
| M107 | 88 | 1095 | 6 | 2014 | 414148 |
| M108 | 88 | 1095 | 15 | 2014 | 444104 |
| M109 | 90 | 721 | 13 | 2015 | 667167 |
| M110 | 97 | 1140 | 4 | 2014 | 551958 |
| M111 | 100 | 1481 | 15 | 2014 | 616467 |

**D.2: Validation data**

| ID | Area served (P$_1$) | Total length (P$_3$) | Irrigation. Valves (P$_6$) | Year (P$_{14}$) | Total Cost LE / Mesqa |
|---|---|---|---|---|---|
| **Validation data** | | | | | |
| **M112** | 78.1 | 1051.6 | 9 | 2010 | 425690 |
| **M113** | 29.7 | 440 | 7 | 2014 | 308403 |
| **M114** | 45.1 | 935 | 8 | 2014 | 460172 |
| **M115** | 22 | 429 | 5 | 2014 | 242751 |
| **M116** | 37.4 | 957 | 10 | 2014 | 430992 |
| **M117** | 49.5 | 663.3 | 6 | 2014 | 333382 |
| **M118** | 88 | 1027.4 | 9 | 2014 | 638823 |
| **M119** | 37.4 | 541.2 | 5 | 2014 | 362775 |
| **M120** | 57.2 | 1210 | 13 | 2014 | 563805 |
| **M121** | 30.8 | 308 | 5 | 2015 | 305263 |
| **M122** | 41.8 | 1136.3 | 6 | 2014 | 389516 |
| **M123** | 61.6 | 2015.2 | 3 | 2014 | 645995 |
| **M124** | 58.3 | 627 | 9 | 2014 | 466561 |
| **M125** | 49.5 | 677.6 | 9 | 2014 | 464815 |
| **M126** | 73.7 | 1265 | 11 | 2011 | 458295 |
| **M127** | 45.1 | 585.2 | 8 | 2012 | 267604 |
| **M128** | 55 | 660 | 11 | 2011 | 335976 |
| **M129** | 58.3 | 1320 | 27 | 2013 | 645938 |
| **M130** | 96.8 | 1204.5 | 6 | 2014 | 443139 |
| **M131** | 49.5 | 693 | 7 | 2013 | 346361 |
| **M132** | 40.7 | 458.7 | 7 | 2010 | 285802 |
| **M133** | 66 | 1383.8 | 11 | 2013 | 536669 |
| **M134** | 50.6 | 517 | 7 | 2014 | 333277 |
| **M135** | 49.5 | 1091.2 | 9 | 2014 | 500704 |
| **M136** | 42.9 | 555.5 | 8 | 2014 | 351073 |
| **M137** | 84.7 | 1532.3 | 17 | 2014 | 750612 |
| **M138** | 110 | 1629.1 | 15 | 2014 | 659620 |
| **M139** | 30.8 | 308 | 5 | 2015 | 285052 |
| **M140** | 27.5 | 583 | 4 | 2014 | 324236 |
| **M141** | 41.8 | 583 | 7 | 2010 | 263636 |
| **M142** | 30.8 | 308 | 5 | 2013 | 245087 |
| **M143** | 73.7 | 1265 | 10 | 2011 | 500766 |
| **M144** | 55 | 1402.5 | 12 | 2010 | 513789 |

# Appendix E: Excel VBA Code for cost model application

```vba
Private Sub CheckBox1_Click()
If CheckBox1 = True Then
ThisWorkbook.Sheets("GUIR").Cells(24, 2).Value = "TRUE"
TextBox9.Value = ThisWorkbook.Sheets("GUIR").Range("H9").Value
TextBox10.Value = Round(ThisWorkbook.Sheets("GUIR").Range("I9").Value)
End If
If CheckBox1 = False Then
ThisWorkbook.Sheets("GUIR").Cells(24, 2).Value = "FALSE"
TextBox9.Value = ""
TextBox10.Value = ""
End If
End Sub
Private Sub CheckBox2_Click()
If CheckBox2 = True Then
ThisWorkbook.Sheets("GUIR").Cells(24, 3).Value = "TRUE"
TextBox9.Value = ThisWorkbook.Sheets("GUIR").Range("H9").Value
TextBox10.Value = Round(ThisWorkbook.Sheets("GUIR").Range("I9").Value)
End If
If CheckBox2 = False Then
ThisWorkbook.Sheets("GUIR").Cells(24, 3).Value = "FALSE"
TextBox9.Value = ""
TextBox10.Value = ""
End If
End Sub
Private Sub CheckBox3_Click()
If CheckBox3 = True Then
ThisWorkbook.Sheets("GUIR").Cells(24, 4).Value = "TRUE"
TextBox9.Value = ThisWorkbook.Sheets("GUIR").Range("H9").Value
TextBox10.Value = Round(ThisWorkbook.Sheets("GUIR").Range("I9").Value)
End If
If CheckBox3 = False Then
ThisWorkbook.Sheets("GUIR").Cells(24, 4).Value = "FALSE"
TextBox9.Value = ""
TextBox10.Value = ""
End If
End Sub
Private Sub CheckBox4_Click()
```

```vba
If CheckBox4 = True Then
ThisWorkbook.Sheets("GUIR").Cells(24, 5).Value = "TRUE"
TextBox9.Value = ThisWorkbook.Sheets("GUIR").Range("H9").Value
TextBox10.Value = Round(ThisWorkbook.Sheets("GUIR").Range("I9").Value)
End If
If CheckBox4 = False Then
ThisWorkbook.Sheets("GUIR").Cells(24, 5).Value = "FALSE"
TextBox9.Value = ""
TextBox10.Value = ""
End If
End Sub
Private Sub CommandButton1_Click()
Dim L  As Double
Dim A As Double
Dim N As Double
Dim Y As Double
Dim C As Double
' after 2015 variables
' NT nuber of years
'IR inflation Rate
' FC Future Cost
Dim NY As Double
Dim IR As Double
Dim FC As Double
Dim NY2015 As Double
L = TextBox1.Value
A = TextBox2.Value
N = TextBox3.Value
Y = TextBox4.Value
C = Round((-37032.81 + L * 0.1691+ A * 2.21 + N * 2.265+ Y * 18.594)^2)
TextBox5.Value = C
TextBox8.Value = Round(C / A)
' to excel
ThisWorkbook.Sheets("GUIR").Range("B31").Value = A
ThisWorkbook.Sheets("GUIR").Range("c31").Value = L
ThisWorkbook.Sheets("GUIR").Range("d31").Value = N
ThisWorkbook.Sheets("GUIR").Range("e31").Value = Y
ThisWorkbook.Sheets("GUIR").Range("f31").Value = C
'after form adjustment
If TextBox4.Value > 2015 Then
```

```vba
NY2015 = (TextBox4.Value - 2015)
IR = TextBox6.Value
FC = C * ((1 + IR / 100) ^ NY2015)
TextBox5.Value = Round(FC)
TextBox8.Value = Round(FC / A)
' to excel
ThisWorkbook.Sheets("GUIR").Range("B31").Value = A
ThisWorkbook.Sheets("GUIR").Range("c31").Value = L
ThisWorkbook.Sheets("GUIR").Range("d31").Value = N
ThisWorkbook.Sheets("GUIR").Range("e31").Value = Y
ThisWorkbook.Sheets("GUIR").Range("f31").Value = FC
ThisWorkbook.Sheets("GUIR").Range("I27").Value = NY2015
ThisWorkbook.Sheets("GUIR").Range("I28").Value = IR
End If
End Sub
Private Sub CommandButton1_Click()
UserForm1.Show
'ThisWorkbook.Sheets("GUIR").Cells(24, 2).Value = "FALSE"
'ThisWorkbook.Sheets("GUIR").Cells(24, 3).Value = "FALSE"
'ThisWorkbook.Sheets("GUIR").Cells(24, 4).Value = "FALSE"
'ThisWorkbook.Sheets("GUIR").Cells(24, 5).Value = "FALSE"
End Sub
```

## Appendix F: Automated fuzzy rules generation (R programming).

```
D4 <- Data_4
D4
data.train <- D4[1 : 80, ]
data.tst <- D4[80 : 111, 1:4 ]
real.val <- matrix(D4[80 : 111, 5], ncol = 1)
options(max.print=999999)
## Define range of input data. Note that it is only for the input variables.
range.data <- apply(data.train, 2, range)
method.type <- "SBC"
control <- list(num.labels = 6, type.mf = "GAUSSIAN", type.defuz = "WAM",
            type.tnorm = "MIN", type.snorm = "MAX", type.implication.func =
"ZADEH",
            name = "sim-0")
object.reg <- frbs.learn(data.train, range.data, method.type, control)
res.test <- predict(object.reg, data.tst)
## Display the FRBS model
summary(object.reg)
## Plot the membership functions
par(mar = rep(2, 4))
plotMF(object.reg)
newdata <- data.tst
## generate the model and save it as object.WM
object.SBC <- frbs.learn(data.train, range.data, method.type, control)
## the prediction process
## The following code can be used for all methods
res <- predict(object.SBC, newdata)
res
> ## Display the FRBS model
> summary(object.reg)
The name of model:  sim-0
Model was trained using:  WM
The names of attributes:  Area surved(P1) Total length (P3) Irr.valves(P6)
year(P14) Total Cost LE / Mesqa
The interval of training data:
```

Area surved(P1) Total length (P3) Irr.valves(P6) year(P14) Total Cost LE / Mesqa

|      | P1 | P3   | P6 | P14  | Total Cost |
|------|-----|------|-----|------|-----------|
| min  | 19  | 198  | 3   | 2010 | 209795    |
| max  | 54  | 1730 | 27  | 2015 | 677433    |

Type of FRBS model:
[1] "MAMDANI"
Type of membership functions:
[1] "GAUSSIAN"
Type of t-norm method:
[1] "Standard t-norm (min)"
Type of s-norm method:
[1] "Standard s-norm"
Type of defuzzification technique:
[1] "Weighted average method"
Type of implication function:
[1] "ZADEH"
The names of linguistic terms on the input variables:
 [1] "v.1_a.1" "v.1_a.2" "v.1_a.3" "v.1_a.4" "v.1_a.5" "v.1_a.6" "v.2_a.1" "v.2_a.2"
 [9] "v.2_a.3" "v.2_a.4" "v.2_a.5" "v.2_a.6" "v.3_a.1" "v.3_a.2" "v.3_a.3" "v.3_a.4"
[17] "v.3_a.5" "v.3_a.6" "v.4_a.1" "v.4_a.2" "v.4_a.3" "v.4_a.4" "v.4_a.5" "v.4_a.6"
The parameter values of membership function on the input variable (normalized):

|      | v.1_a.1 | v.1_a.2 | v.1_a.3 | v.1_a.4 | v.1_a.5 | v.1_a.6 | v.2_a.1 | v.2_a.2 | v.2_a.3 | v.2_a.4 |
|------|---------|---------|---------|---------|---------|---------|---------|---------|---------|---------|
| [1,] | 5.00    | 5.00    | 5.00    | 5.00    | 5.00    | 5.00    | 5.00    | 5.00    | 5.00    | 5.00    |
| [2,] | 0.00    | 0.20    | 0.40    | 0.60    | 0.80    | 1.00    | 0.00    | 0.20    | 0.40    | 0.60    |
| [3,] | 0.07    | 0.07    | 0.07    | 0.07    | 0.07    | 0.07    | 0.07    | 0.07    | 0.07    | 0.07    |
| [4,] | NA      | NA      | NA      | NA      | NA      | NA      | NA      | NA      | NA      | NA      |
| [5,] | NA      | NA      | NA      | NA      | NA      | NA      | NA      | NA      | NA      | NA      |

|      | v.2_a.5 | v.2_a.6 | v.3_a.1 | v.3_a.2 | v.3_a.3 | v.3_a.4 | v.3_a.5 | v.3_a.6 | v.4_a.1 | v.4_a.2 |
|------|---------|---------|---------|---------|---------|---------|---------|---------|---------|---------|
| [1,] | 5.00    | 5.00    | 5.00    | 5.00    | 5.00    | 5.00    | 5.00    | 5.00    | 5.00    | 5.00    |
| [2,] | 0.80    | 1.00    | 0.00    | 0.20    | 0.40    | 0.60    | 0.80    | 1.00    | 0.00    | 0.20    |
| [3,] | 0.07    | 0.07    | 0.07    | 0.07    | 0.07    | 0.07    | 0.07    | 0.07    | 0.07    | 0.07    |
| [4,] | NA      | NA      | NA      | NA      | NA      | NA      | NA      | NA      | NA      | NA      |
| [5,] | NA      | NA      | NA      | NA      | NA      | NA      | NA      | NA      | NA      | NA      |

    v.4_a.3 v.4_a.4 v.4_a.5 v.4_a.6

```
[1,]    5.00    5.00    5.00    5.00
[2,]    0.40    0.60    0.80    1.00
[3,]    0.07    0.07    0.07    0.07
[4,]    NA      NA      NA      NA
[5,]    NA      NA      NA      NA
```
The names of linguistic terms on the output variable:
[1] "c.1" "c.2" "c.3" "c.4" "c.5" "c.6"
The parameter values of membership function on the output variable (normalized):
```
      c.1  c.2  c.3  c.4  c.5  c.6
[1,] 5.00 5.00 5.00 5.00 5.00 5.00
[2,] 0.00 0.20 0.40 0.60 0.80 1.00
[3,] 0.07 0.07 0.07 0.07 0.07 0.07
[4,] NA   NA   NA   NA   NA   NA
[5,] NA   NA   NA   NA   NA   NA
```
The number of linguistic terms on each variables

| Area surved(P1) | Total length (P3) | Irr.valves(P6) | year(P14) | Total Cost LE / Mesqa |
|---|---|---|---|---|
| [1,]     6 | 6 | 6  6 | | 6 |

v.1_a.5 = membership 5 in variable 1

The fuzzy IF-THEN rules:

| V1 | V2 V3 | V4 V5 | V6 V7 | V8 V9 | V10 |
|---|---|---|---|---|---|

1  IF Area surved(P1) is v.1_a.5 and Total length (P3) is v.2_a.2 and Irr.valves(P6)
2  IF Area surved(P1) is v.1_a.5 and Total length (P3) is v.2_a.2 and Irr.valves(P6)
3  IF Area surved(P1) is v.1_a.2 and Total length (P3) is v.2_a.2 and Irr.valves(P6)
4  IF Area surved(P1) is v.1_a.2 and Total length (P3) is v.2_a.2 and Irr.valves(P6)
5  IF Area surved(P1) is v.1_a.4 and Total length (P3) is v.2_a.2 and Irr.valves(P6)
6  IF Area surved(P1) is v.1_a.6 and Total length (P3) is v.2_a.4 and Irr.valves(P6)
7  IF Area surved(P1) is v.1_a.2 and Total length (P3) is v.2_a.3 and Irr.valves(P6)
8  IF Area surved(P1) is v.1_a.5 and Total length (P3) is v.2_a.3 and Irr.valves(P6)
9  IF Area surved(P1) is v.1_a.6 and Total length (P3) is v.2_a.5 and Irr.valves(P6)
10 IF Area surved(P1) is v.1_a.3 and Total length (P3) is v.2_a.3 and
Irr.valves(P6)
11 IF Area surved(P1) is v.1_a.3 and Total length (P3) is v.2_a.3 and
Irr.valves(P6)
12 IF Area surved(P1) is v.1_a.4 and Total length (P3) is v.2_a.4 and
Irr.valves(P6)

13 IF Area surved(P1) is v.1_a.5 and Total length (P3) is v.2_a.2 and Irr.valves(P6)

14 IF Area surved(P1) is v.1_a.4 and Total length (P3) is v.2_a.3 and Irr.valves(P6)

15 IF Area surved(P1) is v.1_a.3 and Total length (P3) is v.2_a.2 and Irr.valves(P6)

16 IF Area surved(P1) is v.1_a.6 and Total length (P3) is v.2_a.1 and Irr.valves(P6)

17 IF Area surved(P1) is v.1_a.4 and Total length (P3) is v.2_a.3 and Irr.valves(P6)

18 IF Area surved(P1) is v.1_a.5 and Total length (P3) is v.2_a.1 and Irr.valves(P6)

19 IF Area surved(P1) is v.1_a.2 and Total length (P3) is v.2_a.1 and Irr.valves(P6)

20 IF Area surved(P1) is v.1_a.3 and Total length (P3) is v.2_a.2 and Irr.valves(P6)

21 IF Area surved(P1) is v.1_a.4 and Total length (P3) is v.2_a.4 and Irr.valves(P6)

22 IF Area surved(P1) is v.1_a.5 and Total length (P3) is v.2_a.3 and Irr.valves(P6)

23 IF Area surved(P1) is v.1_a.4 and Total length (P3) is v.2_a.4 and Irr.valves(P6)

24 IF Area surved(P1) is v.1_a.6 and Total length (P3) is v.2_a.4 and Irr.valves(P6)

25 IF Area surved(P1) is v.1_a.4 and Total length (P3) is v.2_a.2 and Irr.valves(P6)

26 IF Area surved(P1) is v.1_a.4 and Total length (P3) is v.2_a.3 and Irr.valves(P6)

27 IF Area surved(P1) is v.1_a.3 and Total length (P3) is v.2_a.2 and Irr.valves(P6)

28 IF Area surved(P1) is v.1_a.5 and Total length (P3) is v.2_a.2 and Irr.valves(P6)

29 IF Area surved(P1) is v.1_a.5 and Total length (P3) is v.2_a.4 and Irr.valves(P6)

30 IF Area surved(P1) is v.1_a.5 and Total length (P3) is v.2_a.5 and Irr.valves(P6)

31 IF Area surved(P1) is v.1_a.6 and Total length (P3) is v.2_a.2 and Irr.valves(P6)

32 IF Area surved(P1) is v.1_a.1 and Total length (P3) is v.2_a.2 and Irr.valves(P6)

33 IF Area surved(P1) is v.1_a.6 and Total length (P3) is v.2_a.5 and Irr.valves(P6)

34 IF Area surved(P1) is v.1_a.2 and Total length (P3) is v.2_a.1 and Irr.valves(P6)

35 IF Area surved(P1) is v.1_a.4 and Total length (P3) is v.2_a.2 and Irr.valves(P6)

36 IF Area surved(P1) is v.1_a.4 and Total length (P3) is v.2_a.2 and Irr.valves(P6)

37 IF Area surved(P1) is v.1_a.4 and Total length (P3) is v.2_a.3 and Irr.valves(P6)

38 IF Area surved(P1) is v.1_a.3 and Total length (P3) is v.2_a.2 and Irr.valves(P6)

39 IF Area surved(P1) is v.1_a.2 and Total length (P3) is v.2_a.2 and Irr.valves(P6)

40 IF Area surved(P1) is v.1_a.2 and Total length (P3) is v.2_a.2 and Irr.valves(P6)

41 IF Area surved(P1) is v.1_a.3 and Total length (P3) is v.2_a.1 and Irr.valves(P6)

42 IF Area surved(P1) is v.1_a.3 and Total length (P3) is v.2_a.3 and Irr.valves(P6)

43 IF Area surved(P1) is v.1_a.4 and Total length (P3) is v.2_a.1 and Irr.valves(P6)

44 IF Area surved(P1) is v.1_a.4 and Total length (P3) is v.2_a.2 and Irr.valves(P6)

45 IF Area surved(P1) is v.1_a.3 and Total length (P3) is v.2_a.2 and Irr.valves(P6)

46 IF Area surved(P1) is v.1_a.2 and Total length (P3) is v.2_a.1 and Irr.valves(P6)

47 IF Area surved(P1) is v.1_a.3 and Total length (P3) is v.2_a.3 and Irr.valves(P6)

48 IF Area surved(P1) is v.1_a.5 and Total length (P3) is v.2_a.3 and Irr.valves(P6)

49 IF Area surved(P1) is v.1_a.6 and Total length (P3) is v.2_a.2 and Irr.valves(P6)

50 IF Area surved(P1) is v.1_a.6 and Total length (P3) is v.2_a.4 and Irr.valves(P6)

51 IF Area surved(P1) is v.1_a.2 and Total length (P3) is v.2_a.2 and Irr.valves(P6)

52 IF Area surved(P1) is v.1_a.2 and Total length (P3) is v.2_a.1 and Irr.valves(P6)

53 IF Area surved(P1) is v.1_a.5 and Total length (P3) is v.2_a.2 and Irr.valves(P6)

54 IF Area surved(P1) is v.1_a.5 and Total length (P3) is v.2_a.5 and Irr.valves(P6)

55 IF Area surved(P1) is v.1_a.6 and Total length (P3) is v.2_a.2 and Irr.valves(P6)

56 IF Area surved(P1) is v.1_a.3 and Total length (P3) is v.2_a.3 and Irr.valves(P6)

57 IF Area surved(P1) is v.1_a.2 and Total length (P3) is v.2_a.1 and Irr.valves(P6)

58 IF Area surved(P1) is v.1_a.5 and Total length (P3) is v.2_a.4 and Irr.valves(P6)

59 IF Area surved(P1) is v.1_a.2 and Total length (P3) is v.2_a.1 and Irr.valves(P6)

60 IF Area surved(P1) is v.1_a.6 and Total length (P3) is v.2_a.2 and Irr.valves(P6)

61 IF Area surved(P1) is v.1_a.2 and Total length (P3) is v.2_a.1 and Irr.valves(P6)

62 IF Area surved(P1) is v.1_a.6 and Total length (P3) is v.2_a.6 and Irr.valves(P6)

63 IF Area surved(P1) is v.1_a.6 and Total length (P3) is v.2_a.2 and Irr.valves(P6)

| V11 | V12 | V13 | V14 | V15 | V16 | V17 | V18 | V19 | V20 |
|---|---|---|---|---|---|---|---|---|---|
| 1 | is v.3_a.2 and year(P14) | | is v.4_a.6 | THEN Total Cost LE / Mesqa | | | is c.3 | | |
| 2 | is v.3_a.2 and year(P14) | | is v.4_a.5 | THEN Total Cost LE / Mesqa | | | is c.2 | | |
| 3 | is v.3_a.2 and year(P14) | | is v.4_a.5 | THEN Total Cost LE / Mesqa | | | is c.2 | | |
| 4 | is v.3_a.1 and year(P14) | | is v.4_a.5 | THEN Total Cost LE / Mesqa | | | is c.2 | | |
| 5 | is v.3_a.2 and year(P14) | | is v.4_a.5 | THEN Total Cost LE / Mesqa | | | is c.2 | | |
| 6 | is v.3_a.6 and year(P14) | | is v.4_a.4 | THEN Total Cost LE / Mesqa | | | is c.5 | | |

7   is v.3_a.1 and year(P14)  is v.4_a.5 THEN Total Cost LE / Mesqa  is c.3
8   is v.3_a.2 and year(P14)  is v.4_a.4 THEN Total Cost LE / Mesqa  is c.3
9   is v.3_a.3 and year(P14)  is v.4_a.5 THEN Total Cost LE / Mesqa  is c.5
10  is v.3_a.2 and year(P14)  is v.4_a.2 THEN Total Cost LE / Mesqa  is c.2
11  is v.3_a.3 and year(P14)  is v.4_a.2 THEN Total Cost LE / Mesqa  is c.3
12  is v.3_a.3 and year(P14)  is v.4_a.5 THEN Total Cost LE / Mesqa  is c.5
13  is v.3_a.2 and year(P14)  is v.4_a.4 THEN Total Cost LE / Mesqa  is c.2
14  is v.3_a.2 and year(P14)  is v.4_a.5 THEN Total Cost LE / Mesqa  is c.3
15  is v.3_a.2 and year(P14)  is v.4_a.5 THEN Total Cost LE / Mesqa  is c.3
16  is v.3_a.2 and year(P14)  is v.4_a.4 THEN Total Cost LE / Mesqa  is c.1
17  is v.3_a.2 and year(P14)  is v.4_a.1 THEN Total Cost LE / Mesqa  is c.2
18  is v.3_a.2 and year(P14)  is v.4_a.6 THEN Total Cost LE / Mesqa  is c.2
19  is v.3_a.2 and year(P14)  is v.4_a.5 THEN Total Cost LE / Mesqa  is c.2
20  is v.3_a.2 and year(P14)  is v.4_a.1 THEN Total Cost LE / Mesqa  is c.2
21  is v.3_a.2 and year(P14)  is v.4_a.5 THEN Total Cost LE / Mesqa  is c.3
22  is v.3_a.2 and year(P14)  is v.4_a.5 THEN Total Cost LE / Mesqa  is c.3
23  is v.3_a.1 and year(P14)  is v.4_a.5 THEN Total Cost LE / Mesqa  is c.3
24  is v.3_a.3 and year(P14)  is v.4_a.5 THEN Total Cost LE / Mesqa  is c.4
25  is v.3_a.2 and year(P14)  is v.4_a.1 THEN Total Cost LE / Mesqa  is c.1
26  is v.3_a.2 and year(P14)  is v.4_a.3 THEN Total Cost LE / Mesqa  is c.2
27  is v.3_a.1 and year(P14)  is v.4_a.5 THEN Total Cost LE / Mesqa  is c.2
28  is v.3_a.2 and year(P14)  is v.4_a.5 THEN Total Cost LE / Mesqa  is c.3
29  is v.3_a.2 and year(P14)  is v.4_a.5 THEN Total Cost LE / Mesqa  is c.4
30  is v.3_a.2 and year(P14)  is v.4_a.5 THEN Total Cost LE / Mesqa  is c.4
31  is v.3_a.2 and year(P14)  is v.4_a.5 THEN Total Cost LE / Mesqa  is c.3
32  is v.3_a.1 and year(P14)  is v.4_a.5 THEN Total Cost LE / Mesqa  is c.1
33  is v.3_a.3 and year(P14)  is v.4_a.4 THEN Total Cost LE / Mesqa  is c.5
34  is v.3_a.1 and year(P14)  is v.4_a.6 THEN Total Cost LE / Mesqa  is c.2
35  is v.3_a.2 and year(P14)  is v.4_a.3 THEN Total Cost LE / Mesqa  is c.1
36  is v.3_a.2 and year(P14)  is v.4_a.1 THEN Total Cost LE / Mesqa  is c.2
37  is v.3_a.2 and year(P14)  is v.4_a.2 THEN Total Cost LE / Mesqa  is c.2
38  is v.3_a.2 and year(P14)  is v.4_a.3 THEN Total Cost LE / Mesqa  is c.2
39  is v.3_a.2 and year(P14)  is v.4_a.2 THEN Total Cost LE / Mesqa  is c.2
40  is v.3_a.1 and year(P14)  is v.4_a.5 THEN Total Cost LE / Mesqa  is c.3
41  is v.3_a.1 and year(P14)  is v.4_a.5 THEN Total Cost LE / Mesqa  is c.2
42  is v.3_a.2 and year(P14)  is v.4_a.5 THEN Total Cost LE / Mesqa  is c.2
43  is v.3_a.1 and year(P14)  is v.4_a.5 THEN Total Cost LE / Mesqa  is c.1

44  is v.3_a.1 and year(P14)  is v.4_a.5 THEN Total Cost LE / Mesqa  is c.2
45  is v.3_a.1 and year(P14)  is v.4_a.4 THEN Total Cost LE / Mesqa  is c.2
46  is v.3_a.1 and year(P14)  is v.4_a.4 THEN Total Cost LE / Mesqa  is c.1
47  is v.3_a.1 and year(P14)  is v.4_a.5 THEN Total Cost LE / Mesqa  is c.2
48  is v.3_a.3 and year(P14)  is v.4_a.5 THEN Total Cost LE / Mesqa  is c.3
49  is v.3_a.2 and year(P14)  is v.4_a.5 THEN Total Cost LE / Mesqa  is c.2
50  is v.3_a.2 and year(P14)  is v.4_a.5 THEN Total Cost LE / Mesqa  is c.3
51  is v.3_a.1 and year(P14)  is v.4_a.1 THEN Total Cost LE / Mesqa  is c.1
52  is v.3_a.2 and year(P14)  is v.4_a.3 THEN Total Cost LE / Mesqa  is c.1
53  is v.3_a.3 and year(P14)  is v.4_a.2 THEN Total Cost LE / Mesqa  is c.2
54  is v.3_a.3 and year(P14)  is v.4_a.1 THEN Total Cost LE / Mesqa  is c.4
55  is v.3_a.2 and year(P14)  is v.4_a.4 THEN Total Cost LE / Mesqa  is c.2
56  is v.3_a.2 and year(P14)  is v.4_a.5 THEN Total Cost LE / Mesqa  is c.3
57  is v.3_a.1 and year(P14)  is v.4_a.5 THEN Total Cost LE / Mesqa  is c.1
58  is v.3_a.2 and year(P14)  is v.4_a.1 THEN Total Cost LE / Mesqa  is c.2
59  is v.3_a.1 and year(P14)  is v.4_a.5 THEN Total Cost LE / Mesqa  is c.2
60  is v.3_a.1 and year(P14)  is v.4_a.5 THEN Total Cost LE / Mesqa  is c.2
61  is v.3_a.1 and year(P14)  is v.4_a.3 THEN Total Cost LE / Mesqa  is c.1
62  is v.3_a.2 and year(P14)  is v.4_a.5 THEN Total Cost LE / Mesqa  is c.6
63  is v.3_a.1 and year(P14)  is v.4_a.3 THEN Total Cost LE / Mesqa  is c.1

```
> ## Plot the membership functions
> par(mar = rep(2, 4))
> plotMF(object.reg)
> newdata <- data.tst
> ## generate the model and save it as object.WM
> object.SBC <- frbs.learn(data.train, range.data, method.type, control)
  |===============================================
| 62%
Error in loadNamespace(name) : there is no package called 'e1071'
In addition: Warning message:
In if (class(data.train) != "matrix") { :
  the condition has length > 1 and only the first element will be used
> ## the prediction process
> ## The following code can be used for all methods
> res <- predict(object.SBC, newdata)
Error in predict(object.SBC, newdata) : object 'object.SBC' not found
> res
```